\documentclass[10pt,journal,compsoc]{IEEEtran}
\ifCLASSOPTIONcompsoc
  \usepackage[nocompress]{cite}
\else
  \usepackage{cite}
\fi

\ifCLASSINFOpdf
\else
\fi

\usepackage{graphicx}
\usepackage{amsmath}
\usepackage{amssymb}
\usepackage{booktabs}
\usepackage{xcolor}
\usepackage[english]{babel} 
\usepackage[T1]{fontenc}
\usepackage{array}
\usepackage{multirow}
\usepackage{amstext}
\PassOptionsToPackage{normalem}{ulem}
\usepackage{ulem}
\usepackage{bbding}
\usepackage{rotating}
\usepackage{colortbl}
\usepackage[colorlinks,linkcolor=red]{hyperref}

\begin{document}
\title{ZITS++: Image Inpainting by Improving the Incremental Transformer on Structural Priors}

\author{Chenjie Cao$^\star$,
        Qiaole Dong$^\star$,
        Yanwei Fu$^\dag$
        
\thanks{$^\star$ indicates equal contributions, $^\dag$ refers to the corresponding author.}

\IEEEcompsocitemizethanks{\IEEEcompsocthanksitem Chenjie Cao, Qiaole Dong, and Yanwei Fu are with Fudan University, China.  E-mail: \{20110980001,qldong18,yanweifu\}@fudan.edu.cn.
}
}

\markboth{Journal of \LaTeX\ Class Files,~Vol.~14, No.~8, August~2015}%
{Shell \MakeLowercase{\textit{et al.}}: Bare Demo of IEEEtran.cls for Computer Society Journals}


\IEEEtitleabstractindextext{%
\begin{abstract}
Image inpainting involves filling missing areas of a corrupted image.
Despite impressive results have been achieved recently, restoring images with both vivid textures and reasonable structures remains a significant challenge. Previous methods have primarily addressed regular textures while disregarding holistic structures due to the limited receptive fields of Convolutional Neural Networks (CNNs).
To this end, we study learning a Zero-initialized residual addition based Incremental Transformer on Structural priors (ZITS++), an improved model upon our conference work, ZITS~\cite{dong2022incremental}.
Specifically, given one corrupt image, we present the Transformer Structure Restorer (TSR) module to restore holistic structural priors at low image resolution, which are further upsampled by Simple Structure Upsampler (SSU) module to higher image resolution.
To recover image texture details, we use the Fourier CNN Texture Restoration (FTR) module, which is strengthened by Fourier and large-kernel attention convolutions. 
Furthermore, to enhance the FTR, the upsampled structural priors from TSR are further processed by Structure Feature Encoder (SFE) and optimized with the Zero-initialized Residual Addition (ZeroRA) incrementally. Besides, a new masking positional encoding is proposed to encode the large irregular masks. 
\textcolor{black}{Compared with ZITS, ZITS++ improves the FTR's stability and inpainting ability with several techniques. More importantly, we comprehensively explore the effects of various image priors for inpainting and investigate how to utilize them to address high-resolution image inpainting with extensive experiments. This investigation is orthogonal to most inpainting approaches and can thus significantly benefit the community. Codes, dataset, and models will be released in \url{https://github.com/ewrfcas/ZITS-PlusPlus}.}
% Extensive experiments on various datasets validate the efficacy of our model compared with other competitors. We also conduct extensive ablation to compare and verify various priors for image inpainting tasks.
% Typically, FTR can be independently pre-trained  without image structural priors.
\end{abstract}

\begin{IEEEkeywords}
Image inpainting, Large-kernel attention convolution, Learning-based edges, Positional encoding.
\end{IEEEkeywords}}

\maketitle

\IEEEdisplaynontitleabstractindextext

% \fi
\IEEEpeerreviewmaketitle

\IEEEraisesectionheading{\section{Introduction}
\label{sec:intro}}

\begin{figure*}
\begin{centering}
\includegraphics[width=0.99\linewidth]{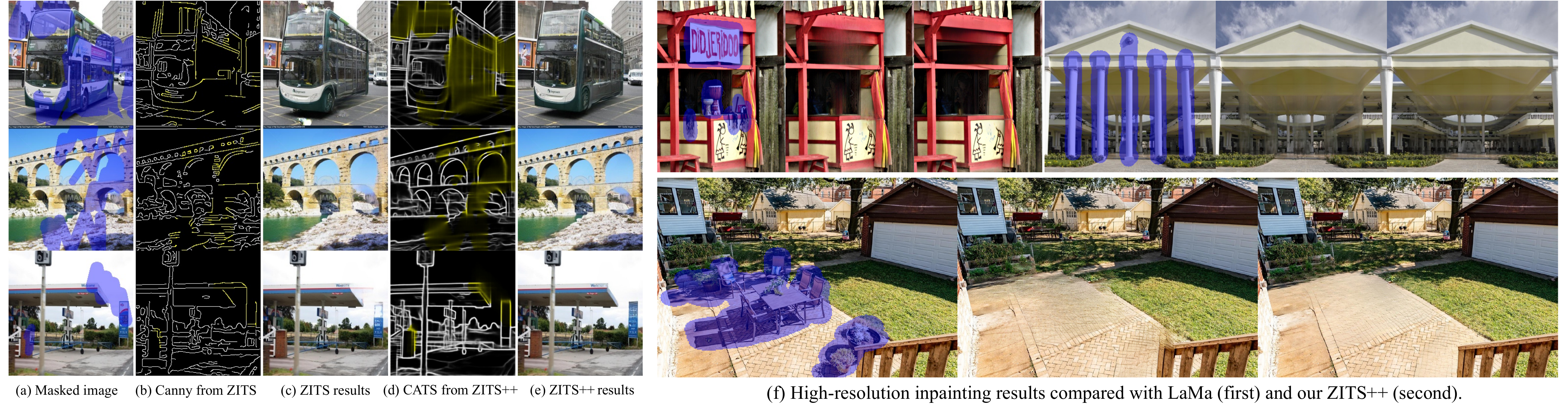}
\par\end{centering}
\vspace{-0.1in}
 \caption{Left (a)-(e): Comparisons of ZITS~\cite{dong2022incremental} and ZITS++ enhanced with canny and CATS respectively.
 Right (f): High-resolution (1024$\times$768, 1024$\times$1024, 680$\times$1024) inpainted results. Each image group: original image with inpainting mask, inpainted results by LaMa~\cite{suvorov2021resolution}, and results of ZITS++. Please zoom-in for details.
 \label{fig:teaser}}
 \vspace{-0.15in}
\end{figure*}

\begin{figure}
\begin{centering}
\includegraphics[width=0.75\linewidth]{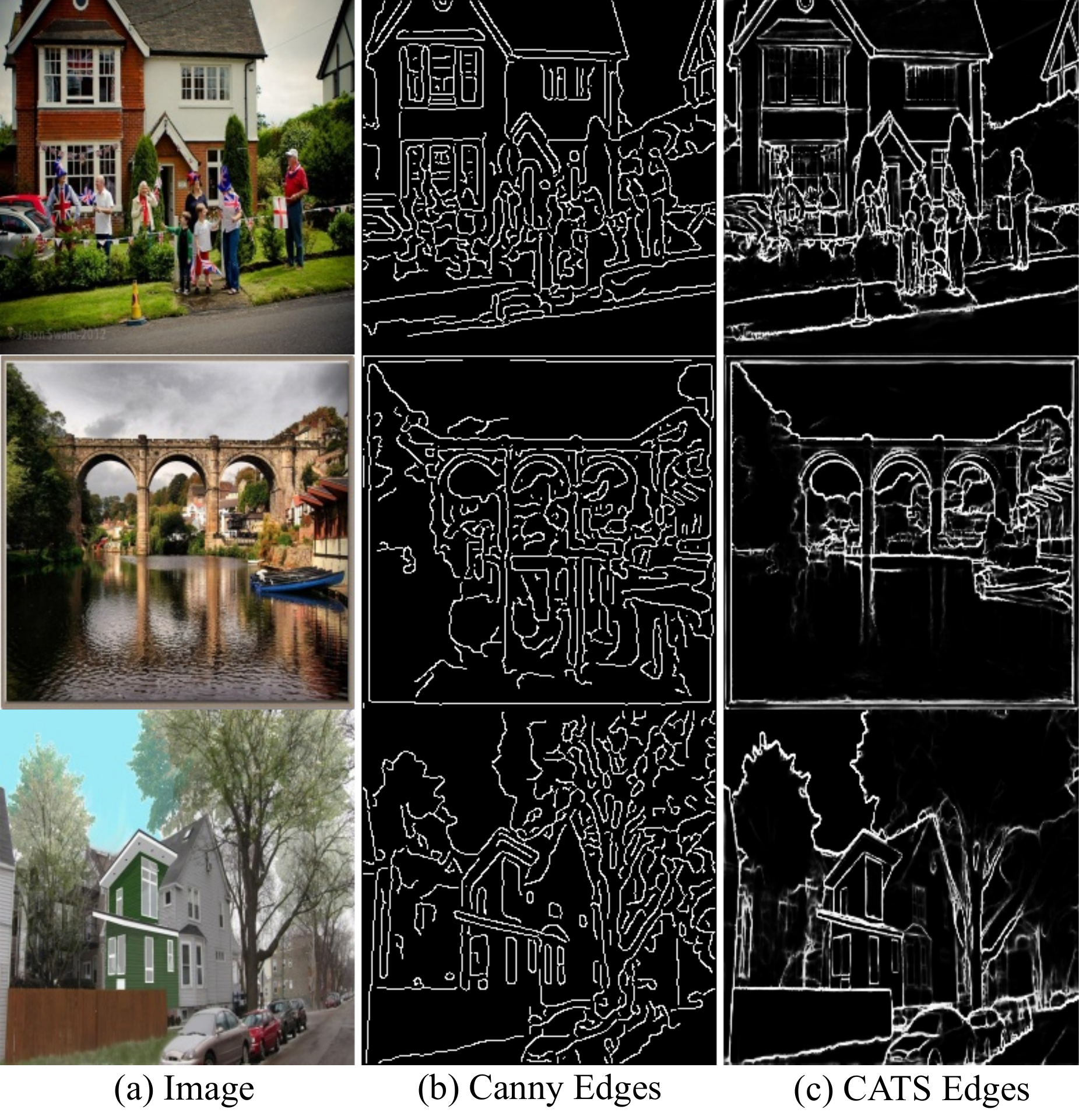}
\par\end{centering}
\vspace{-0.1in}
 \caption{\textcolor{black}{Edge results from (b) canny detection~\cite{canny1986computational} and (c) CATS~\cite{huan2021unmixing} learning based edge detection. We use $\sigma=2$ for canny, which is also adapted by most other inpainting methods~\cite{nazeri2019edgeconnect,cao2021learning,guo2021image}. Note that CATS edges have much clearer boundary structures and ignore high-frequency textures. But canny produces many meaningless edges from textures in column 1(b).}
 \label{fig:canny_cats_compare}}
 \vspace{-0.15in}
\end{figure}

\begin{figure*}
\begin{centering}
\includegraphics[width=0.9\linewidth]{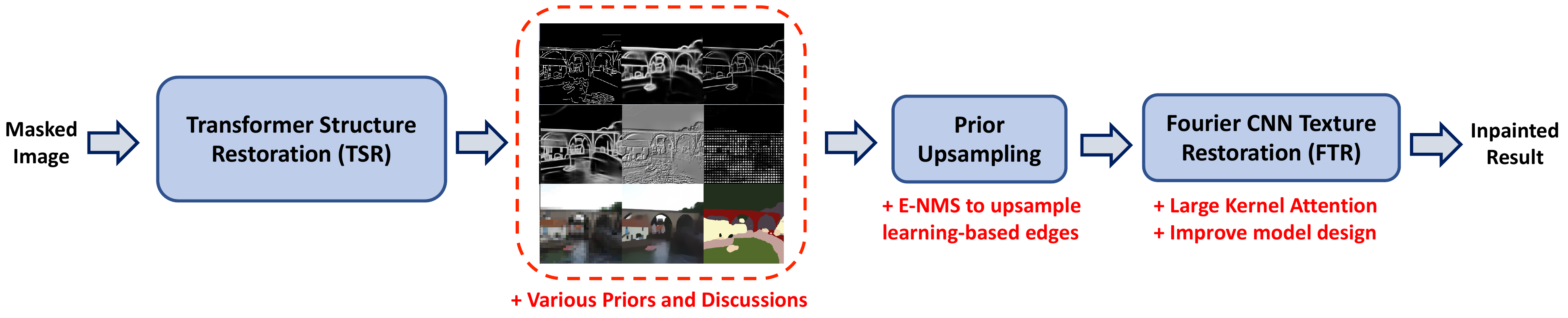}
\par\end{centering}
\vspace{-0.1in}
 \caption{The conceptual overview of ZITS++ based on ZITS~\cite{dong2022incremental}. We further highlight the newly proposed contributions in red.
 \label{fig:bird_eye_view}}
 \vspace{-0.15in}
\end{figure*}

Image inpainting is a long-standing challenge that aims at filling missing areas in pictures with plausible pixels. It has various real-world applications, including object removal~\cite{elharrouss2020image}, photo restoration, and image editing~\cite{jo2019sc}.
To achieve realistic outcomes, the inpainted images should maintain both semantically coherent textures in the low-level color space and visually reasonable structures for high-level human perception. 

Many classical image inpainting algorithms~\cite{Bertalmo2000ImageI,Levin2003LearningHT,Roth2005FieldsOE,Hays:2007,criminisi2003object} heuristically search similar patches for the reconstruction of missing image regions. 
Benefiting from the excellent capacities of CNNs~\cite{krizhevsky2012imagenet} and Generative Adversarial Networks (GANs)~\cite{goodfellow2014generative}, existing deep learning methods~\cite{guo2021image,cao2021learning,wan2021high,yi2020contextual,zhao2021large,li2022mat} could address the image inpainting tasks in some common cases.
\textcolor{black}{Particularly, to achieve high-fidelity reconstructions with realistic textures, deep learning models should have large receptive fields and preserve holistic structures~\cite{dong2022incremental}. 
However, preserving both good textures and holistic structures of the corrupted image remains intractable for these methods, despite some pioneering works~\cite{nazeri2019edgeconnect,wan2021high,suvorov2021resolution,yi2020contextual,cao2021learning,dong2022incremental} have partially solved such challenges.}

% Some pioneering works can partially solve these problems. 
Many inpainting works employ the attention mechanism to expand limited receptive fields~\cite{yu2018generative,yi2020contextual,zeng2020high}. Additionally, Fast Fourier Convolution (FFC) of global receptive fields is utilized in~\cite{suvorov2021resolution} to  
encode features in the frequency domain for resolution-robust inpainting. But the holistic structure is not explicitly modeled in~\cite{suvorov2021resolution}.  
Some transformer-based  methods~\cite{wan2021high,yu2021diverse} with long-range dependency first fill Low-Resolution (LR) tokens and then upsample them with CNNs to address the huge memory footprint of transformers for large-scale images.
\textcolor{black}{On the other hand, various image priors have been explored for inpainting to preserve specific information, rather than directly address the ill-posed image inpainting}, such as canny edges~\cite{nazeri2019edgeconnect,guo2021image}, segmentation~\cite{liao2020guidance,song2018spg}, Relative Total Variation (RTV)~\cite{ren2019structureflow,liu2020rethinking}, learning based edges~\cite{xu2020e2i}, gradients~\cite{yang2020learning}, and even semantic prior features~\cite{zhangijcai2021}.
Critically, Cao \textit{et al.}~\cite{cao2021learning} propose the concept of sketch tensor space, and exploit the structural priors of edges and wireframes~\cite{huang2018learning} in the space to facilitate holistic structure recovery for image inpainting. However, these methods typically involve multi-stage or multi-model designs and are costly to be trained from scratch.

\textcolor{black}{Our previous conference work, ZITS~\cite{dong2022incremental}, alleviates these issues with a well-designed architecture. Specifically, ZITS leverages a transformer-based model to recover canny edges and wireframe lines in low-resolution (LR) images and then upsamples these structural priors by a simple CNN module to High-Resolution (HR) cases. The upsampled edges and lines are used to strengthen a LaMa-based texture restoration. Moreover, ZITS also provides position clues for masked regions, which play a critical role in image generation~\cite{islam2020much,xu2020positional,lin2021infinitygan}. To overcome the heavy training cost of the multi-stage model, ZITS repurposes an incremental training strategy called Zero-initialized Residual Addition (ZeroRA)~\cite{bachlechner2020rezero} to finetune the off-the-shelf LaMa with holistic structure guidance.
Despite these advancements, ZITS still faces some intractable dilemmas. 
\textit{1) Canny edges used in ZITS fail to distinguish meaningful structures.} As shown in Fig.~\ref{fig:canny_cats_compare}(b), canny edges produce confusing textures in complicated environments rather than informative low-level structures.
\textit{2) There is a need for comprehensive investigations into various image priors.}
While ZITS only considers canny edges and lines, other image priors also need to be explored. Additionally, understanding how these priors can be used in HR images is important for the inpainting community.
\textit{3) Improving LaMa's performance for texture recovery.} Although LaMa~\cite{suvorov2021resolution} performs well in texture recovery with robust resolutions, it still suffers from unstable training and could be further enhanced for the inpainting task.
}

% high level
In this paper, we systematically address these problems upon our conference version~\cite{dong2022incremental} and propose an improved ZeroRA-based Incremental Transformer Structure (ZITS++) inpainting framework. 
\textcolor{black}{We first briefly introduce our main ideas as follows.}
Particularly, Learning-based Edges (L-Edges) are advocated as the important representations for meaningful holistic structures as illustrated in Fig.~\ref{fig:canny_cats_compare}(c).
A transformer-based model inherited from ZITS is presented to restore holistic structures that incorporate L-Edges, wireframe lines, and gradients.
\textcolor{black}{Critically, to adaptively leverage the structure upsampling mechanism of ZITS, we further iteratively upscale the grayscale L-Edges to arbitrary resolutions filtered by the Edge Non-Maximum Suppression (E-NMS).}
For the FFC-based Texture Restoration, we make significant improvements by incorporating Large Kernel Attention (LKA) modules~\cite{guo2022visual} and other useful techniques. 
Importantly, we comprehensively study various image priors, including canny edges~\cite{canny1986computational}, wireframes~\cite{huang2018learning}, L-Edges~\cite{xie2015holistically}, gradients, semantic segmentation, RTV~\cite{xu2012structure}, LR-RGB and Histograms of Oriented Gradients (HOG)~\cite{dalal2005histograms}. 
Through empirical analysis in Sec.~\ref{sec:priors} and Sec.~\ref{sec:ablation_priors}, we find that L-Edges, wireframe lines, and gradients are generally more effective than other priors. Moreover, L-Edges could also be effectively generalized to HR with E-NMS, which thoroughly outperforms canny edge. 
We highlight our new contributions of ZITS++ in Fig.~\ref{fig:bird_eye_view}. ZITS++ inherits the advantages of ZITS, including the decoupled structure and texture recoveries, flexible structure upsampling, and a lightweight finetuning strategy based on ZeroRA. We further enhance ZITS++ with stronger texture learning ability and more informative structural prior combinations, as demonstrated through convincing experiments and discussions. 

\textcolor{black}{The workflow of ZITS++ and related technical details are listed in this paragraph.} Formally, ZITS++ first recovers structural priors by a Transformer Structure Restorer (TSR) in 256$\times$256. 
% These priors are upsampled by the Simple Structure Upsampler (SSU) and passed into Structure Feature Encoder (SFE) to extract structural features. 
After the prior restoration, E-NMS is leveraged to eliminate ambiguous blur of the L-Edges' boundaries. 
Then the output grayscale priors are upsampled by a Simple Structure Upsampler (SSU) which consists of a 4-layer CNN to meet the desired resolution.
Upsampled priors are passed into Structure Feature Encoder (SFE) to extract structural features.
After that, the FCC-based Texture Restoration (FTR) is used to restore textures, trained with ZeroRA and Masking Positional Encoding (MPE).
Remarkably, we propose several novel techniques to improve the training process and significantly enhance FTR's performance. 
Specifically, we promote the \emph{maxpool} as the mask resizing strategy of PatchGAN instead of the \emph{nearest} in LaMa. This technique leads to stable training as empirically evaluated in Sec.~\ref{sec:FTR}.
Compared with the vanilla attention mechanism, LKA implemented by integrated convolutions~\cite{guo2022visual} is further repurposed in FTR and enjoys both \emph{large receptive fields} and \emph{scale invariance}, which greatly strengthens the FTR in HR inpainting. 
To our knowledge, this is the first time that extremely large kernels have been used in inpainting tasks.

Finally, we conduct extensive experiments on several datasets, including Places2~\cite{zhou2017places}, Indoor~\cite{cao2021learning, huang2018learning, Silberman:ECCV12}, MatterPort3D~\cite{chang2017matterport3d}, FFHQ~\cite{karras2019style}, and our newly collected HR-Flickr. The HR-Flickr dataset, which contains a group of high-quality images, serves as a test set for HR image inpainting. These experiments reveal that our proposed model outperforms other state-of-the-art competitors.

\textcolor{black}{As a substantial extension of ZITS~\cite{dong2022incremental}, we highlight the new contributions of ZITS++ as follows: 
(1) The initial work ZITS effectively recovers LR canny edges and lines with TSR and upscales them to HR for the image inpainting. This work further explores more image priors with extensive experiments. We find that L-Edges enjoy more valuable information for image inpainting.
(2) We introduce the E-NMS technique to flexibly upsample L-Edges to arbitrary resolutions, making them adaptive to the SSU introduced in ZITS.
This technique provides critical structural clues for HR inpainting. 
(3) Our FTR is further improved with the maxpool mask resizing strategy and LKA, resulting in more stable training and better performance.
(4) Additionally, we contribute to the inpainting community with a high-quality HR image dataset--HR-Flickr, and conduct extensive inpainting experiments on various resolutions.}

\section{Related Work}

\subsection{Image Inpainting}
Many classical approaches are explored to solve inpainting problems, which can be grouped as diffuision-based~\cite{li2017localization} and patch-based~\cite{ruzic2015context} methods. Unfortunately, these methods are typically restricted to low-level image features, and hard to generate expressive results with large holes. Learning-based methods equipped with GANs achieve great success in image inpainting~\cite{pathak2016context,nazeri2019edgeconnect,zhao2021large,li2022mat}. 
Concretely, Pathak \emph{et al.}~\cite{pathak2016context} show that adversarial training is critical to generate clear results for the inpainting. Furthermore, many \textit{bespoke} inpainting convolutions are designed~\cite{liu2018image,yu2019free,zeng2022aggregated} to gracefully model generalized features for masked regions. Moreover, the attention mechanism is also well explored in the inpainting~\cite{yu2018generative,yu2019free,yi2020contextual}, which takes long-range relation and aggregates reliable features from unmasked regions to masked ones.
Recently, StyleGAN~\cite{karras2020analyzing} based co-modulated inpainting models~\cite{zhao2021large,li2022mat} have achieved high-quality generations enhanced with the style learning. Nevertheless, they~\cite{zhao2021large,li2022mat} are quite likely to hallucinate incompatible image content over the mask.
\textcolor{black}{Diffusion models~\cite{dhariwal2021diffusion} have also developed rapidly in image generation recently.
However, training an inpainting-specific diffusion model
demands prohibitive computational cost with huge model parameters~\cite{rombach2022high,saharia2022palette,nichol2021glide}. While much more diffusion steps should be considered to infer the inpainted results as an inverse problem with off-the-shelf diffusion models~\cite{lugmayr2022repaint,chung2022come,chung2022improving}. Thus this is still very time-consuming.}
Although these pioneering works have tackled various problems in inpainting, they are generally incapable of reconstructing structural priors to maintain faithful inpainting results.

\begin{figure*}
\begin{centering}
\includegraphics[width=0.75\linewidth]{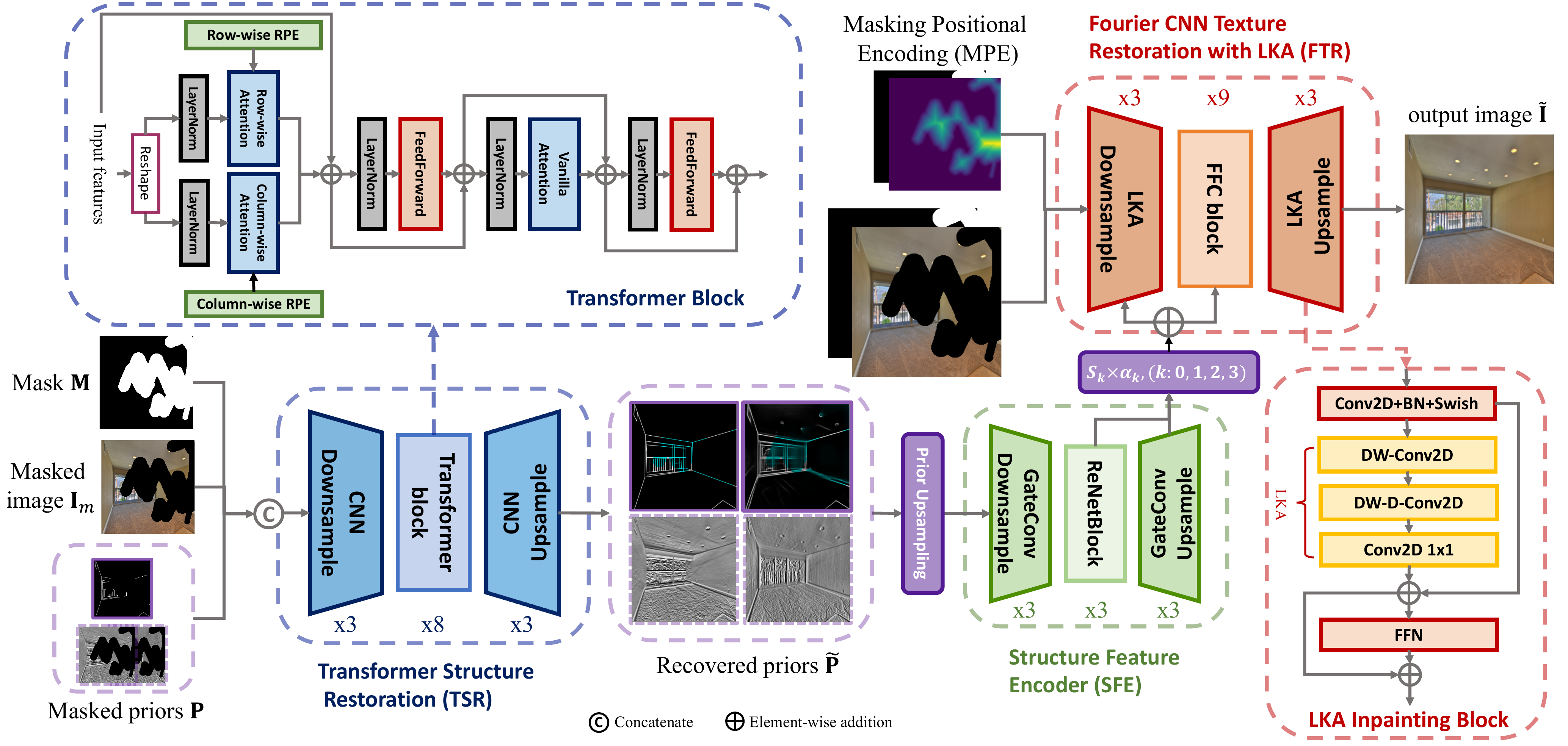}
\par\end{centering}
\vspace{-0.1in}
 \caption{Overview of ZITS++. Given corrupted images, TSR is used to restore structures at low resolution. Then, the upsampled sketch space is encoded by SFE and added to the FTR through ZeroRA to restore the textures. The top left corner shows details about the transformer block, \textcolor{black}{while the bottom right corner illustrates the LKA inpainting block.}
\label{fig:overview}}
\vspace{-0.15in}
\end{figure*}

\subsection{Inpainting by Auxiliaries}
Auxiliary information such as canny edges~\cite{nazeri2019edgeconnect,yang2020learning}, segmentation maps~\cite{song2018spg, liao2020guidance}, L-Edges~\cite{xu2020e2i} and gradients~\cite{yang2020learning} shown promising improvements to image inpainting. 
Specifically, EdgeConnect~\cite{nazeri2019edgeconnect} utilizes canny edges to facilitate inpainting images with certain structures. E2I~\cite{xu2020e2i} firstly tried to restore images with a two-stage inpainting model enhanced by HED~\cite{xie2015holistically}. Guo~\textit{et al.}~\cite{guo2021image} propose a two-stream network for image inpainting, which models the structure-constrained texture synthesis and texture-guided structure reconstruction in a coupled manner. SGE-Net~\cite{liao2020guidance} iteratively updates the semantic segmentation maps and the corrupted image.
Moreover, RTV is used to filter high-frequency textures for structural recovery~\cite{ren2019structureflow} and the decoupling structure/texture feature learning~\cite{liu2020rethinking}.
Our previous work~\cite{cao2021learning} further proposes learning a sketch tensor space, composed of canny edges and lines for inpainting man-made scenes. 
\textcolor{black}{Besides, for video inpainting, optical flow plays an important role to guide the sequence completion~\cite{gao2020flow,li2022towards,zhang2022flow}.}
In this paper, inspired by the comprehensive study on auxiliary priors in Sec.~\ref{sec:priors} for the single view inpainting, L-Edges and lines are employed to model the structural priors.
However, quite differs from~\cite{cao2021learning}, the transformer is explored to model structural priors in ZITS++. Preliminary investigations~\cite{chen2020generative} have shown the excellent capability of transformers in modeling structural relationships for natural image synthesis.
\textcolor{black}{Besides, we provide sufficient discussions about various image priors for image inpainting and their performance extended to HR images. Hopefully, this could bring useful insights to the community.}

\subsection{Deep Learning Modules}
\textcolor{black}{
In this section, we provide some discussions about relevant deep learning modules. Transformers have been utilized in some inpainting works, while CNNs \textbf{with large kernels} are first employed for image inpainting in this paper.
}

\noindent\textcolor{black}{\textbf{Transformers for Image Inpainting.}} Transformers~\cite{vaswani2017attention} achieved good performance on many vision tasks by learning long-range interactions on sequential image patches~\cite{dosovitskiy2020image,esser2021taming,ramesh2021zeroshot,he2022masked}.
Transformer is also employed to inpaint image in~\cite{wan2021high,yu2021diverse} at low resolution, and further guides GAN-based CNN to produce high-quality results.
Unfortunately, directly using transformers to learn large image patches demands a huge memory footprint and computations. In contrast, this work utilizes a transformer to build LR holistic structure reconstruction, which helps to guide HR image inpainting. Note that the upsampling for grayscale structures enjoys fewer ambiguities and information loss, which successfully transfers structure clues to HR.

\noindent\textcolor{black}{\textbf{CNNs with Large Kernels.}}
% Since the attention mechanism has achieved remarkable success in many vision tasks~\cite{dosovitskiy2020image,he2022masked} benefited by long-range dependencies, 
CNNs with large kernels attract some research attention recently. Except for the global CNN learning in the Fourier domain~\cite{chi2020fast}, Ding \emph{et al.}~\cite{ding2022scaling} propose to enlarge the kernel size to 31$\times$31 with several tricks and achieve better performance in many downstream tasks. The kernel size is further enlarged to 51 with the sparsity in~\cite{liu2022more}. To balance the computation, large kernel convolutions are decomposed into a depth-wise convolution, a depth-wise dilation convolution, and a point-wise convolution in~\cite{guo2022visual}.
We show that such approximately large kernels can work properly in image inpainting compared with vanilla convolutions, dilated ones, and even FFC.

\section{Method}

\noindent\textbf{Overview.} The whole pipeline of ZITS++ is illustrated in Fig.~\ref{fig:overview}. Given masked image $\textbf{I}_m$, \textcolor{black}{masked priors $\textbf{P}$,}
and binary mask $\textbf{M}$, we concatenate and input them to the TSR model for \textcolor{black}{restored priors $\mathbf{\tilde{P}}=\mathrm{TSR}(\textbf{I}_m,\textbf{P},\textbf{M})$ (Sec.~\ref{sec:TSR}). 
Note that the priors of L-Edges are directly produced by TSR. And we include an extensive discussion of priors in  Sec.~\ref{sec:priors}. 
Then the selected grayscale structures (lines and edges) can be easily combined and upsampled into arbitrary sizes for HR inpainting (Sec.~\ref{sec:SSU}).}
Gated convolution based SFE extracts multi-scale features $\{\mathbf{S}_k\}_{k=0}^3=\mathrm{SFE}(\mathbf{\tilde{P}},\textbf{M})$ from upsampled sketches. We incrementally add $\{\mathbf{S}_k\}_{k=0}^3$ to related layers of the FTR enhanced by \textcolor{black}{LKA} and Fourier convolutions  as $\tilde{\textbf{I}}=\mathrm{FTR}(\textbf{I}_m,\textbf{M},\{\alpha_k\cdot\textbf{S}_k\}_{k=0}^3)$ (Sec.~\ref{sec:FTR}) with the residual addition of zero-initialized trainable parameter $\alpha_k$ (ZeroRA) (Sec.~\ref{sec:SFEandMPEandZeroRA}).

\subsection{Transformer Structure Restoration (TSR)}
\label{sec:TSR}

Since the transformer shows an ability to get expressive global structure recoveries~\cite{wan2021high}, we leverage the capacity of the transformer to restore holistic structures in a relatively low resolution.
For the input \textcolor{black}{masked priors $\mathbf{P}$,} and mask $\textbf{M}$ in $256\times256$, we firstly downsample them with three convolutions to reduce computation for attention learning. Such simple convolutions can also inject beneficial convolutional inductive bias for vision transformers compared with the patch-based MLP embedding~\cite{xiao2021early}.
Then we add a learnable absolute position embedding to the feature at each spatial position and get $\mathbf{X}\in\mathbb{R}^{h\times w\times c}$ for the input to attention layers, where both the height $h$ and width $w$ are 32, while $c=256$ is the feature channel.

To overcome the quadratic complexity of standard self-attention~\cite{vaswani2017attention}, we alternately use axial attention modules~\cite{ho2019axial,huang2019ccnet} and standard attention modules as illustrated in   Fig.~\ref{fig:overview}(top-left). The axial attention module can be implemented easily by adjusting the tensor shape for row-wise/column-wise and then processing them with dot product-based self-attention respectively. To improve the spatial relation, we also provide Relative Position Encoding (RPE)~\cite{raffel2019exploring} for each axial-attention module. For the input feature $\mathbf{X}\in\mathbb{R}^{h\times w\times c}$, we suppose that \textcolor{black}{$\mathbf{x}_{ri},\mathbf{x}_{rj},\mathbf{x}_{ci},\mathbf{x}_{cj}\in\mathbb{R}^{c}$} mean feature vectors of rows $i,j$ and columns $i,j$ of $\mathbf{X}$.
Then the row and column-wise RPE based axial attention scores $\mathbf{A}^{row}, \mathbf{A}^{col}$ can be written as
\begin{equation}
\begin{split}
\mathbf{A}_{i,j}^{row}&=\mathbf{x}_{ri}\mathbf{W}_{rq}\mathbf{W}_{rk}^T\mathbf{x}_{rj}^T+R_{i,j}^{row},\\
\mathbf{A}_{i,j}^{col}&=\mathbf{x}_{ci}\mathbf{W}_{cq}\mathbf{W}_{ck}^T\mathbf{x}_{cj}^T+R_{i,j}^{col},
\end{split}
\label{eq:axial_attention}
\end{equation}
where $\mathbf{W}_{rq},\mathbf{W}_{rk},\mathbf{W}_{cq},\mathbf{W}_{ck}$ are trainable parameters for query and key in row and column; $R_{i,j}^{row}$ is the trainable RPE value between row $i$ and $j$, and $R_{i,j}^{col}$ means the RPE value between columns $i,j$. Then, the attention scores are processed by the softmax operation.
To stabilize the training, we advocate the pre-norm trick in~\cite{xiong2020layer}.
Compared with the $O(n^2)$ complexity of the standard self-attention, the axial attention only has $O(2n^{\frac{3}{2}})$, which allows us can handle more attention layers for a better model capacity. 
Additionally, we also retain some vanilla attention modules in TSR for learning better global correlation.

After the encoding of stacked transformer blocks, features are upsampled by three transpose convolutions for outputting structures $\mathbf{\tilde{P}}$ in 256$\times$256 resolution. 
Various priors could be recovered by TSR; and corresponding
loss functions are specified in Sec.~\ref{sec:priors}.  
The motivation for processing the structural priors in 256$\times$256 is to save the computation and memory for the attention-based TSR. 
And we further show our SSU module in Sec.~\ref{sec:SSU} can flexibly upsample these priors to a higher resolution.
\textcolor{black}{Note that more detailed structural priors are unnecessary as compared between L-Edges and canny edges in our experiments. Because TSR should focus on the low-frequency recovery while FTR could make up for the high-frequency missing.}

\subsection{Upsampling Priors}
\label{sec:SSU}

\begin{figure}
\begin{centering}
\includegraphics[width=0.9\linewidth]{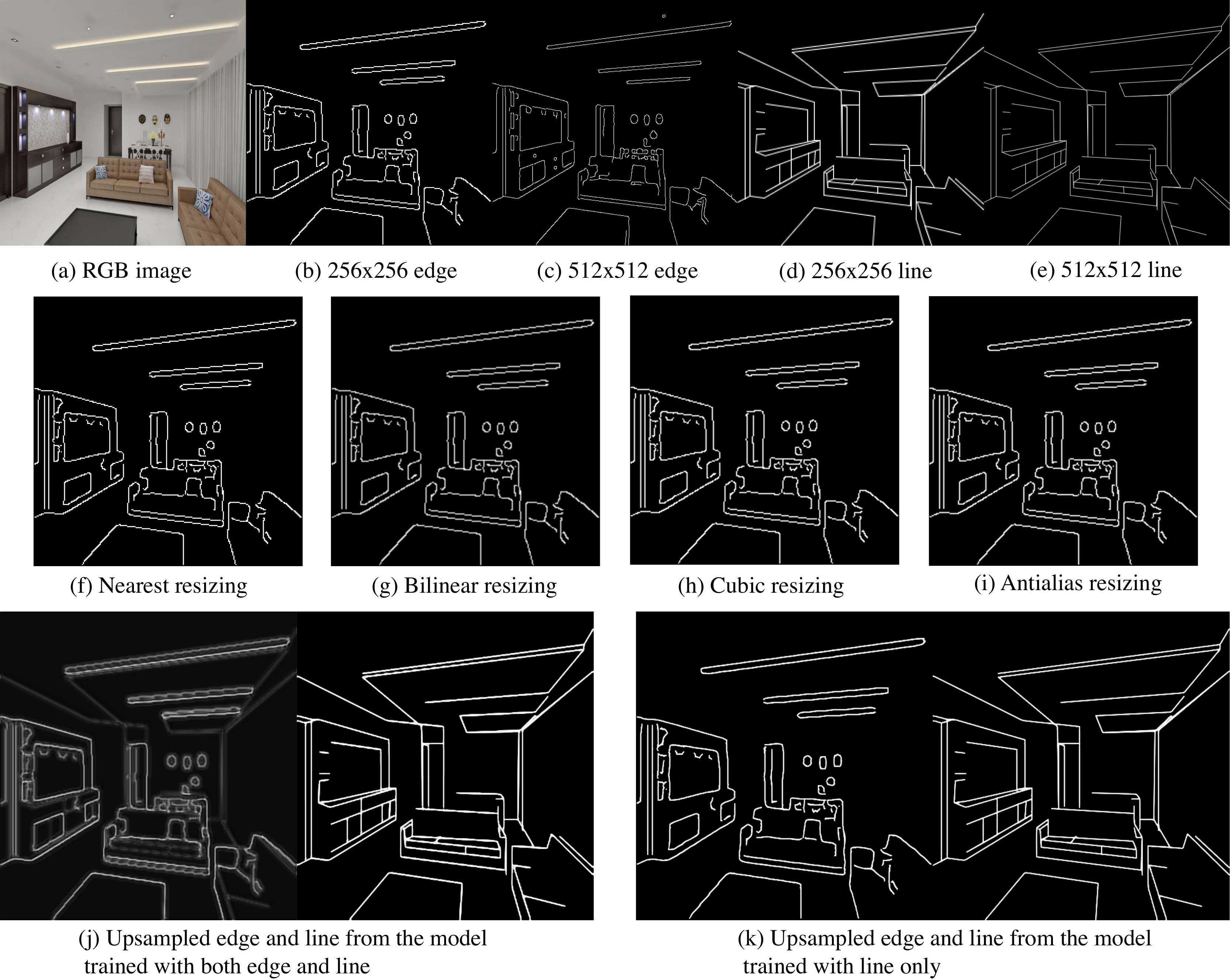}
\par\end{centering}
\vspace{-0.1in}
 \caption{(a)--(e) indicate the ground truth images and structures. 
%  Canny edges are got by $\sigma=2$ for 256$\times$256 and $\sigma=2.5$ for 512$\times$512. 
\textcolor{black}{(b) and (c) show obvious ambiguities between canny edges extracted from different image scales.} (f)--(i) show resizing edges of different interpolations. The learning-based upsampling edge from (k) has significantly superior quality compared with the one from (j). \label{fig:upsample}}
\vspace{-0.15in}
\end{figure}

\begin{figure}
\begin{centering}
\includegraphics[width=0.8\linewidth]{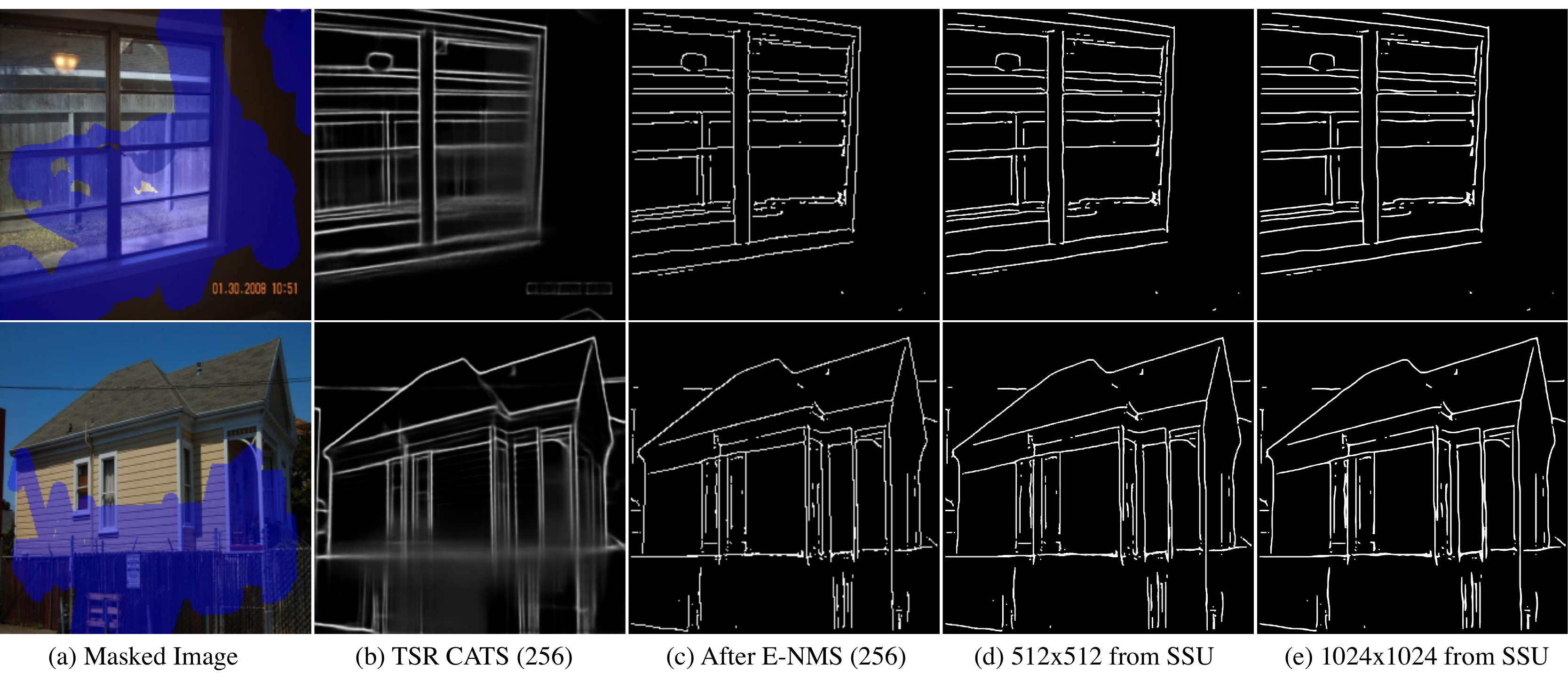}
\par\end{centering}
\vspace{-0.1in}
\caption{\textcolor{black}{Illustration of E-NMS and SSU work well for CATS edges, produced from TSR.}\label{fig:E-NMS}}
\vspace{-0.15in}
\end{figure}

\noindent\textbf{\textcolor{black}{Upsampling Edges and Lines.}} To capture holistic structures for higher resolution images, we should upsample the generated priors $\mathbf{\tilde{P}}$ to arbitrary scales without obvious degeneration. 
Particularly, upsampled structures can achieve superior performance as empirically analyzed in Sec.~\ref{sec:ablation_priors}.
However, vanilla interpolation-based resizing causes the aliasing effect to lines as well as edges as in Fig.~\ref{fig:upsample}(f)--(i). Such artifacts are more serious for large image sizes and deteriorate the inpainted results. To address this issue,
we train a CNN-based SSU to upsample lines to a doubled size. Since lines obtained from a wireframe parser have good discrete representations~\cite{huang2018learning,xue2020holistically}, \emph{i.e.}, a line can be indicated as positions of two endpoints and their relation, we can draw line maps in various resolutions without any ambiguities as in Fig.~\ref{fig:upsample}(d) and Fig.~\ref{fig:upsample}(e). 
Through iterative calling, we can get high-quality lines at higher resolutions.
Remarkably, despite the lines can be upsampled successfully, edges fail to get correct results with the same training strategy as shown in Fig.~\ref{fig:upsample}(j). 
Because there are ambiguities in the canny edge from different image sizes as in Fig.~\ref{fig:upsample}(b) and Fig.~\ref{fig:upsample}(c).
\textcolor{black}{And L-Edges suffer from a similar issue, \textit{i.e.}, edges extracted from different image scales enjoy inconsistent results.}
Interestingly, we find that if the upsampling module is only trained in lines, it can also produce smoothed edge maps at higher resolution, thanks to the generalization of the network as in Fig.~\ref{fig:upsample}(k). 
\textcolor{black}{Such a good property can also be generalized for upsampling L-Edges after E-NMS as introduced below.}

\noindent\textbf{\textcolor{black}{E-NMS for L-Edges.}}
We promote the learning-based edges -- CATS~\cite{huan2021unmixing}, rather than canny edges in \cite{dong2022incremental}. 
Such priors can also be well generalized to the cases of higher resolution with the newly introduced E-NMS and our SSU trained with lines. 
As in Fig.~\ref{fig:E-NMS}, we take E-NMS in edge evaluation of~\cite{ZitnickECCV14edgeBoxes} and implement E-NMS in PyTorch to filter uncertain edge predictions. Then the filtered edges are further binarized. Thus SSU can iteratively upsample CATS edges as lines. E-NMS significantly eliminates ambiguous blur and artifacts near the boundaries and improves the performance during HR inpainting as in Sec.~\ref{sec:ablation_priors}. To avoid losing information potentially, we maintain low confident areas with resized CATS prediction, while high confident boundaries are processed as E-NMS boundaries. 
\textcolor{black}{E-NMS let L-Edges retain sufficient information in arbitrary resolutions. So it is unnecessary to iteratively Mask-Predict edges as in~\cite{dong2022incremental}, which saves lots of inference time.}
More details about the E-NMS are discussed in the supplementary.

\begin{figure}
\begin{centering}
\includegraphics[width=0.95\linewidth]{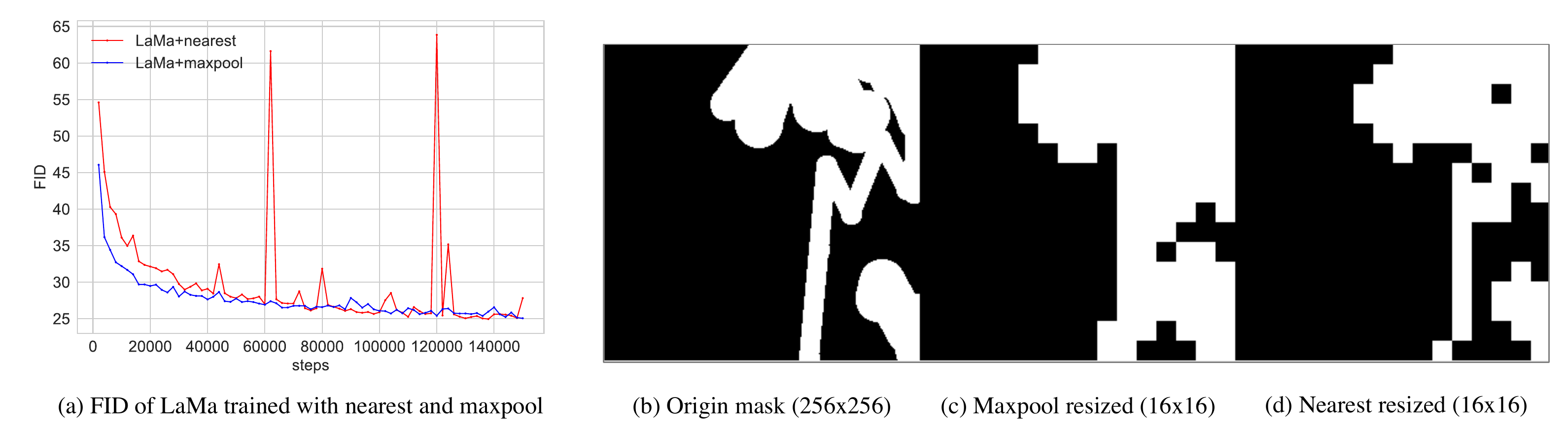}
\par\end{centering}
\vspace{-0.1in}
\caption{\textcolor{black}{The influence of different mask resizing strategies for LaMa on the Places2 subset of  25,000 training and 500 validation images over 5 scenes. LaMa worked with nearest resizing is unstable.}\label{fig:mask_resize}}
\vspace{-0.15in}
\end{figure}

\begin{table}
\caption{\textcolor{black}{Quantitative ablations of FTR based on the Places2 subset of 25,000 training and 500 validation images over 5 scenes. `AddConv' means additional convolution types utilized in the upsampling and downsampling modules of FTR. `Conv2D+' indicates using more 2D convolutions to balance the computation compared with FFC and LKA. Numbers in brackets beside the LKA mean effective kernel sizes.\label{table:FTR_ablation}}}
\vspace{-0.1in}
\centering
\footnotesize
\renewcommand\tabcolsep{1.0pt}
\begin{tabular}{c|c|c|c|c|cccc}
\toprule 
\multicolumn{2}{c|}{Resizing} & \multicolumn{2}{c|}{Activation} & \multirow{2}{*}{AddConv} & \multirow{2}{*}{PSNR$\uparrow$} & \multirow{2}{*}{SSIM$\uparrow$} & \multirow{2}{*}{FID$\downarrow$} & \multirow{2}{*}{LPIPS$\downarrow$}\tabularnewline
\cline{1-4} \cline{2-4} \cline{3-4} \cline{4-4} 
\multicolumn{1}{c|}{{\scriptsize{}Nearest}} & {\scriptsize{}Maxpool} & \multicolumn{1}{c|}{{\scriptsize{}ReLU}} & {\scriptsize{}Swish} &  &  &  &  & \tabularnewline
\midrule 
\CheckmarkBold{} &  & \CheckmarkBold{} &  & -- & 25.00 & 0.876 & 25.09 & 0.109\tabularnewline
 & \CheckmarkBold{} & \CheckmarkBold{} &  & -- & 25.07 & 0.876 & 25.05 & 0.110\tabularnewline
 & \CheckmarkBold{} & \CheckmarkBold{} &  & Conv2D+ & 24.99 & 0.874 & 25.42 & 0.111\tabularnewline
 & \CheckmarkBold{} & \CheckmarkBold{} &  & FFC & 25.07 & 0.876 & 24.15 & 0.107\tabularnewline
 & \CheckmarkBold{} & \CheckmarkBold{} &  & LKA(14) & 25.15 & \textbf{0.878} & 23.76 & 0.105\tabularnewline
 & \CheckmarkBold{} & \CheckmarkBold{} &  & LKA(21) & 25.12 & 0.877 & \textbf{23.31} & 0.104\tabularnewline
 & \CheckmarkBold{} & \CheckmarkBold{} &  & LKA(28) & 25.13 & 0.877 & 23.56 & 0.105\tabularnewline
 & \CheckmarkBold{} &  & \CheckmarkBold{} & LKA(21) & \textbf{25.18} & \textbf{0.878} & 23.44 & \textbf{0.102}\tabularnewline
\bottomrule 
\end{tabular}
\vspace{-0.1in}
\end{table}

\subsection{\textcolor{black}{Fourier CNN Texture Restoration with LKA (FTR)}}
\label{sec:FTR}
%\textbf{LaMa.} 

For the texture restoration, we adopt 
Fourier convolutions~\cite{chi2020fast} from LaMa~\cite{suvorov2021resolution} for learning in the frequency domain. Note that FTR also works as the pre-trained inpainting model for our ZeroRA finetuning with structural priors from TSR.
As in Fig.~\ref{fig:overview}, FTR is an autoencoder-based model. The key module of FTR is the FFC layer with two branches: 1) the local branch uses conventional convolutions and 2) the global branch convolutes features after the fast Fourier transform. Then two branches are combined for larger receptive fields and local invariance during the inpainting~\cite{suvorov2021resolution}. 
Unfortunately, LaMa didn't explicitly learn reasonable holistic structures, which is addressed in ZITS++.

\noindent\textcolor{black}{\textbf{Improving FTR from LaMa.} Although LaMa can produce high-quality inpainted results, the training of LaMa is quite unstable even with the EMA as compared in Fig.~\ref{fig:mask_resize}(a) on the Places2 subset. We expose that such an unstable training suffers from the resizing way of PatchGAN~\cite{isola2017image}. For the patch-wise fake loss optimization of the discriminator, LaMa leverages the \emph{nearest} resizing, where some partially masked patches are defined as unmasked ones as in Fig.~\ref{fig:mask_resize}(d). Instead, we find that using \emph{maxpool} resizing as Fig.~\ref{fig:mask_resize}(c) can greatly stabilize the training process. It means that partially masked patches should be considered as `fake' patches rather than `real' ones for inpainting.
Besides, we upgrade all activation in the generator of FTR to Swish~\cite{ramachandran2017searching}, which has been shown to be effective in many generation works~\cite{karras2020analyzing,esser2021taming}. Detailed ablation studies of the improved FTR are shown in Tab.~\ref{table:FTR_ablation}.
}

\noindent\textcolor{black}{\textbf{LKA Inpainting Module.} 
It has been well explored that large receptive fields are important for image inpainting~\cite{suvorov2021resolution,zeng2022aggregated}. FFC used in LaMa~\cite{suvorov2021resolution} has the global receptive field in the frequency domain. However, FFC fails to learn reliable feature relations between masked regions and unmasked ones.
% Dilated convolutions help to learn more meaningful features for masked regions from unmasked ones, which are widely used in many inpainting models~\cite{nazeri2019edgeconnect,cao2021learning,zeng2022aggregated}. 
On the other hand, various attention mechanisms with long-range dependencies are also incorporated into many inpainting methods~\cite{yu2018generative,yi2020contextual,li2022mat}. But these attention-based methods inevitably overfit certain resolutions without scale invariance. 
To unify the advantages of both CNN and attention, we propose to inject large-kernel based convolutions into FTR. 
We follow~\cite{guo2022visual} to decompose LKA with $K\times K$ receptive fields into a $\lceil\frac{K}{d}\rceil\times\lceil\frac{K}{d}\rceil$ Depth-Wise Convolution (DW-Conv2D) with dilation $d$, a $(2d-1)\times(2d-1)$ depth-wise dilation convolution (DW-D-Conv2D), and a point-wise convolution (Conv2d 1$\times$1) as  in the lower right corner of Fig.~\ref{fig:overview}. LKAs are incorporated into upsampling and downsampling blocks of FTR, which are orthogonal to FFCs. Because LKAs devote to learning better feature representations for masked regions with large receptive fields, FFCs tend to facilitate the global learning for regular textures in the frequency domain. 
%As discussed in~\cite{ding2022scaling},
Furthermore, 
we also use shortcuts with $3\times3$ convolutions as Feed Forward Networks to ensure the generalization of LKA. Our empirical ablations in Tab.~\ref{table:FTR_ablation} show that LKAs work better than FFCs and vanilla convolutions. Thus LKA with $K=21$ is used in ZITS++.
}

\subsection{ZeroRA Learning for Structural Priors}
\label{sec:SFEandMPEandZeroRA}
\subsubsection{Structure Feature Encoder (SFE)}
For the given restored \textcolor{black}{priors $\mathbf{\tilde{P}}$} in arbitrary scales,  a fully convolutional network is employed to process them into a feature space. Our SFE is also an autoencoder model with 3 layers of downsampling convolutions (encoder), 3 layers of residual blocks with dilated convolutions~\cite{yu2015multi} (middle), and 3 layers of upsampling convolutions (decoder). For the encoder and the decoder in SFE, we use Gated Convolutions (GCs)~\cite{yu2019free} to transfer useful features selectively. GC learns another sigmoid activation with the same channels. Then the sigmoid features are multiplied by the convoluted ones as outputs. Although GCs are widely used in image inpainting for better generalization to irregular masks, we use GCs to filter useful features to FTR. Because the grayscale sketch space is sparse, and not all features are necessary for the inpainting. Then, 4 coarse-to-fine feature maps $\{\mathbf{S}_k\}_{k=0}^3$ from the last middle layer and 3 decoder layers are selected to transfer structural features to FTR as
\textcolor{black}{
\begin{equation}
\mathbf{S}_0,\mathbf{S}_1,\mathbf{S}_2,\mathbf{S}_3=\mathrm{SFE}(\mathbf{\tilde{P}},\mathbf{M}),
\label{eq:sfe}
\end{equation}
}
where $\mathbf{M}$ indicates the resized binary mask.

\begin{figure}
\begin{centering}
\includegraphics[width=0.75\linewidth]{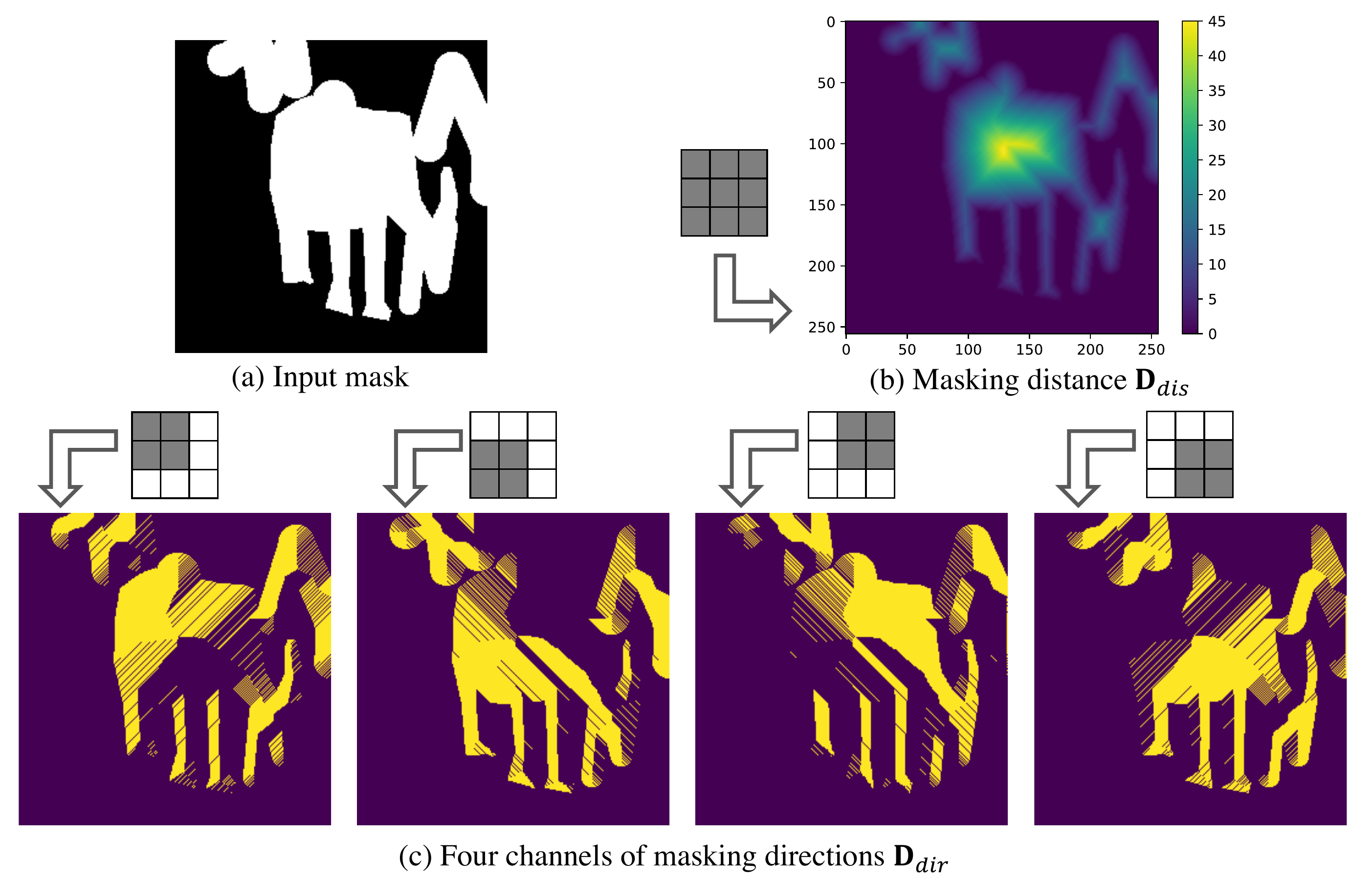}
\par\end{centering}
\vspace{-0.1in}
 \caption{The illustration of our masking relative position encoding. (a) Input mask, (b) masking distance $\mathbf{D}_{dis}$ and the all-one 3$\times$3 kernel, (c) masking directions $\mathbf{D}_{dir}$ and their kernels. \label{fig:MPE}}
\vspace{-0.15in}
\end{figure}

\subsubsection{Masking Positional Encoding (MPE)}
Although the zero-padding in CNNs can provide some position information~\cite{islam2020much}, it only contains information about spatial anchors~\cite{xu2020positional}. Therefore, generated central regions from GANs tend to repeat meaningless artifacts without specific position encoding. When the image size is large, the effect of zero-padding will be further weakened, which causes more repeated artifacts~\cite{lin2021infinitygan} to generators.

During the inpainting, position information for unmasked regions is unnecessary, because the model always knows the ground truth of unmasked image regions. However, we think that position information is still critical for masked regions, especially when mask areas are large for HR images. Limited by the receptive fields of CNNs, the model with large masks may lose the direction and position information, which causes meaningless artifacts. Although FFC can extend the feature learning to the frequency domain, it is insensitive to distinguish masked or unmasked regions. Therefore, we propose to use position encoding in masked regions called MPE for the image inpainting, which is orthogonal to and improves upon the FFC in FTR.

Specifically, our MPE can be expressed as the masking distance $\mathbf{E}_{dis}$ and the masking directions $\mathbf{E}_{dir}$ as shown in Fig.~\ref{fig:MPE}. 
Given an inversed 256$\times$256 binary mask, where one indicates unmasked regions and zero indicates masked regions, we use a 3$\times$3 all-one kernel to calculate the masking distance $\mathbf{D}_{dis}$ for each position in masked regions as shown in Fig.~\ref{fig:MPE}(b). Then, the distance is clipped and mapped by the Sinusoidal Positional Encoding (SPE)~\cite{vaswani2017attention} to get $\mathbf{E}_{dis}\in\mathbb{R}^{256\times256\times d}$
\begin{equation}
\begin{split}
\mathbf{E}_{dis,2i}&=\mathrm{sin}(\mathrm{clip}(\mathbf{D}_{dis},0,D_{max})/10000^{\frac{i}{d}}),\\
\mathbf{E}_{dis,2i+1}&=\mathrm{cos}(\mathrm{clip}(\mathbf{D}_{dis},0,D_{max})/10000^{\frac{i}{d}}),
\end{split}
\label{eq:masking_distance}
\end{equation}
where $i$ indicates the channel index; $D_{max}=128$, and $d=64$ means the total channels of $\mathbf{E}_{dis}$, which is the same as the first convolution of FTR. Since SPE can only provide absolute positional information~\cite{xu2020positional}, $\mathbf{E}_{dis}$ can be further resized by the nearest interpolation to various scales during the training for learning relative positional information in arbitrary resolutions.
For masking directions, we use 4 different binary kernels to get the 4-channel one-hot vector $\mathbf{D}_{dir}\in\mathbb{R}^{256\times256\times4}$. Values of $\mathbf{D}_{dir}$ depend on which kernel covers the masked regions firstly. $\mathbf{D}_{dir}$ shows the nearest direction from a masked position to an unmasked one as shown in Fig.~\ref{fig:MPE}(c). 
Note that the masking direction is a multi-label vector, because a pixel may have more than one shortest direction. Then $\mathbf{D}_{dir}$ is projected to a $d$ dimension features with learnable embedding parameters $\mathbf{W}_{dir}\in\mathbb{R}^{4\times d}$ as
\begin{equation}
\mathbf{E}_{dir}=\mathbf{D}_{dir}\times\mathbf{W}_{dir}\in\mathbb{R}^{256\times256\times d}.
\label{eq:masking_direction}
\end{equation}
$\mathbf{E}_{dis}$ and $\mathbf{E}_{dir}$ are added as MPE to the first layer of FTR.

\subsubsection{Zero-initialized Residual Addition (ZeroRA)}
Since most inpainting methods are based on sophisticated GANs nowadays, training the inpainting model incrementally is non-trivial. However, benefiting from various auxiliary information~\cite{nazeri2019edgeconnect,cao2021learning,liao2020guidance}, incrementally training is flexible to improve the image inpainting. 
To improve the pre-trained inpainting model (\emph{i.e.}, FTR in ZITS++) incrementally with holistic structures, we propose to use ZeroRA, which has been leveraged in~\cite{bachlechner2020rezero} to replace the layer normalization in the transformer. Compared to training from scratch with auxiliary information, ZeroRA can save 11 days' computations on $256\times 256$ resolution Places2 with 3 V100 16 GB GPUs. The idea of ZeroRA is simple. For the given input feature $x$, the output feature $x'$ is got from adding a skip connection with function $F$ to $x$ with a zero-initialized trainable residual weight $\alpha$ as
\begin{equation}
x'=x+\alpha\cdot F(x).
\label{eq:rezero}
\end{equation}
For simple linear-based models, if $\alpha$ is initialized in zero, the input-output Jacobian will be initialized to 1, which makes the training stable. For more complex cases, experiments in~\cite{bachlechner2020rezero} also prove the effectiveness of ZeroRA. Since ZeroRA can replace the layer normalization in the transformer, it can also improve the expressive power of the model without degrading variances to early layers.

In our case, we use ZeroRA to incrementally add structural information from SFE to FTR. Specifically, 4 zero-initialized $\{\alpha_k\}_{k=0}^3$ are utilized to fuse 4 related feature maps $\{\mathbf{S}_k\}_{k=0}^3$ from SFE. For the feature $\mathbf{X}_k$ of FTR encoder layer $k$, which is based on Conv-BatchNorm-ReLU, we add residuals as follows
\begin{equation}
\begin{split}
\mathbf{X}_{k+1}&=\mathrm{Conv}(\mathbf{X}_k+\alpha_k\cdot\mathbf{S}_k),\\
\mathbf{X}_{k+1}&=\mathrm{ReLU}(\mathrm{BatchNorm}(\mathbf{X}_{k+1})).
\end{split}
\label{eq:rezero_ftr}
\end{equation}
Another benefit of ZeroRA-based incremental learning is that it preserves the equivalent output of the pre-trained model at the start of finetuning. 
This helps to stabilize the training and adaptively transfer the necessary information.
% There is another advantage of ZeroRA-based incremental learning. The model output is equivalent to the pre-trained one at the beginning of finetuning, which can effectively stabilize the training, and transfer necessary information adaptively.
Our ablation studies show that the ZeroRA is important to incrementally finetune the pre-trained inpainting model with additional information. 
\textcolor{black}{We could also apply ZeroRA to other locations except before the convolution, while the equivalence at the beginning of finetuning is unchanged. Empirically, we find that ZeroRA before the convolution is stable enough in our experiments.}

\subsection{Training Pipeline and Loss Functions}
\label{sec:loss}

\textcolor{black}{We first pre-train FTR as a regular GAN-based inpainting model. Meanwhile, the TSR is trained separately for the prior reconstruction. Note that L-Edges and lines are jointly learned by a single TSR because they enjoy complementary information for each other. Then we finetune FTR with priors recovered from TSR through SFE. ZeroRA is utilized to facilitate the finetuning. More details are in Sec.~\ref{sec:imp_details}.}

We adopt the same loss functions as~\cite{suvorov2021resolution}, which include L1 loss, adversarial loss, feature match loss, and High Receptive Field  (HRF) perceptual loss~\cite{suvorov2021resolution}.
First, L1 loss is only calculated within the unmasked regions as
\begin{equation}
\mathcal{L}_{L1}=(1-\mathbf{M})\odot|\mathbf{\hat{I}}-\mathbf{\tilde{I}}|_1,
\label{eq:l1_loss}
\end{equation}
where $\mathbf{M}$ indicates 0-1 mask that 1 means masked regions; $\odot$ means the element-wise multiplication; $\mathbf{\hat{I}},\mathbf{\tilde{I}}$ indicate the ground truth and predicted images respectively. The adversarial loss consists of the discriminator loss $\mathcal{L}_D$ and the generator loss $\mathcal{L}_G$. Moreover, we only regard features from masked regions as fake samples in $\mathcal{L}_D$. The PatchGAN~\cite{isola2017image} based discriminator is written as $D$ and the combination of FTR and SFE can be seen as the generator $G$, Then the adversarial loss can be indicated as
\begin{equation}
\begin{split}
\negthickspace\negthickspace
\mathcal{L}_{D}=&-\mathbb{E}_{\mathbf{\hat{I}}}\left[\mathrm{log}D(\mathbf{\hat{I}})\right]-\mathbb{E}_{\mathbf{\tilde{I}},\mathbf{M}}\left[\mathrm{log}D(\mathbf{\tilde{I}})\odot(1-\mathbf{M})\right]\\
&-\mathbb{E}_{\mathbf{\tilde{I}},\mathbf{M}}\left[\mathrm{log}(1-D(\mathbf{\tilde{I}}))\odot\mathbf{M}\right],\\
&\mathcal{L}_{G}=-\mathbb{E}_{\mathbf{\tilde{I}}}\left[\mathrm{log}D(\mathbf{\tilde{I}})\right],\\
&\mathcal{L}_{adv}=\mathcal{L}_D+\mathcal{L}_G+\lambda_{GP}\mathcal{L}_{GP},
\end{split}
\label{eq:adv_loss}
\end{equation}
where $\mathcal{L}_{GP}=\mathbb{E}_{\mathbf{\hat{I}}}||\triangledown_{\mathbf{\hat{I}}}D(\mathbf{\hat{I}})||^2$ is the gradient penalty~\cite{gulrajani2017improved} and $\lambda_{GP}=1e-3$. We also use the feature match loss~\cite{wang2018high} $\mathcal{L}_{fm}$, which is based on L1 loss between discriminator features of true and fake samples. $\mathcal{L}_{fm}$ is usually used to stable the GAN training. It can also slightly improve performance. 
Furthermore, we use the HRF loss $\mathcal{L}_{hrf}$ in~\cite{suvorov2021resolution} as
\begin{equation}
\mathcal{L}_{hrf}=\mathbb{E}(\left[\phi_{hrf}(\mathbf{\hat{I}})-\phi_{hrf}(\mathbf{\tilde{I}})\right]^2),
\label{eq:hrfpl}
\end{equation}
where $\phi_{hrf}$ indicates a pre-trained segmentation ResNet50 with dilated convolutions.
As discussed in~\cite{suvorov2021resolution}, using HRF loss instead of the perceptual loss can improve the quality of the inpainting model. The final loss of our model in the incremental training can be written as
\begin{equation}
\small
\begin{split}
\negthickspace
\mathcal{L}_{final}=&\lambda_{L1}\mathcal{L}_{L1}+\lambda_{adv}\mathcal{L}_{adv}+\lambda_{fm}\mathcal{L}_{fm}+\lambda_{hrf}\mathcal{L}_{hrf},
\end{split}
\label{eq:final_loss}
\end{equation}
where $\lambda_{L1}=10,\lambda_{adv}=10,\lambda_{fm}=100,\lambda_{hrf}=30$.

\subsection{\textcolor{black}{Various Priors for Inpainting}}
\label{sec:priors}

\begin{figure*}
\begin{centering}
\includegraphics[width=0.95\linewidth]{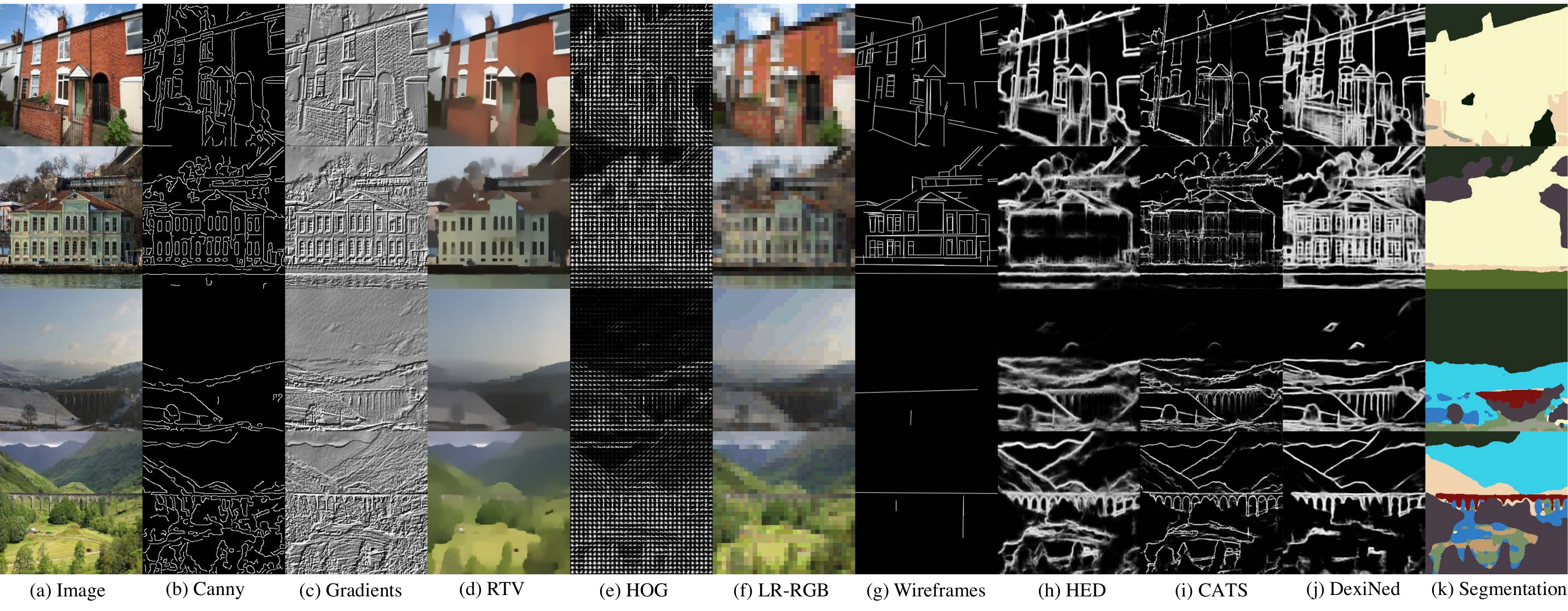}
\par\end{centering}
\vspace{-0.1in}
 \caption{\textcolor{black}{Visualizations of different priors discussed in this paper. We combine gradients in both x and y directions for a consistent representation. \label{fig:qualitative_priors}}}
\vspace{-0.15in}
\end{figure*}

\textcolor{black}{In this section, we introduce various priors, which are roughly categorized as classical priors, and learning-based priors. The classical priors have canny, gradients, RTV, HOG, and LR-RGB, while the latter one will consider the wireframes, L-Edges, and semantic segmentation. Learning-based priors employ deep models to extract priors from images.  
These priors will be negatively affected by masked images~\cite{liao2020guidance}, as no visual content is available. Visualizations of these priors are in Fig.~\ref{fig:qualitative_priors}.
In this subsection, we still use the Place2 subset of 25,000 training and 500 validation images over 5 scenes for the pilot study.}

\subsubsection{Classical Priors}

\begin{figure}
\begin{centering}
\includegraphics[width=0.95\linewidth]{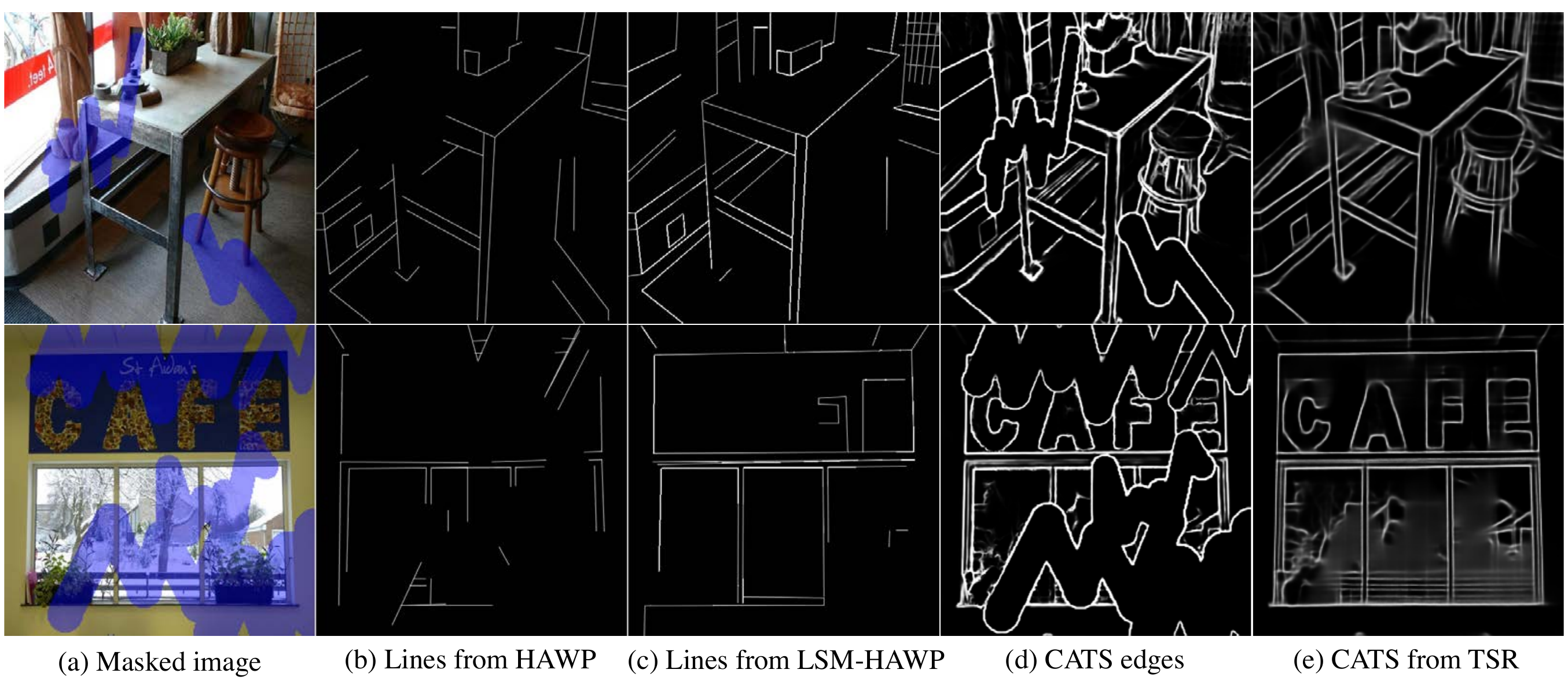}
\par\end{centering}
\vspace{-0.1in}
 \caption{\textcolor{black}{Learning-based priors extracted from masked images. From left to right, (a) masked images, (b) lines from HAWP~\cite{xue2020holistically}, (c) lines from masking augmented LSM-HAWP~\cite{cao2021learning}, (d) masked CATS edges from~\cite{xie2015holistically}, (e) predicted CATS edges from our TSR.\label{fig:masked_priors}}}
\vspace{-0.15in}
\end{figure}

\noindent\textcolor{black}{\textbf{Canny Edges.} 
% The canny edge detector~\cite{canny1986computational} has been widely used in image inpainting~\cite{nazeri2019edgeconnect,cao2021learning,guo2021image} with clear boundaries.
Following~\cite{nazeri2019edgeconnect}, we set $\sigma=2$ as the Gaussian blur coefficient for the $256\times256$ canny detection.
% As a comprehensive edge detector, canny can get binary edge maps with clear boundaries. But canny is only strengthened by low-level image features, which loses texture-less structures and gets excessive high-frequency details as shown in Fig.~\ref{fig:canny_cats_compare}. 
Canny edge priors ${\mathbf{\tilde{P}}_{ce}}$ are optimized with  Binary Cross-Entropy (BCE) loss in TSR. We use the Mask-Predict~\cite{ghazvininejad2019mask} trick and sigmoid temperature parameters for better canny edge predictions as in~\cite{dong2022incremental}.
}

\noindent\textcolor{black}{\textbf{Gradients.} 
% Compared with canny edges, the gradient is a more basic image low-level feature, which is usually investigated to describe edges~\cite{canny1986computational}, lines~\cite{von2008lsd}, and even image representation~\cite{dalal2005histograms}. 
The gradient prior is also used as an auxiliary loss for inpainting in~\cite{yang2020learning}.
As in~\cite{yang2020learning}, we use Sobel filters to extract ground truth RGB gradients $\mathbf{\hat{P}}_{g}$ in two directions, \emph{i.e.}, horizontal and vertical gradient maps. 
Gradient priors $\mathbf{\tilde{P}}_{g}$ are optimized by $L_1$ loss with edge regularization as~\cite{yang2020learning}:
\begin{equation}
\mathcal{L}_{grad}=\beta_1\mathbb{E}[||\mathbf{\tilde{P}}_{g}-\mathbf{\hat{P}}_{g}||_1]+\beta_2\mathbb{E}[(g*\mathbf{C})\odot||\mathbf{\tilde{P}}_{g}-\mathbf{\hat{P}}_{g}||_1],
\label{eq:prior_gradient}
\end{equation}
where $*$ and $\odot$ means convolution and element-wise multiplication; $g$ is a $10\times10$ Gaussian filter with standard deviation 1; $\mathbf{C}$ is canny edge; $\beta_1=0.1$; $\beta_2=20$. 
}

\noindent\textcolor{black}{\textbf{Relative Total Variation (RTV).} Xu~\emph{et al.}~\cite{xu2012structure} propose to use RTV maps to decouple structures and textures.
%
% Thus edge-preserved smooth image can be achieved as shown in Fig.~\ref{fig:qualitative_priors}(d), whose high-frequency details are filtered and low-frequency structures are remained.
Such decoupling has been demonstrated useful to image inpainting~\cite{ren2019structureflow,liu2020rethinking}. We follow~\cite{ren2019structureflow} to use Gaussian kernel $\sigma=3$ and smooth degree $\lambda=0.015$ to extract structural features from $256\times256$ images. Then, $L_1$ loss is taken to optimize RTV priors $\mathbf{\tilde{P}}_{rtv}$.
}

\noindent\textcolor{black}{\textbf{Others}. Additionally, we also compare the priors of HOG~\cite{dalal2005histograms} and Low-Resolution RGB Pixels (LR-RGB), as details are in the supplementary.
% Please refer to the implementation details in our supplementary.
}

\subsubsection{Learning-based Priors}

\noindent\textcolor{black}{\textbf{Lines.} 
The common solution of wireframe parsing is to achieve endpoint heatmaps at first, and then verify valid endpoint pairs through verification modules~\cite{huang2018learning,zhou2019end,xue2020holistically}. However, these wireframe parsers are unable to produce reliable endpoint heatmaps from corrupted images as discussed in~\cite{cao2021learning}. Thus we employ the masking augmented wireframe parser, --LSM-HAWP~\cite{cao2021learning} to extract lines from masked images.
Specifically, we use anti-aliased lines to link endpoint pairs to achieve the ground truth line priors $\mathbf{\hat{P}}_l$.
For simplicity, we directly mask ground truth priors $\mathbf{\hat{P}}_l$ as masked ones $\mathbf{P}_l$ for training. 
During the inference, LSM-HAWP can be well generalized to masked images as illustrated in Fig.~\ref{fig:masked_priors}. The reconstructed lines $\mathbf{\tilde{P}}_l$ are also in grayscale, so we can optimize them with BCE.
}

\noindent\textcolor{black}{\textbf{L-Edges.}
L-Edges such as HED~\cite{xie2015holistically}, CATS~\cite{huan2021unmixing}, and DexiNed~\cite{poma2021dense} enjoy good properties to decouple structures and textures. 
% But there are various learning-based edge detectors with different characteristics. 
% In this paper, we have tried HED~\cite{xie2015holistically}, CATS~\cite{huan2021unmixing}, and DexiNed~\cite{poma2021dense}
So it is interesting to understand which type of L-Edges is in good favor of the inpainting task. 
As in Fig.~\ref{fig:qualitative_priors}(h)-(j), we give some observations:  HED edges are more indecisive near boundaries; CATS edges can get clearer results, but lose some continuous information; DexiNed edges can extract very dense  L-Edges greatly benefited by learning from fine-grained annotations. 
However, these L-Edges also suffer from corrupted images, as the example of CATS edges in Fig.~\ref{fig:masked_priors}(d).  
To this end, we make TSR work for both edge detection and recovery trained with BCE on masked regions. This strategy can be well generalized to masked images without information leakage as in Fig.~\ref{fig:masked_priors}(e).
And, only one forward pass is needed to achieve binary L-Edge priors $\mathbf{\tilde{P}}_{le}$ from TSR.
% Fortunately, the influence of masked regions is limited to local areas near the masking boundary as shown in Fig.~\ref{fig:masked_priors}. So we can simply dilate masking regions with $10\times10$ to eliminate these artifacts. 
% We also use BCE to optimize the binary L-Edge priors $\mathbf{\tilde{P}}_{le}$.
}

\noindent\textcolor{black}{\textbf{Semantic Segmentation.}
Priors from semantic segmentations show good performance in inpainting~\cite{song2018spg,liao2020guidance}. But such priors are largely influenced by corrupted images~\cite{liao2020guidance}. Here we leverage an off-the-shelf ADE20K~\cite{DBLP:journals/corr/ZhouZPFBT16} pre-trained segmentation SETR model~\cite{zheng2021rethinking} to extract ground truth segmentation priors $\mathbf{\hat{P}}_{s}$ from unmasked images. The masked priors $\mathbf{P}_{s}$ are directly masked from $\mathbf{\hat{P}}_{s}$. Particularly, note that such a prior setting can only serve as the performance upper bound of segmentation priors, in order to help understand how much segmentation can help inpainting.
% Different from edge detectors~\cite{xie2015holistically,huan2021unmixing,poma2021dense}, segmentation models are vulnerable to masked inputs, because they need global understanding to get correct semantic predictions.
% because we cannot ensure that real-world inpainting cases contain complete input images without any masks. 
% We select SETR~\cite{zheng2021rethinking}, one of the state-of-the-art methods trained on ADE20K~\cite{DBLP:journals/corr/ZhouZPFBT16} for ground truth $\mathbf{\hat{P}}_{s}$. 
The number of SETR categories is reduced from 150 to 20 with a threshold of 0.0115 for the label frequency in the Places2 subset (Sec.~\ref{sec:ablation_priors}). The rest 130 categories are merged into another new label (21st category).
The CE is used to optimize the generated $\mathbf{\tilde{P}}_{s}$.
}

\subsubsection{Discussion of Various Priors}
\label{sec:dis_priors}

\textcolor{black}{In this work, we utilize the  L-Edge CATS, wireframe lines, and gradients as our priors for the inpainting on the 256$\times$256 images. Gradients are not used for the inpainting at higher resolution, as we empirically compare and discuss various priors in Sec.~\ref{sec:ablation_priors}.}

\noindent\textcolor{black}{\textbf{Why L-Edges outperform other priors?}
In general, L-Edges produce more significant \emph{boundaries} rather than trivial edges from textures (canny) as visualized in  Fig.~\ref{fig:canny_cats_compare}. Such structural boundaries of L-Edges reflect the scene-level understanding of the image.
Here we adopt CATS from all L-Edges since CATS outperforms HED and DexiNed for inpainting. Particularly,  
HED and DexiNed give blurry edges over the image regions of boundaries and high-frequency details respectively as
illustrated in Fig.~\ref{fig:all_priors}.
Furthermore,
CATS can be well generalized to HR cases with the E-NMS and upsampling of SSU in Sec.~\ref{sec:SSU}.
In contrast to other priors, such as gradients and RTV, SSU cannot tackle the upsampling of them, which leads to degradation and artifacts for unseen scales (1024) during the training as in Tab.~\ref{table:HR_priors} and Fig.~\ref{fig:HR_prior_qua}. 
For the consistent LR inpainting (256$\times$256), our model further includes the gradient that outperforms all other priors.
}

\noindent\textcolor{black}{\textbf{Effect of Wireframe Lines.}
From Tab.~\ref{table:multi_prior_compare}, the wireframe line works compatibly with CATS and gradients, and prominently improves FID and LPIPS for both image inpainting at $256\times256$ and $512\times512$ resolution. Such phenomenon indicates that lines can provide simple but strong structural priors to overcome artifacts during the generation as in Fig.~\ref{fig:all_priors}(b), especially for large masking regions and HR inpainting. Moreover, lines enjoy great advantages for man-made scenes as studied in~\cite{cao2021learning}.
In practice, by using the same TSR, lines can be jointly trained and inference with CATS at negligible additional cost.
}

\section{Experiments}
\label{sec:exp}

\noindent \textbf{Datasets}.
Our model is trained on two datasets: Places2~\cite{zhou2017places} and our custom indoor dataset (Indoor). For Places2, we use about 1,800k images from various scenes as the training set, and 36,500 images as the validation. To better demonstrate the structural recovery, we collect 5,000 images from ShanghaiTech~\cite{huang2018learning} and 15,055 images from NYUDepthV2~\cite{Silberman:ECCV12} to build the custom 20,055 Indoor training dataset. For the Indoor validation, we collect 1,000 images which consist of 462 and 538 images from ShanhaiTech and NYUDepthV2 respectively. Places2 and Indoor can all be tested in both 256$\times$256 and 512$\times$512.

\noindent \textcolor{black}{\textbf{High-Resolution Testing Dataset}.
To further explore the performance of HR inpainting, we also test the inpainting ability on both MatterPort3D~\cite{chang2017matterport3d} and our released HR image dataset.
MatterPort3D~\cite{chang2017matterport3d} comprises 1,965 indoor images in 1280$\times$1024, which is used to evaluate the HR inpainting ability in man-made scenes. We resized them into 1024$\times$1024.
\textcolor{black}{For more comprehensive comparisons in challenging HR scenes, we introduce HR-Flickr as below.}
}

\begin{figure}
\begin{centering}
\includegraphics[width=0.9\linewidth]{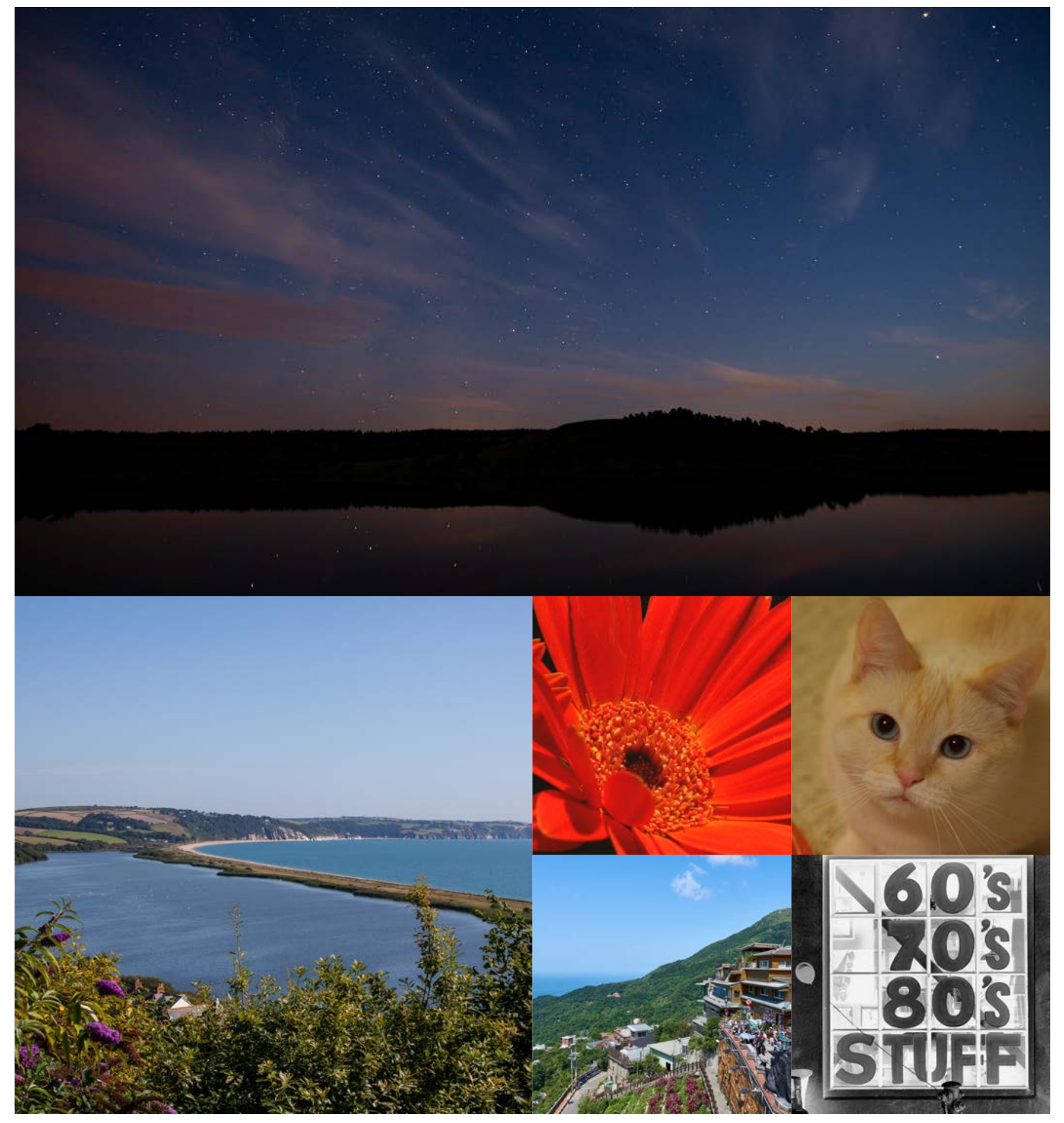}
\par\end{centering}
\vspace{-0.1in}
\caption{High-quality images from HR-Flickr with 1k, 2k, and 4k resolutions.\label{fig:hr-flickr-images}}
\vspace{-0.15in}
\end{figure}

\textcolor{black}{\subsection{HR-Flickr Dataset}}
For the HR cases with more diversity, we newly release a group of high-quality images collected from Flickr with downloading permissions from owners, called HR-Flickr. 
HR-Flickr consists of 500/500/490 HR images with resolutions 1k, 2k, and 4k respectively as shown in Fig.~\ref{fig:hr-flickr-images}. Images from each group of specific resolution have maximum side lengths which are equal to the corresponding group resolution (1024, 2048, 4096).
Moreover, images with persons are eliminated by segmentation methods and manual checking from HR-Flickr to reduce the privacy risk as far as possible. Compared with other HR datasets, such as DIV2K and Flickr2K~\cite{Lim_2017_CVPR_Workshops}, our dataset contains 4k images \textcolor{black}{from Single-Lens Reflex cameras (SLR), which enjoy high quality without any interpolated zoom-in.}  Furthermore, HR-Flickr has more diverse and challenging cases for image inpainting, while we manually filter images with too simple textures or monotonous scenes. \textcolor{black}{Moreover, we carefully remove images of personal information to protect privacy.}

\subsection{Implementation Details}
\label{sec:imp_details}

\noindent \textbf{Training Settings.}
Our ZITS++ is implemented with PyTorch. For the training of TSR, we use the Adam optimizer of learning rate 6e-4 with 1,000 steps warmup and cosine decay. TSR is trained with 150k and 400k steps for Indoor and Places2. On the other hand, we first train the FTR with Adam optimizer of learning rates 1e-3 and 1e-4 for the generator and discriminator respectively. And FTR is trained with 100k steps on Indoor and 800k steps on Places2. Then, we incrementally finetune them with ZeroRA for 50k and 150k steps on Indoor and Places2 respectively, and reduce the generator learning rate to 3e-4. Besides, we warmup the learning rate for training the SFE with 2,000 steps.
For the training of TSR and FTR, input images are resized into 256$\times$256. For the incremental finetuning, we separately train two versions of ZITS++, which are the version trained in 256$\times$256 and the version trained in random size from 256 to 512. The second model can handle some situations with higher-resolution inputs. And the MPE is also changed to a relative position encoding for the random size training.

\noindent  \textbf{Mask Settings.}
To tackle the real-world object removal task, we follow the mask setting from~\cite{cao2021learning}, which includes irregular masking brushes and segmentation masks with masking rates from 10\% to 50\%. Different from~\cite{cao2021learning}, we \textcolor{black}{additionally use some LaMa irregular masks~\cite{suvorov2021resolution} (both thin and thick types) for better diversity, and} randomly combine irregular and segmentation masks with 20\% to improve the learning difficulty. 
%Masks are divided into 4 different rates to evaluate the inpainting generalization.

\noindent\textbf{Competitors.}
% \label{sec:compare_methods}
Our conference version~\cite{dong2022incremental} uses canny edges and lines is denoted as  ZITS~\cite{dong2022incremental}, while ZITS++ indicates our improved version in this paper enhanced by the priors of CATS, gradients, and lines, as well as several novel modules and techniques.
We compare the proposed model with other state-of-the-art methods, which include Edge Connect (EC)~\cite{nazeri2019edgeconnect}, Contextual Residual Aggregation (HiFill)~\cite{yi2020contextual}, Multi-scale Sketch Tensor inpainting (MST)~\cite{cao2021learning}, Co-Modulation GAN (Co-Mod)~\cite{zhao2021large}, \textcolor{black}{Mask-Aware Transformer (MAT)}~\cite{li2022mat}, and Large Mask inpainting (LaMa)~\cite{suvorov2021resolution}. All competitors are compared in Places2. We also retrain EC, MST, and LaMa for the Indoor dataset to discuss the structure recovery. Note that the LaMas compared below are all trained with the same total steps as ZITS++.

\subsection{Quantitative Comparisons}

\begin{table} \small
\caption{Quantitative results on Indoor and Places2 in 256$\times$256. \textcolor{black}{The best results are in bold, while the second-best ones are underlined.}\label{table:main_results}}
\vspace{-0.1in}
\begin{centering}
\renewcommand\tabcolsep{1.5pt}
\footnotesize
\begin{tabular}{c|cccc|cccc}
\toprule 
& \multicolumn{4}{c|}{Indoor} & \multicolumn{4}{c}{Places2}\tabularnewline
 & PSNR$\uparrow$ & SSIM$\uparrow$ & FID$\text{\ensuremath{\downarrow}}$ & LPIPS$\text{\ensuremath{\downarrow}}$ & PSNR$\uparrow$ & SSIM$\uparrow$ & FID$\text{\ensuremath{\downarrow}}$ & LPIPS$\text{\ensuremath{\downarrow}}$\tabularnewline
\midrule
EC~\cite{nazeri2019edgeconnect} & 24.07 & 0.884 & 22.02 & 0.135 & 23.31 & 0.839 & 6.21 & 0.149\tabularnewline
MST~\cite{cao2021learning} & 24.52 & 0.894 & 21.65 & 0.122 & 24.02 & 0.862 & 3.53 & 0.137\tabularnewline
HiFill~\cite{yi2020contextual} & - & - & - & - & 20.76 & 0.770 & 21.33 & 0.246\tabularnewline
Co-Mod~\cite{zhao2021large} & - & - & - & - & 22.57 & 0.843 & 1.49 & 0.122\tabularnewline
LaMa~\cite{suvorov2021resolution} & 25.20 & 0.902 & 16.97 & 0.112 & 24.37 & 0.869 & 1.63 & 0.155\tabularnewline
\textcolor{black}{ZITS~\cite{dong2022incremental}} & \underline{25.57} & \underline{0.907} & \underline{15.93} & \underline{0.098} & \underline{24.42} & \underline{0.870} & \underline{1.47} & \underline{0.108}\tabularnewline
\textcolor{black}{ZITS++(ours)} & \textbf{26.15} & \textbf{0.917} & \textbf{14.61} & \textbf{0.090} & \textbf{25.13} & \textbf{0.884} & \textbf{1.19} & \textbf{0.094}\tabularnewline
\bottomrule 
\end{tabular}
\par\end{centering}
\end{table}

\begin{table} \small
\caption{\textcolor{black}{P-IDS and U-IDS on Indoor and Places2 in 256$\times$256.} The best results are in bold, while the second best ones are underlined.\label{table:ids_results}}
\vspace{-0.1in}
\begin{centering}
\renewcommand\tabcolsep{1.5pt}
\footnotesize
\begin{tabular}{cc|cccccc|cc}
\toprule
 & (\%) & EC & MST & HiFill & LaMa & ZITS & ZITS++  & Co-Mod & MAT\tabularnewline
\midrule
\multirow{2}{*}{Indoor} & P-IDS$\uparrow$ & 0.80 & 1.00 & - & 5.00 & \uline{7.90} & \textbf{11.1} & - & -\tabularnewline
 & U-IDS$\uparrow$ & 14.55 & 15.25 & - & 24.30 & \uline{30.55} & \textbf{33.05} & - & -\tabularnewline
\midrule
\multirow{2}{*}{Places2} & P-IDS$\uparrow$ & 3.06 & 10.38 & 1.21 & 20.21 & 22.49 & 23.93 & \textbf{28.65} & \uline{25.42}\tabularnewline
 & U-IDS$\uparrow$ & 23.60 & 31.80 & 14.11 & 38.56 & 39.89 & \uline{40.61} & \textbf{40.96} & 39.15\tabularnewline
\bottomrule
\end{tabular}
\par\end{centering}
\end{table}

\noindent\textbf{Quantitative Inpainting Results.}
In Tab.~\ref{table:main_results}, we utilize PSNR, SSIM~\cite{wang2004image}, FID~\cite{heusel2018gans}, and LPIPS~\cite{zhang2018unreasonable} to assess the performance of all compared methods \textcolor{black}{including our conference version} on the Indoor and Places2 datasets in 256$\times$256 with mixed segmentation and irregular masks. 
More results with different masking rates are shown in the supplementary.
For Indoor, our \textcolor{black}{ZITS++ and ZITS achieve the best and the second best results on all metrics.} MST can get slightly better results compared with EC, which is benefited from the usage of lines. 
LaMa can get more acceptable FID and LPIPS while our ZITS can achieve significant improvements based on LaMa due to the seamlessly embedded structural information and positional encoding.
Note that the gap between ZITS++ and MST is also caused by the different quality of recovered structures as discussed below.
For Places2, HiFill fails to get good results with large masks, which may be caused by its limited model capacity. Note that Co-Mod has a low FID and LPIPS on Places2. However, Co-Mod is trained with a sophisticated StyleGAN~\cite{karras2020analyzing} with much more training data compared with others. And our ZITS can even achieve better results compared with Co-Mod with limited data scale and training steps. Moreover, ZITS++ improved in this paper can even outperform ZITS. In general, our method has superior performance compared with LaMa, which is valuable with limited finetune steps. And LaMa in Tab.~\ref{table:main_results} is trained with the same total steps as ZITS++.

\noindent\textbf{Comparison of P-IDS and U-IDS.}
Following~\cite{zhao2021large}, we also test the feature space separability of generated images with P-IDS and U-IDS in Tab.~\ref{table:ids_results}. To avoid the overfitting of SVM, we use the 768-d features from the pre-trained InceptionV3 model instead of the 2048-d ones on the Indoor dataset with just 1,000 testing images. ZITS++ outperforms all competitors in Indoor. But Co-Mod and MAT achieve superior P-IDS than our method in Places2. We think this phenomenon is reasonable, caused by different training formulations between Co-Mod~\cite{zhao2021large} and LaMa~\cite{suvorov2021resolution}. 
Both Co-Mod and MAT enjoy the co-modulation training based on StyleGAN2~\cite{karras2020analyzing} with latent vectors, which employ little to no reconstruction loss (such as L1 and perceptual loss). Co-modulation based methods enjoy fewer blur and texture artifacts, but they often suffer from hallucinated generations and fail to result in faithful inpainted images. Contrarily, ZITS is based on the training strategy of LaMa, which has relatively high weights for reconstruction losses as in Sec.~\ref{sec:loss}, and enjoys high-fidelity generations as verified in qualitative comparisons (Sec.~\ref{sec:qualitative}).
Hence it is hard to judge whether the inpainting results are faithful through the feature separability in P-IDS and U-IDS.
But we think our prior guidance is orthogonal to the co-modulation GAN training; unifying them could be seen as interesting future work.

\begin{table} 
\small
\caption{Quantitative Precision (P.), Recall (R.) and F1-score (F1) of \textcolor{black}{L-Edges (CATS~\cite{huan2021unmixing})} and lines on Indoor and Places2. \label{table:edgeline_res}}
\vspace{-0.1in}
\begin{centering}
\renewcommand\tabcolsep{2.8pt}
\small
\begin{tabular}{c|c|ccc|ccc|c}
\toprule 
\multicolumn{1}{c}{} &  & \multicolumn{3}{c|}{L-Edge} & \multicolumn{3}{c|}{Line} & Avg\tabularnewline
\multicolumn{1}{c}{} &  & P. & R. & F1 & P. & R. & F1 & F1\tabularnewline
\midrule 
\multirow{2}{*}{Indoor} & MST & 34.97 & 24.15 & 28.55 & 30.64 & 25.75 & 27.85 & 28.20\tabularnewline
 & Ours & \textbf{58.72} & \textbf{48.54} & \textbf{52.07} & \textbf{61.10} & \textbf{66.42} & \textbf{62.66} & \textbf{57.36}\tabularnewline
\midrule 
\multirow{2}{*}{Places2} & MST & 28.56 & 19.27 & 23.00 & 32.00 & 18.36 & 23.23 & 23.12\tabularnewline
 & Ours & \textbf{42.05} & \textbf{53.49} & \textbf{45.57} & \textbf{45.11} & \textbf{59.33} & \textbf{49.75} & \textbf{47.66}\tabularnewline
\bottomrule
\end{tabular}
\par\end{centering}
\end{table}

\noindent\textbf{Quantitative Results of \textcolor{black}{L-Edges} and Lines.}
We show quantitative results of \textcolor{black}{L-Edges and lines reconstructed} on Indoor and Places2 in Tab.~\ref{table:edgeline_res}. Our TSR can get much better results on both Indoor and Places2 compared with MST~\cite{cao2021learning}. It demonstrates that the transformer-based TSR is amenable to learning holistic structures in a sparse tensor space, which can benefit the results of ZITS++ a lot as shown in Tab.~\ref{table:main_results}. 
Moreover, the newly introduced priors could be well recovered by TSR without the Mask-Predict~\cite{ghazvininejad2019mask} trick used in ZITS.
\textcolor{black}{Since both L-Edges and lines are trained jointly in MST and TSR, line metrics may be different from ones produced with the canny edge in our previous work~\cite{dong2022incremental}. But the conclusion is consistent, \textit{i.e.}, the transformer-based TSR outperforms the CNN-based MST in the structure recovery.}

\subsection{Qualitative Comparisons}
\label{sec:qualitative}

\begin{figure}
\begin{centering}
\includegraphics[width=0.85\linewidth]{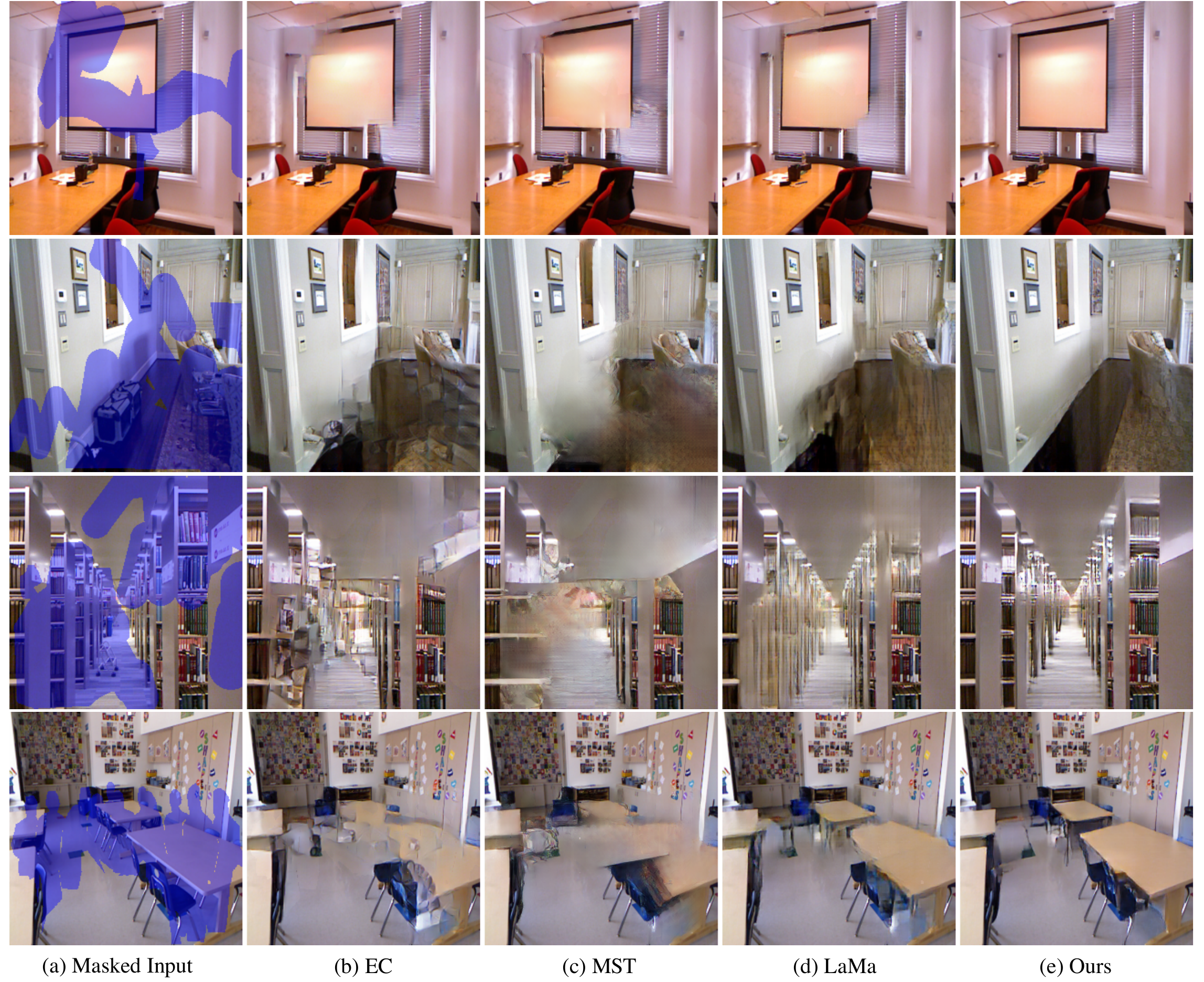}
\par\end{centering}
\vspace{-0.1in}
 \caption{Qualitative 256$\times$256 results of Indoor dataset compared among EC~\cite{nazeri2019edgeconnect}, MST~\cite{cao2021learning}, LaMa~\cite{suvorov2021resolution}, and our ZITS++ results.
 \label{fig:qualitative_indoor}}
 \vspace{-0.15in}
\end{figure}
% Please Zoom-in for details.

\begin{figure*}
\begin{centering}
\includegraphics[width=0.9\linewidth]{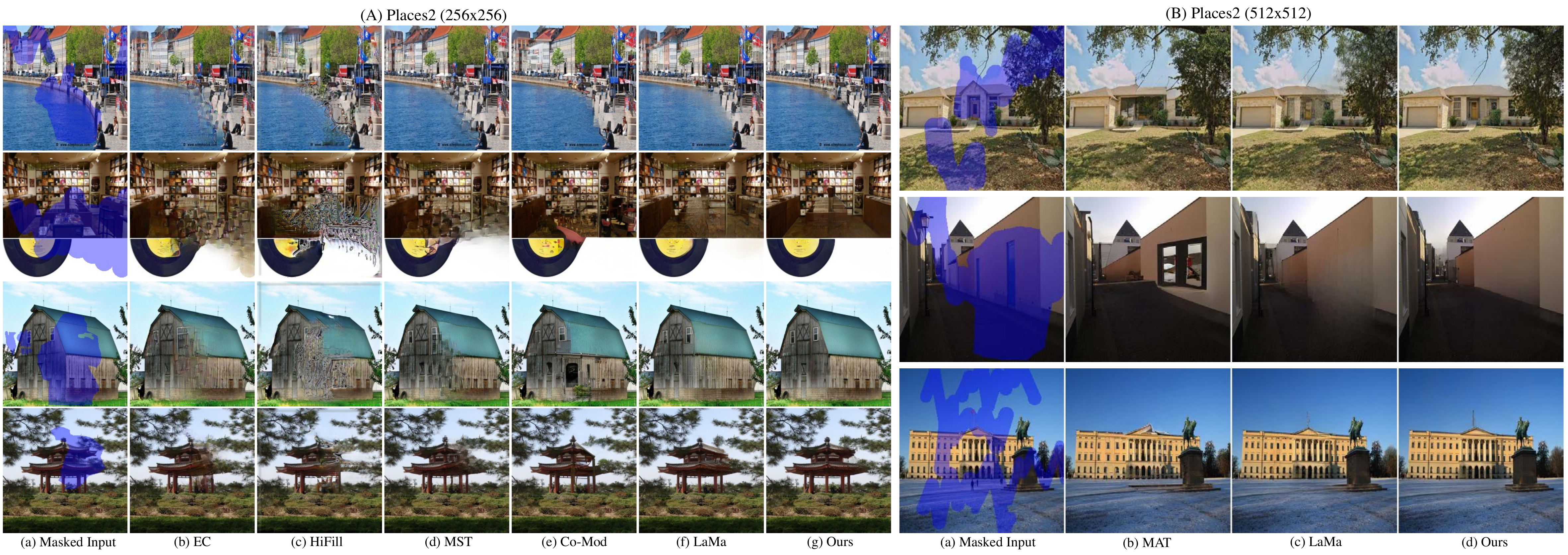}
\par\end{centering}
\vspace{-0.1in}
 \caption{(A) Places2 256$\times$256 results compared with EC~\cite{nazeri2019edgeconnect}, HiFill~\cite{yi2020contextual}, MST~\cite{cao2021learning}, Co-Mod~\cite{zhao2021large}, LaMa~\cite{suvorov2021resolution}, and our ZITS++. 
 \textcolor{black}{(B) Places2 512$\times$512 results compared with MAT~\cite{li2022mat}, LaMa~\cite{suvorov2021resolution}, and our ZITS++.}
 \label{fig:qualitative_places2}}
 \vspace{-0.15in}
\end{figure*}

% \subsubsection{Qualitative Inpainting Results}

\noindent \textbf{Qualitative Inpainting Results.}
We show the qualitative inpainting results of Indoor in Fig.~\ref{fig:qualitative_indoor} and Places2 in Fig.~\ref{fig:qualitative_places2}. Compared with other methods, our ZITS++ can tackle more reasonable structures, especially our method can obtain clearer borderlines. \textcolor{black}{Furthermore, ZITS++ achieves prominent improvements in the structure recovery compared with LaMa; and outperforms the state-of-the-art MAT~\cite{li2022mat} with much more faithful results.} Note that both LaMa and ZITS++ are trained with the same steps.

\begin{figure}
\begin{centering}
\includegraphics[width=1.0\linewidth]{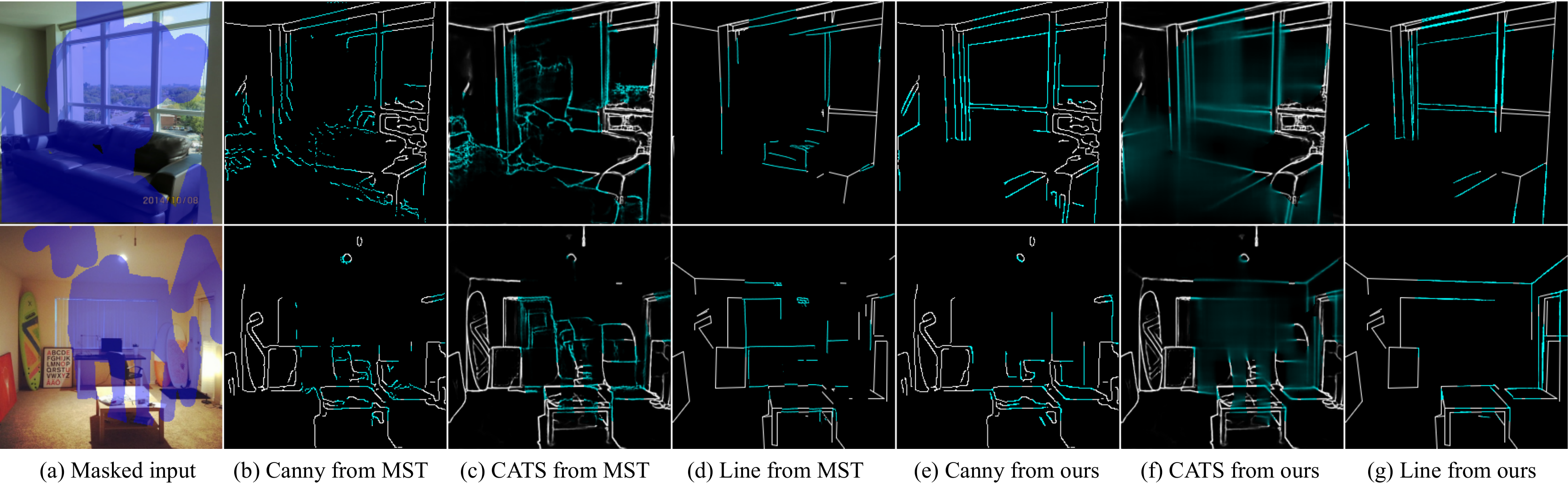}
\par\end{centering}
\vspace{-0.1in}
 \caption{The recovery comparison of canny edges, \textcolor{black}{L-Edges (CATS)} and lines on Indoor dataset (256$\times$256).
 \label{fig:qualitative_str}}
 \vspace{-0.15in}
\end{figure}

% \subsubsection{Qualitative Results of Edges and Lines} 
\noindent \textbf{Qualitative Results of Edges and Lines.}
We compare the structure recovery results in Indoor of Fig.~\ref{fig:qualitative_str} between the proposed transformer-based TSR and CNN-based MST. TSR can achieve more reasonable and expressive results of \textcolor{black}{L-Edges,} canny edges and lines. More qualitative structural results are shown in the supplementary.

\subsection{Ablation Studies}
\label{sec:ablation}
Quantitative ablation studies on Indoor are shown in Tab.~\ref{table:ablation}. MPE and GCs can slightly improve the performance of FTR. Besides, if adding structural information from TSR without ZeroRA, the improvement is limited. So ZeroRA is useful for incremental learning with good convergence. 
Moreover, \textcolor{black}{newly proposed priors, L-Edge and gradient, and the LKA enhanced FTR can further improve the inpainting performance as our full model.}
% the full model achieves the best performance.

\begin{figure}
\begin{centering}
\includegraphics[width=0.7\linewidth]{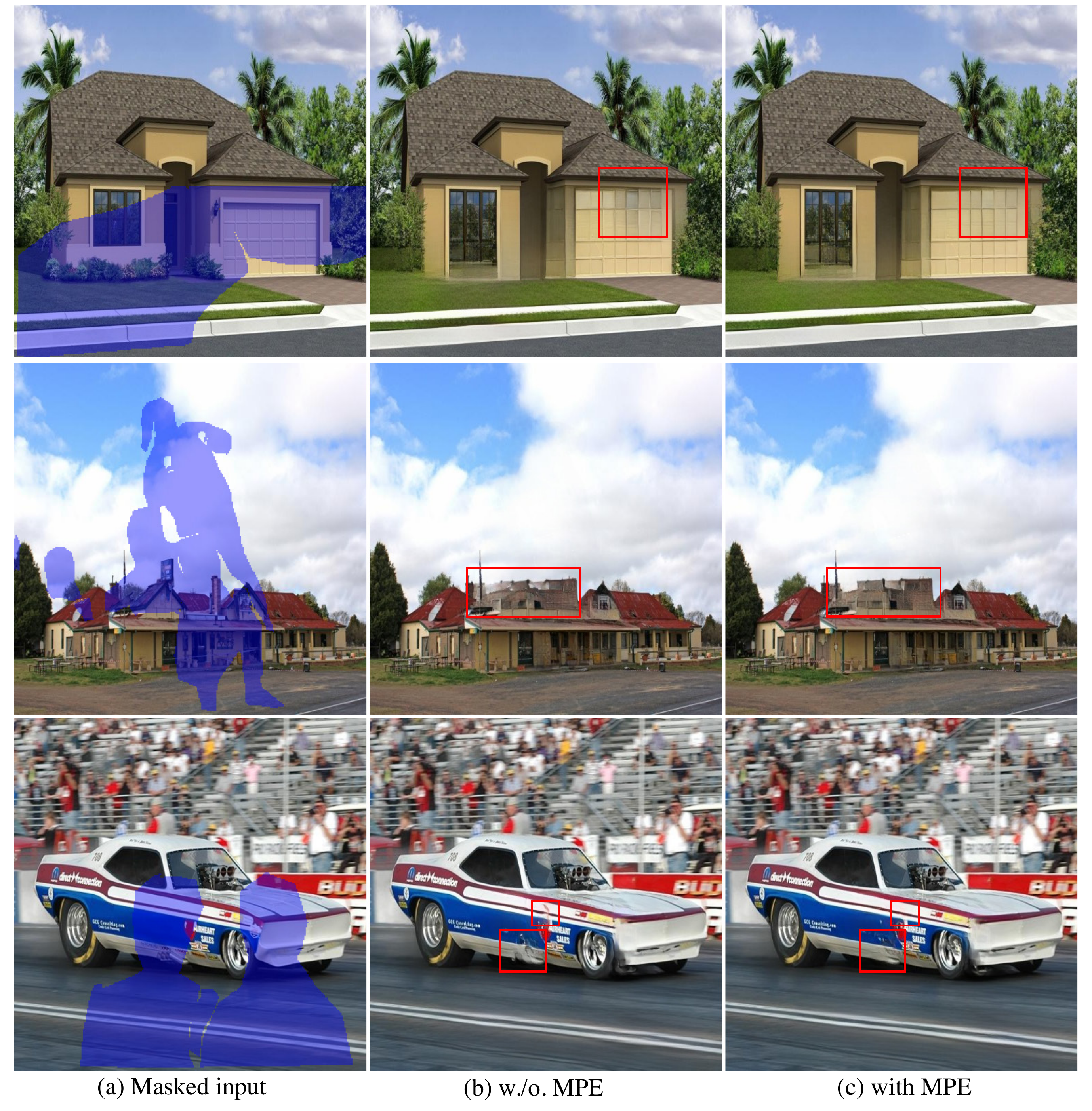}
\par\end{centering}
\vspace{-0.1in}
 \caption{Ablations of 512$\times$512 Places2 with and without MPE.
 \label{fig:MPE_abla}}
 \vspace{-0.15in}
\end{figure}

\begin{table}
\small
\caption{MPE ablation \textcolor{black}{of ZITS and ZITS++ finetuned on 512$\times$512 Places2.}
% finetuned with dynamic resolutions from 256 to 512
\label{table:abla_MPE}}
\vspace{-0.1in}
\centering
\begin{tabular}{cccccc}
\toprule 
 &  & PSNR$\uparrow$  & SSIM$\uparrow$  & FID$\downarrow$  & LPIPS$\downarrow$\tabularnewline
\midrule 
\multirow{2}{*}{ZITS} & with MPE  & \textbf{24.23}  & \textbf{0.881}  & \textbf{26.08}  & \textbf{0.133}\tabularnewline
 & w./o. MPE  & 24.20  & 0.880  & 26.29  & 0.135 \tabularnewline
\midrule
\multirow{2}{*}{ZITS++} & with MPE  & \textbf{24.50} & \textbf{0.885} & \textbf{25.64} & \textbf{0.118}\tabularnewline
 & w./o. MPE  & 24.45 & 0.884 & 25.95 & 0.119\tabularnewline
\bottomrule
\end{tabular}
\end{table}

\begin{table}
\footnotesize
\caption{Ablation studies with different settings on Indoor. \textcolor{black}{`NewPriors' indicate L-Edge and gradient used in ZITS++ rather than the canny edge of ZITS, while `LKA' means improved FTR as in Sec.~\ref{sec:FTR}.} \label{table:ablation}}
\vspace{-0.1in}
\renewcommand\tabcolsep{0.9pt}
\begin{centering}
\begin{tabular}{ccccccc|cccc}
\toprule 
FTR & SFE & MPE & ReZero & GCs & NewPriors & LKA & PSNR$\uparrow$ & SSIM$\uparrow$ & FID$\downarrow$ & LPIPS$\downarrow$\tabularnewline
\midrule
\CheckmarkBold{} &  &  &  &  &  &  & 25.20 & 0.902 & 16.97 & 0.112\tabularnewline
\CheckmarkBold{} &  & \CheckmarkBold{} &  &  &  &  & 25.31 & 0.903 & 16.44 & 0.110\tabularnewline
\CheckmarkBold{} & \CheckmarkBold{} & \CheckmarkBold{} &  & \CheckmarkBold{} &  &  & 25.28 & 0.905 & 16.15 & 0.102\tabularnewline
\CheckmarkBold{} & \CheckmarkBold{} &  & \CheckmarkBold{} & \CheckmarkBold{} &  &  & 25.46 & 0.906 & 16.22 & 0.107\tabularnewline
\CheckmarkBold{} & \CheckmarkBold{} & \CheckmarkBold{} & \CheckmarkBold{} &  &  &  & 25.51 & 0.906 & 16.15 & 0.103\tabularnewline
\CheckmarkBold{} & \CheckmarkBold{} & \CheckmarkBold{} & \CheckmarkBold{} & \CheckmarkBold{} &  &  & 25.57 & 0.907 & 15.93 & 0.098\tabularnewline
\CheckmarkBold{} & \CheckmarkBold{} & \CheckmarkBold{} & \CheckmarkBold{} & \CheckmarkBold{} & \CheckmarkBold{} &  & 26.14 & 0.916 & 14.66 & \textbf{0.089}\tabularnewline
\CheckmarkBold{} & \CheckmarkBold{} & \CheckmarkBold{} & \CheckmarkBold{} & \CheckmarkBold{} & \CheckmarkBold{} & \CheckmarkBold{} & \textbf{26.15} & \textbf{0.917} & \textbf{14.61} & 0.090\tabularnewline
\bottomrule
\end{tabular}
\par\end{centering}
\end{table}

\noindent \textbf{Effects of MPE.}
We further exploit the effects of MPE in HR inpainting. 
FTR is trained without MPE at first. Then we use the ZeroRA technique to finetune the model with and without MPE of the same steps. Results in Tab.~\ref{table:abla_MPE} show that the simple MPE-based finetuning effectively improves the 512-inpainting in FID. From Fig.~\ref{fig:MPE_abla}, ZITS with MPE generates images with natural and smooth colors.

\begin{figure}
\begin{centering}
\includegraphics[width=0.99\linewidth]{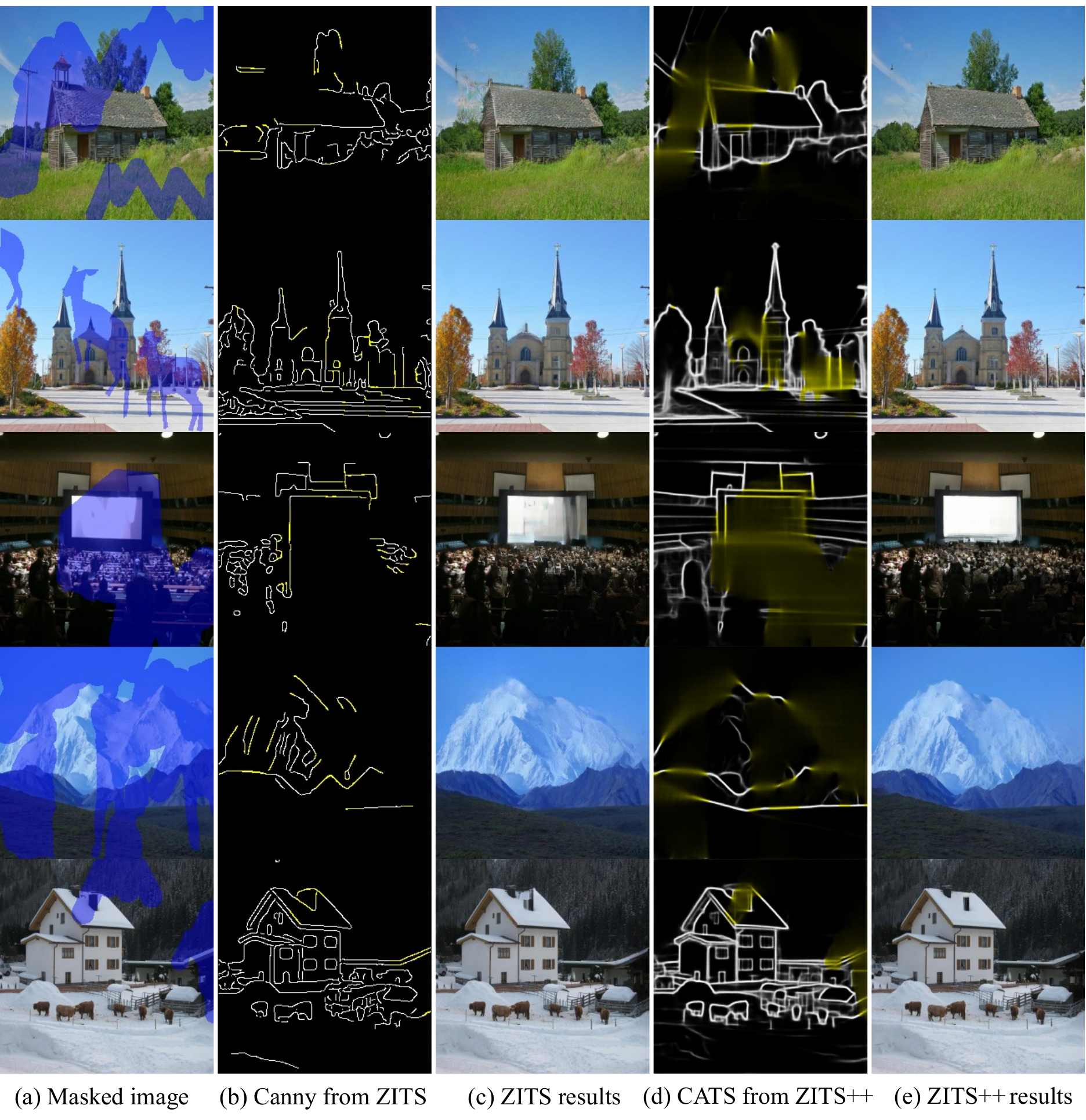}
\par\end{centering}
\vspace{-0.1in}
 \caption{\textcolor{black}{Qualitative comparisons between canny edges from ZITS and CATS edges from ZITS++ in Places2. Yellow edges indicate restored regions in (b)(d).}
 \label{fig:compare_zits_zits+}}
 \vspace{-0.15in}
\end{figure}

\begin{figure*}
\begin{centering}
\includegraphics[width=0.99\linewidth]{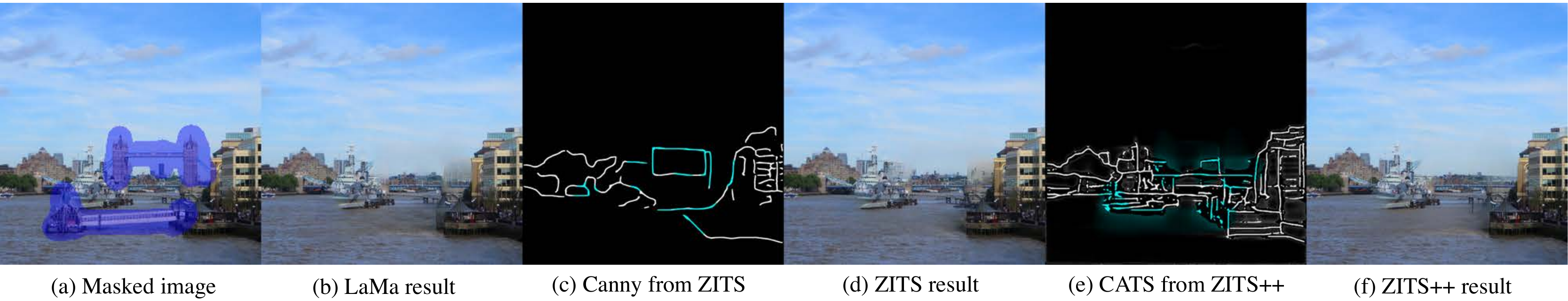}
\par\end{centering}
\vspace{-0.1in}
 \caption{\textcolor{black}{ZITS++ with CATS can also achieve better performance compared with ZITS in 1k inpainting. Please zoom in for details.}
 \label{fig:compare_zits_zits+_1k}}
 \vspace{-0.15in}
\end{figure*}

% \textcolor{black}{\subsubsection{Comparisons between Canny and CATS Edges}}
\noindent\textbf{Comparisons between Canny and CATS Edges.}
Since the edge prior is simultaneously utilized in our conference model ZITS~\cite{dong2022incremental} (canny) and the newly proposed ZITS++ (CATS), we further compare their performance in Fig.~\ref{fig:compare_zits_zits+} and Fig.~\ref{fig:compare_zits_zits+_1k}. From Fig.~\ref{fig:compare_zits_zits+}, CATS can achieve more reasonable structure recovery benefiting from the superior scene understanding from L-Edge. Moreover, CATS results pay more attention to valuable object boundaries rather than textural canny edges filtered by gradients. Such a decoupling benefits both TSR and FTR for their own functions as mentioned in Sec.~\ref{sec:dis_priors}. Besides, from the HR inpainting results of 1024$\times$1024 shown in Fig.~\ref{fig:compare_zits_zits+_1k}, our ZITS++ address a failed HR case in ZITS with a better generation of distant buildings.
Therefore, the CATS enhanced ZITS++ enjoys superior performance for image inpainting.

\subsubsection{Prior Ablations}
\label{sec:ablation_priors}

\noindent\textcolor{black}{\textbf{Settings.}
To expose the effect of different priors for image inpainting, we finetune the whole Places2 pre-trained LaMa baseline with 50,000 steps in ZeroRA on the Places2 subset of  25,000 training and 500 validation images over 5 scenes.
For each prior shown in Tab.~\ref{table:single_prior_compare}, we train a TSR with the specific loss function mentioned in Sec.~\ref{sec:priors}. For combined priors shown in Tab.~\ref{table:multi_prior_compare}, we train one TSR for priors of the same loss function, \emph{e.g.}, CATS and lines are restored with the same TSR optimized with BCE, and gradients are restored with another TSR optimized with regularized $\mathcal{L}_{grad}$ in Eq.(\ref{eq:prior_gradient}).
Other implemented details follow Sec.~\ref{sec:imp_details}.
PSNR, SSIM~\cite{wang2004image}, FID~\cite{heusel2018gans}, and LPIPS~\cite{zhang2018unreasonable} are utilized as metrics.
}

\begin{table} \small
\caption{\textcolor{black}{Quantitative ablations on 256$\times$256 Places2 subset with different priors. `Seg.' indicate semantic segmentations.
The baseline (first row) is also finetuned with the same steps on the subset.
Canny+Line are priors used in our conference version~\cite{dong2022incremental}. 
SSU means whether this prior can be upsampled by SSU to HR.
\label{table:single_prior_compare}}}
\vspace{-0.15in}
\centering
\small
\begin{tabular}{c|c|c|c|c|c}
\toprule 
Priors & SSU & PSNR$\uparrow$ & SSIM$\uparrow$ & FID$\downarrow$ & LPIPS$\downarrow$\tabularnewline
\midrule 
\rowcolor{purple!25}
-- & -- & 25.09 & 0.873 & 23.62 & 0.101\tabularnewline
\rowcolor{blue!25}
Canny & $\text{\ensuremath{\checkmark}}$ & 25.36 & 0.879 & 21.68 & 0.097\tabularnewline
\rowcolor{blue!25}
Canny+Line & $\text{\ensuremath{\checkmark}}$ & 25.35 & 0.878 & 21.56 & 0.094\tabularnewline
\rowcolor{green!25}
HED & $\text{\ensuremath{\checkmark}}$ & 25.25 & 0.877 & 21.81 & 0.097\tabularnewline
\rowcolor{green!25}
CATS & $\text{\ensuremath{\checkmark}}$ & 25.44 & 0.879 & 21.50 & 0.097\tabularnewline
\rowcolor{green!25}
DexiNed & $\text{\ensuremath{\checkmark}}$ & 25.29 & 0.878 & 21.54 & 0.097\tabularnewline
Gradient & $\times$ & \textbf{25.80} & \textbf{0.886} & \textbf{20.98} & \textbf{0.091}\tabularnewline
HOG & $\times$ & 25.35 & 0.878 & 21.97 & 0.097\tabularnewline
LR-RGB & $\times$ & 25.21 & 0.876 & 22.08 & 0.097\tabularnewline
RTV & $\times$ & 25.75 & 0.881 & 21.27 & 0.093\tabularnewline
Seg. & $\times$ & 25.32 & 0.876 & 21.09 & 0.096\tabularnewline
\bottomrule 
\end{tabular}
\vspace{-0.1in}
\end{table}

\begin{table}  \small
\caption{\textcolor{black}{Quantitative ablations on the Places2 subset of 256$\times$256 and 512$\times$512 with different prior combinations. `Grad' indicates the gradient prior.\label{table:multi_prior_compare}}}
\vspace{-0.15in}
\centering
\footnotesize
\renewcommand\tabcolsep{1.0pt}
\begin{tabular}{ccc|cccc|cccc}
\toprule 
\multicolumn{3}{c|}{Priors} & \multicolumn{4}{c|}{256$\times$256} & \multicolumn{4}{c}{512$\times$512}\tabularnewline
\midrule 
{\scriptsize{}CATS} & {\scriptsize{}Line} & {\scriptsize{}Grad} & {\scriptsize{}PSNR$\uparrow$} & {\scriptsize{}SSIM$\uparrow$} & {\scriptsize{}FID$\text{\ensuremath{\downarrow}}$} & {\scriptsize{}LPIPS$\text{\ensuremath{\downarrow}}$} & {\scriptsize{}PSNR$\uparrow$} & {\scriptsize{}SSIM$\uparrow$} & {\scriptsize{}FID$\text{\ensuremath{\downarrow}}$} & {\scriptsize{}LPIPS$\text{\ensuremath{\downarrow}}$}\tabularnewline
\midrule 
 &  &  & 25.09 & 0.873 & 23.62 & 0.101 & 24.86 & 0.880 & 26.79 & 0.123\tabularnewline
\CheckmarkBold{} &  &  & 25.28 & 0.877 & 21.50 & 0.097 & 25.09 & 0.887 & 24.27 & 0.203\tabularnewline
\CheckmarkBold{} & \CheckmarkBold{} &  & 25.34 & 0.878 & 20.69 & 0.095 & 25.25 & 0.888 & 23.88 & 0.206\tabularnewline
\CheckmarkBold{} & \CheckmarkBold{} & \CheckmarkBold{} & \textbf{25.77} & \textbf{0.887} & \textbf{20.46} & \textbf{0.092} & \textbf{25.37} & \textbf{0.891} & \textbf{23.24} & \textbf{0.200}\tabularnewline
\bottomrule
\end{tabular}
\vspace{-0.1in}
\end{table}

\begin{table}  \small
\caption{\textcolor{black}{Quantitative ablations for different priors used in HR-Flickr (1024$\times$1024) without training directly.\label{table:HR_priors}}}
\vspace{-0.1in}
\centering
\small
\begin{tabular}{c|cccc}
\toprule 
Priors & PSNR$\text{\ensuremath{\uparrow}}$ & SSIM$\text{\ensuremath{\uparrow}}$ & FID$\downarrow$ & LPIPS$\downarrow$\tabularnewline
\midrule
CATS & 29.03 & \textbf{0.925} & 14.88 & 0.066\tabularnewline
CATS+E-NMS & 29.01 & \textbf{0.925} & \textbf{14.72} & \textbf{0.065}\tabularnewline
Grad & 29.11 & \textbf{0.925} & 14.97 & 0.067\tabularnewline
RTV & \textbf{29.13} & \textbf{0.925} & 15.65 & 0.066\tabularnewline
\bottomrule
\end{tabular}
\end{table}

\begin{figure*}
\begin{centering}
\includegraphics[width=0.99\linewidth]{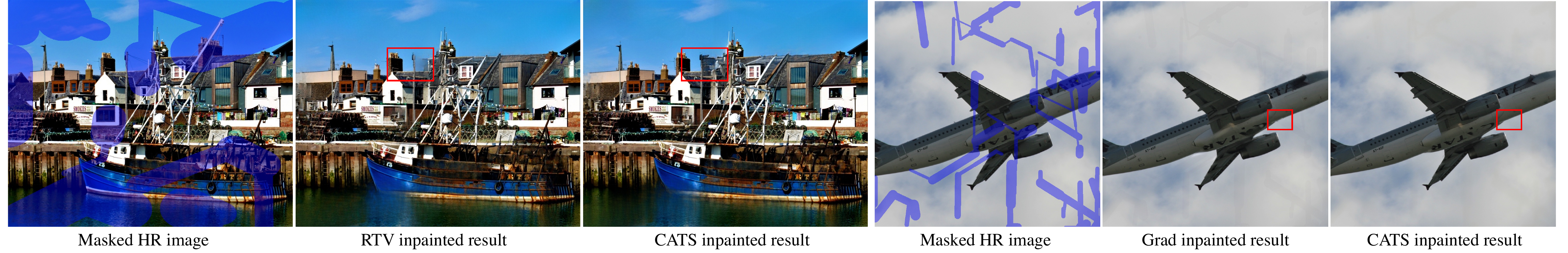}
\par\end{centering}
\vspace{-0.1in}
 \caption{\textcolor{black}{HR-Flickr inpainted results based on CATS, Grad, and RTV. For HR inpainting (1k), left: RTV suffers from color leakage; right: Grad suffers from serious color difference and inconsistent boundaries. Please zoom in for details. \label{fig:HR_prior_qua}}}
\vspace{-0.1in}
\end{figure*}

\begin{figure*}
\begin{centering}
\includegraphics[width=0.9\linewidth]{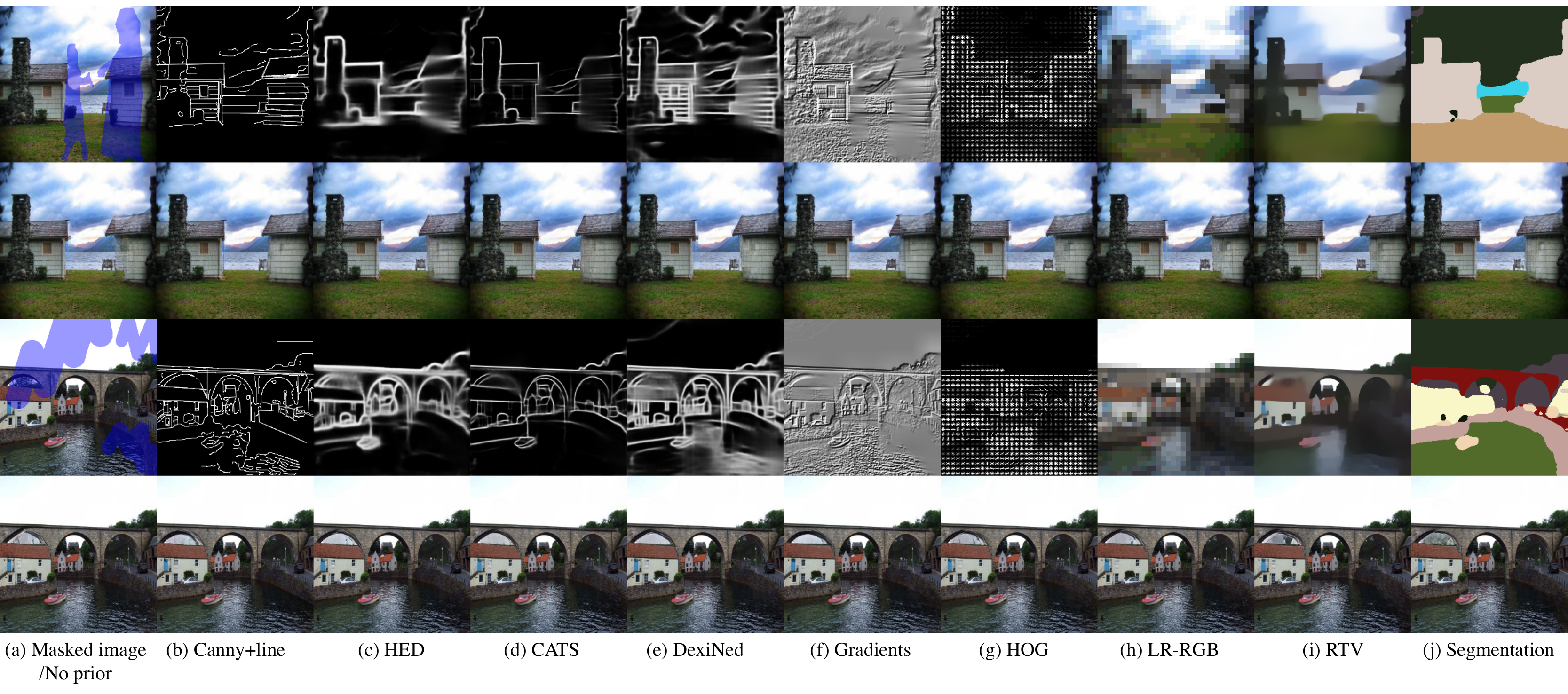}
\par\end{centering}
\vspace{-0.1in}
 \caption{\textcolor{black}{Qualitative results compared with all priors in this paper. For each group, the first row indicates masked images and restored priors, while the second one shows inpainted results with various priors. \label{fig:all_priors}}}
\vspace{-0.15in}
\end{figure*}

\noindent\textcolor{black}{\textbf{Prior Comparisons.} We show quantitative results compared with different priors in Tab.~\ref{table:single_prior_compare}. And qualitative results are illustrated in Fig.~\ref{fig:all_priors}.
Note that the baseline model is also finetuned with the same steps on the subset. All priors can improve the performance based on the baseline\footnote{\textcolor{black}{Wireframe lines do not work as an independent prior for image inpainting in this paper. Because lines are more effective in man-made scenes, while some cases have no lines at all. The effectiveness of lines based on the canny edge has been demonstrated in MST~\cite{cao2021learning}.}}.
Both lines and edges after E-NMS can be upsampled by our SSU, while other priors suffer from artifacts in boundaries during the HR inpainting as in Tab.~\ref{table:HR_priors} and Fig.~\ref{fig:HR_prior_qua}.
HOG and LR-RGB have discouraged improvements in Tab.~\ref{table:single_prior_compare} without proper structures. 
%HOG features and LR-RGB are insufficient to represent good structures for image inpainting. 
% LR-RGB suffers from ambiguous dilemmas. 
Our conference version (Canny+Line) gets moderate performance compared with other priors, while L-Edges (HED, CATS, DexiNed) achieve better results in FID. Furthermore, CATS enjoys the best performance among the three L-Edge priors. 
% More analysis about L-Edges has been discussed in Sec.~\ref{sec:dis_prior}.
Moreover, semantic segmentation works well, but we have only reported the upper bound of semantic performance here due to the gap of segmentation results from corrupted images. 
% And these two priors lose some structural details as shown in Fig.~\ref{fig:all_priors}(i)(j). 
Gradients also greatly facilitate the inpainting performance, especially in PSNR and SSIM. 
% And we will further discuss about properties of the gradient prior in Sec.~\ref{sec:dis_prior}. 
Though RTV is also competitive, it still enjoys less priority compared with L-Edges and gradients. Besides, gradient and RTV fail to be generalized to HR with the SSU.
}

\noindent\textcolor{black}{\textbf{Prior Combinations.} 
We further try to compare prior combinations among CATS, gradient, and line in Tab.~\ref{table:multi_prior_compare}. 
Structural priors achieve prominent improvement to the baseline. Specifically, CATS and gradient can work compatibly and achieve superior results for all metrics. Besides, wireframe lines benefit both $256\times256$ and $512\times512$ inpainting with better FID and LPIPS. Note that the gradient can not be generalized to untrained scales. So we only use it for the 256$\times$256 inpainting.
% More experiments based on the whole Places2 compared with other state-of-the-art models are discussed in Sec.~\ref{sec:exp}.
}

\subsection{Results of High-Resolution Inpainting}

We also compare the results of HiFill, Co-Mod, LaMa, \textcolor{black}{MAT, our ZITS and ZITS++} in Places2(512) of Tab.~\ref{table:high_res}. Besides, LaMa, ZITS, and ZITS++ are further compared in Indoor(512) and MatterPort3D(1k) in Tab.~\ref{table:high_res}. \textcolor{black}{Comparisons of HR-Flickr are shown in Tab.~\ref{table:high_res_hrflickr}.} ZITS and ZITS++ are firstly trained in 256$\times$256 and then finetuned with dynamic resolutions from 256 to 512 with 150k steps. Models tested in Indoor(512) and MatterPort3D(1k) are both trained in Indoor training set, \textcolor{black}{while methods tested for HR-Flickr are trained on Places2.} For Places2(512), we randomly select 1,000 samples from 36,500 for the 512 testing. \textcolor{black}{Our ZITS++ can outperform LaMa and even previous ZITS, which illustrates the effectiveness of the CATS edges and the updated FTR. Besides, ZITS++ can also get better 1k results in MatterPort3D.} More HR results can be seen in the supplementary.

\begin{table}\small
\small
\caption{Quantitative results of 512$\times$512 in Indoor and Places2, and 1024$\times$1024 MatterPort3D. \textcolor{black}{The best results are in bold, while the second-best ones are underlined.} \label{table:high_res}}
\vspace{-0.1in}
\renewcommand\tabcolsep{2.8pt}
\begin{centering}
\begin{tabular}{c|c|cccc}
\toprule
\multicolumn{1}{c}{} &  & PSNR$\uparrow$ & SSIM$\uparrow$ & FID$\downarrow$ & LPIPS$\downarrow$\tabularnewline
\midrule 
\multirow{3}{*}{Indoor(512)} & LaMa~\cite{suvorov2021resolution} & 24.42 & 0.911 & 21.48 & 0.143 \tabularnewline
 & ZITS~\cite{dong2022incremental} & \underline{25.36} & \underline{0.919} & \underline{18.76} & \underline{0.117} \tabularnewline
& ZITS++ & \textbf{25.68} & \textbf{0.923} & \textbf{17.73} & \textbf{0.110} \tabularnewline
\midrule
\multirow{6}{*}{Places2(512)} & HiFill~\cite{yi2020contextual} & 20.10 & 0.764 & 65.47 & 0.291\tabularnewline
 & MAT~\cite{li2022mat} & 21.68 & 0.838 & 32.43 & 0.165\tabularnewline
 & Co-Mod~\cite{zhao2021large} & 22.00 & 0.843 & 30.04 & 0.166\tabularnewline
 & LaMa~\cite{suvorov2021resolution} & 24.15 & 0.877 & 27.86 & 0.149\tabularnewline
 & ZITS~\cite{dong2022incremental} & \underline{24.23} & \underline{0.881} & \underline{26.08} & \underline{0.133}\tabularnewline
 & ZITS++ & \textbf{24.50} & \textbf{0.885} & \textbf{25.64} & \textbf{0.118} \tabularnewline
\midrule
\multirow{3}{*}{MatterPort3D(1k)} & LaMa~\cite{suvorov2021resolution} & 26.40 & 0.944 & 14.04 & 0.133 \tabularnewline
 & ZITS~\cite{dong2022incremental} & \underline{26.55} & \underline{0.946} & \underline{12.34} & \underline{0.116} \tabularnewline
  & ZITS++ & \textbf{27.49} & \textbf{0.950} & \textbf{10.86} & \textbf{0.111} \tabularnewline
\bottomrule
\end{tabular}
\par\end{centering}
\end{table}

\begin{table}  \small
\small
\caption{\textcolor{black}{Quantitative results of HR-Flickr with thin, medium (Med), and thick types of masks defined in~\cite{kulshreshtha2022feature}. SD means inpainting results from StableDiffusion-0.8B~\cite{rombach2022high}, which fails to produce results that are larger than 1K due to the memory limitation (48GB). Best results are in bold.} \label{table:high_res_hrflickr}}
\vspace{-0.1in}
\renewcommand\tabcolsep{1.5pt}
\begin{centering}
\begin{tabular}{c|c|ccc|cc|cc}
\toprule
 &  & \multicolumn{3}{c|}{HR-Flickr(1K)} & \multicolumn{2}{c|}{HR-Flickr(2K)} & \multicolumn{2}{c}{HR-Flickr(4K)}\tabularnewline
\midrule
 & Mask & LaMa & ZITS++ & SD & LaMa & ZITS++ & LaMa & ZITS++\tabularnewline
\midrule
\multirow{3}{*}{PSNR$\uparrow$} & Thin & 28.71 & \textbf{29.32} & 27.93 & 27.80 & \textbf{28.40} & 27.88 & \textbf{28.66}\tabularnewline
 & Med & 26.15 & \textbf{26.60} & 24.76 & 25.01 & \textbf{25.69} & 24.92 & \textbf{25.44}\tabularnewline
 & Thick & 24.26 & \textbf{24.68} & 22.50 & 23.52 & \textbf{23.76} & 23.22 & \textbf{23.37}\tabularnewline
\midrule
\multirow{3}{*}{SSIM$\text{\ensuremath{\uparrow}}$} & Thin & 0.921 & \textbf{0.927} & 0.918 & 0.918 & \textbf{0.925} & 0.925 & \textbf{0.934}\tabularnewline
 & Med & 0.904 & \textbf{0.912} & 0.903 & 0.910 & \textbf{0.921} & 0.915 & \textbf{0.927}\tabularnewline
 & Thick & 0.895 & \textbf{0.903} & 0.883 & 0.898 & \textbf{0.908} & 0.907 & \textbf{0.917}\tabularnewline
\midrule
\multirow{3}{*}{FID$\downarrow$} & Thin & 15.02 & \textbf{13.54} & 24.72 & 19.26 & \textbf{16.46} & 29.05 & \textbf{24.44}\tabularnewline
 & Med & 31.93 & \textbf{28.39} & 36.94 & 40.86 & \textbf{35.91} & 47.43 & \textbf{45.59}\tabularnewline
 & Thick & 40.79 & \textbf{37.95} & 44.45 & 47.84 & \textbf{45.12} & 51.59 & \textbf{50.91}\tabularnewline
\midrule
\multirow{3}{*}{LPIPS$\downarrow$} & Thin & 0.068 & \textbf{0.063} & 0.082 & 0.085 & \textbf{0.077} & 0.098 & \textbf{0.085}\tabularnewline
 & Med & 0.109 & \textbf{0.098} & 0.103 & 0.124 & \textbf{0.114} & 0.132 & \textbf{0.131}\tabularnewline
 & Thick & 0.140 & \textbf{0.130} & 0.139 & \textbf{0.153} & 0.155 & \textbf{0.154} & 0.163\tabularnewline
\bottomrule
\end{tabular}
\par\end{centering}
\end{table}

\subsection{Inpainting on Face Dataset}
\begin{figure}
\begin{centering}
\includegraphics[width=0.95\linewidth]{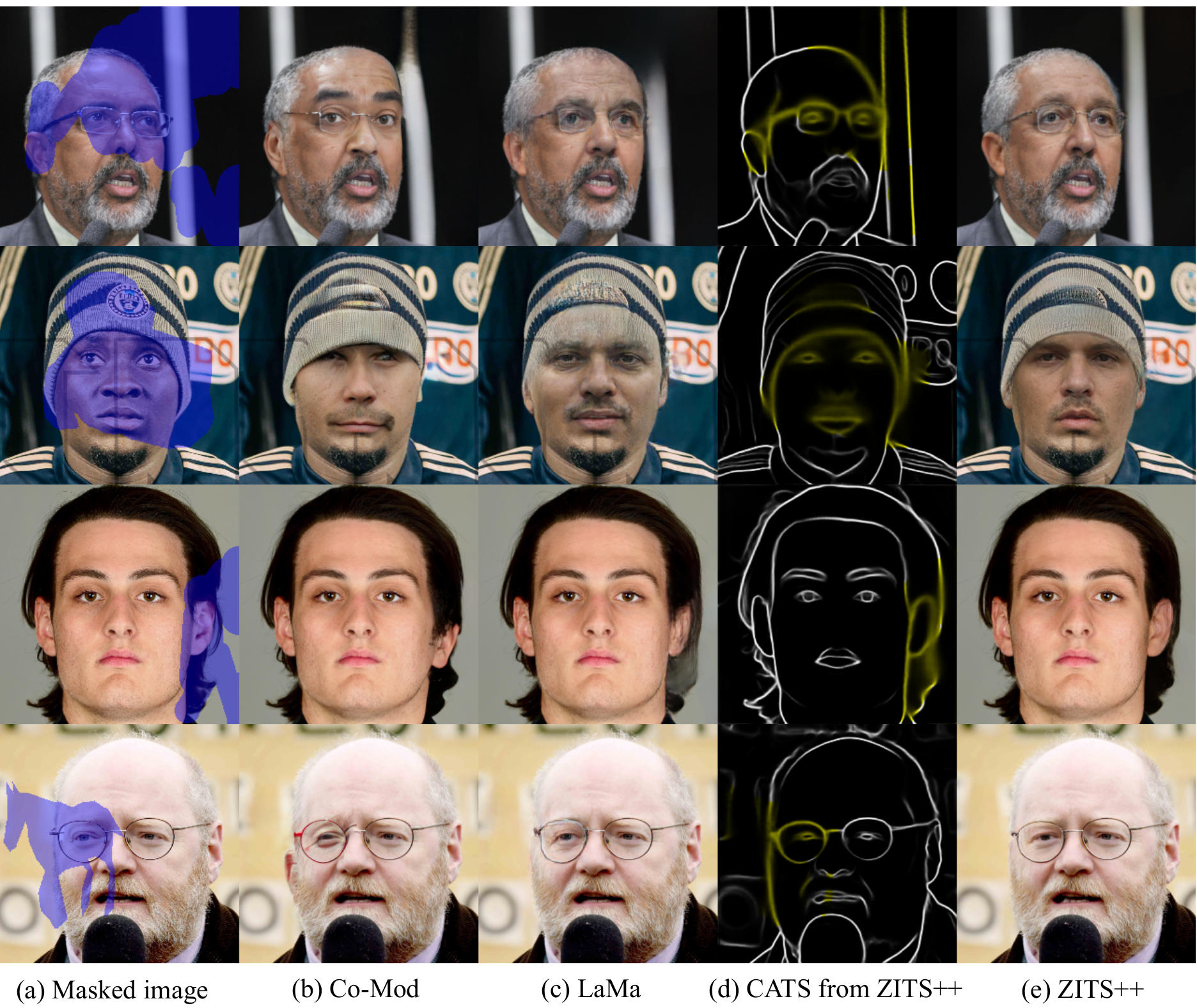}
\par\end{centering}
\vspace{-0.1in}
 \caption{\textcolor{black}{Qualitative inpainting results (256$\times$256) on FFHQ.}
 \label{fig:qualitative_ffhq+}}
 \vspace{-0.15in}
\end{figure}

\textcolor{black}{We provide inpainting results compared on the face dataset FFHQ~\cite{karras2019style} in Fig.~\ref{fig:qualitative_ffhq+} and Tab.~\ref{table:ffhq_main_results}, which is split into 68,000/2,000 images as training and test sets.
Thanks to the effectiveness of CATS, our ZITS++ outperforms Co-Mod and LaMa without the line prior. As shown in Fig.~\ref{fig:qualitative_ffhq+}, CATS L-Edges preserve structures of glasses, hats, and ears.
}

\begin{table} \small
\caption{\textcolor{black}{Quantitative results on FFHQ in 256$\times$256.} \textcolor{black}{The best results are in bold.}\label{table:ffhq_main_results}}
\vspace{-0.1in}
\begin{centering}
\footnotesize
\begin{tabular}{c|cccc}
\toprule
 & PSNR$\uparrow$ & SSIM$\uparrow$ & FID$\downarrow$ & LPIPS$\downarrow$\tabularnewline
\hline
Co-Mod~\cite{zhao2021large} & 25.25 & 0.889 & 5.85 & 0.085\tabularnewline
LaMa~\cite{suvorov2021resolution} & 26.60 & 0.903 & 6.38 & 0.078\tabularnewline
ZITS++ & \textbf{27.56} & \textbf{0.918} & \textbf{5.50} & \textbf{0.069}\tabularnewline
\bottomrule
\end{tabular}
\par\end{centering}
\end{table}

\subsection{Model Parameters and Inference Speed}

\begin{table}
\small
\caption{\textcolor{black}{Comparisons of model Parameters (Param.) and inference speed for different stages in 256 and 1024 resolutions respectively. ZITS++ contains one TSR for lines and L-Edges, while ZITS++$^*$ is enhanced with one more TSR for gradients.}\label{table:efficiency_speed}}
\vspace{-0.1in}
\centering
\begin{tabular}{cccc}
\toprule 
\multirow{2}{*}{Model} & \multirow{2}{*}{Param.} & \multicolumn{2}{c}{Stage1/Stage2/All (ms/img)}\tabularnewline
\cmidrule{3-4} \cmidrule{4-4} 
 &  & 256x256  & 1024x1024\tabularnewline
\midrule 
EC~\cite{nazeri2019edgeconnect}  & 22M & 11.2/11.1/22.3 & 149.1/149.3/298.4\tabularnewline
MST~\cite{cao2021learning}  & 26M & 16.9/13.6/30.5 & 217.8/190.1/408.0\tabularnewline
Co-Mod~\cite{zhao2021large}  & 80M & --/42.8/42.8 & --/60.4/60.4\tabularnewline
MAT~\cite{li2022mat}  & 62M & 121.0/34.9/155.9 & 330.2/149.1/479.3\tabularnewline
ICT~\cite{wan2021high}  & 122M & 8287/8.00/8295 & OOM\tabularnewline
LaMa~\cite{suvorov2021resolution}  & 27M & --/34.3/34.3 & --/103.8/103.8\tabularnewline
ZITS~\cite{dong2022incremental}  & 68M & 182.8/43.2/226.0 & 182.9/232.8/415.7\tabularnewline
ZITS++  & 83M & 44.4/47.4/91.8 & 44.5/338.7/383.2\tabularnewline
ZITS++$^*$ & 101M & 72.2/49.0/121.2 & --\tabularnewline
\bottomrule
\end{tabular}
\vspace{-0.1in}
\end{table}

\textcolor{black}{We further compare the model parameters and inference speed in Tab.~\ref{table:efficiency_speed}. All methods are validated with official codes, while we adjust the model designs of Co-Mod and MAT to make them suitable for 256 resolution. Most methods listed in Tab.~\ref{table:efficiency_speed} could be seen as two-stage models. The first stages of EC, MST, ICT, ZITS, and ZITS++ are working for the prior reconstruction. Note that the transformer-based first stage of ICT is very time-consuming for generating low-resolution images autoregressively. MAT could also be considered as a two-stage model, which contains an additional Conv-U-Net in the second stage for refinement.
Co-Mod is the fastest method in Tab.~\ref{table:efficiency_speed} due to its efficient CUDA implementation of~\cite{karras2020analyzing}.
Benefited by the informative L-Edges and the effective E-NMS technique, the TSR of ZITS++ gets rid of the costly Mask-Predict used in~\cite{dong2022incremental}. So ZITS++ is much faster than ZITS in stage 1 even with three priors (ZITS++$^*$). Moreover, thanks to the SSU (Sec.~\ref{sec:SSU}), our TSR can tackle the prior recovery in arbitrary resolutions with the same cost.
Compared with other state-of-the-art two-stage methods, the efficiency of ZITS++ is still competitive.
}

\section{Limitations and Failure Cases}

\begin{figure}
\begin{centering}
\includegraphics[width=0.95\linewidth]{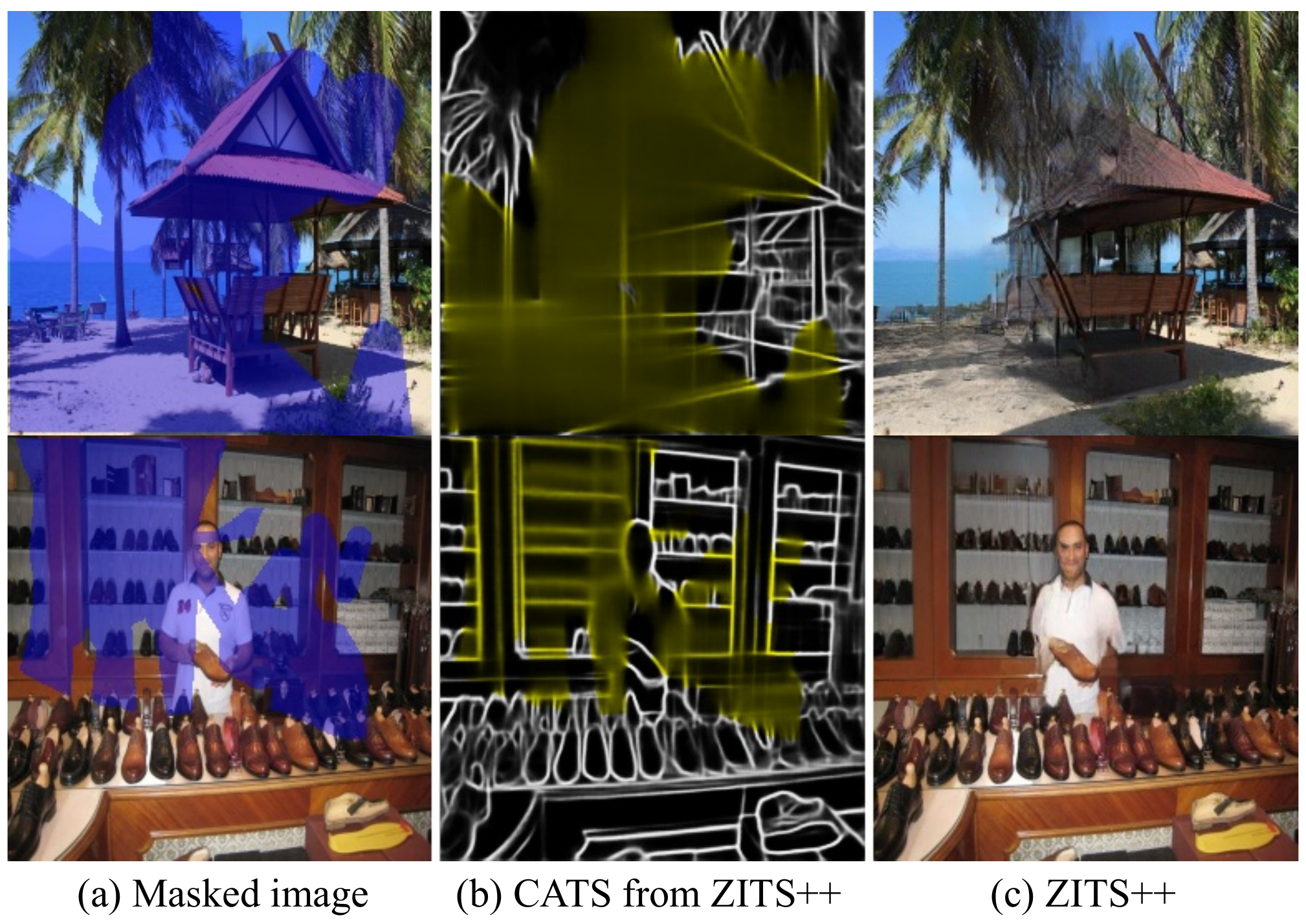}
\par\end{centering}
\vspace{-0.1in}
\caption{Failure cases of our method. ZITS++ is trained on Places2.\label{fig:failed_case}}
\vspace{-0.15in}
\end{figure}

\textcolor{black}{We summarize the limitation of our method in this section. As shown in the first row of Fig.~\ref{fig:failed_case}, our method fails to recover complex man-made buildings under extremely large masks.
Although ZITS++ can achieve good performance in object removal, it still suffers from some intractable cases, such as incomplete masks for human bodies in complicated environments as shown in the second row of Fig.~\ref{fig:failed_case}. Though our TSR can successfully understand the body structure, our FTR cannot tackle it properly.
We think that such failure is restricted by the limited model capacity. Combining our prior learning with models enjoying larger capacity, \textit{e.g.}, diffusion models, is an interesting future work.}

\section{Conclusions}

In this paper, we propose an incremental structure enhanced inpainting model called ZITS++. We use a transformer-based structure restorer to get much better holistic structures compared with previous methods. Then, a novel ZeroRA strategy is leveraged to incorporate auxiliary structures into a pre-trained inpainting model with a few finetuning steps. The proposed masking positional encoding can further improve the inpainting performance. 
Moreover, we further study different image priors for inpainting and select to leverage CATS edges~\cite{huan2021unmixing} instead of  canny~\cite{canny1986computational} used in the conference version~\cite{dong2022incremental}, which provides more informative structures to the inpainting model. 
Besides, we comprehensively upgraded LaMa~\cite{suvorov2021resolution} used as the pre-trained FTR with LKA and several useful techniques.
The newly proposed ZITS++ can achieve significant improvements based on the state-of-the-art model in experiments of various resolutions, even outperforming our previous ZITS.

\section{Acknowledgement}
This work was supported in part by the National Natural Science Foundation of China Grants (62076067, 62176061), and STCSM Project (No.22511105000).

% \section{}

% % use section* for acknowledgment
% \ifCLASSOPTIONcompsoc
%   % The Computer Society usually uses the plural form
%   \section*{Acknowledgments}
% \else
%   % regular IEEE prefers the singular form
% \fi


\ifCLASSOPTIONcaptionsoff
  \newpage
\fi



%\bibliographystyle{IEEEtran}
\bibliographystyle{IEEEtran}
\bibliography{egbib}

\begin{thebibliography}{10}
\providecommand{\url}[1]{#1}
\csname url@samestyle\endcsname
\providecommand{\newblock}{\relax}
\providecommand{\bibinfo}[2]{#2}
\providecommand{\BIBentrySTDinterwordspacing}{\spaceskip=0pt\relax}
\providecommand{\BIBentryALTinterwordstretchfactor}{4}
\providecommand{\BIBentryALTinterwordspacing}{\spaceskip=\fontdimen2\font plus
\BIBentryALTinterwordstretchfactor\fontdimen3\font minus
  \fontdimen4\font\relax}
\providecommand{\BIBforeignlanguage}[2]{{%
\expandafter\ifx\csname l@#1\endcsname\relax
\typeout{** WARNING: IEEEtran.bst: No hyphenation pattern has been}%
\typeout{** loaded for the language `#1'. Using the pattern for}%
\typeout{** the default language instead.}%
\else
\language=\csname l@#1\endcsname
\fi
#2}}
\providecommand{\BIBdecl}{\relax}
\BIBdecl

\bibitem{dong2022incremental}
Q.~Dong, C.~Cao, and Y.~Fu, ``Incremental transformer structure enhanced image
  inpainting with masking positional encoding,'' \emph{arXiv preprint
  arXiv:2203.00867}, 2022.

\bibitem{suvorov2021resolution}
R.~Suvorov, E.~Logacheva, A.~Mashikhin, A.~Remizova, A.~Ashukha, A.~Silvestrov,
  N.~Kong, H.~Goka, K.~Park, and V.~Lempitsky, ``Resolution-robust large mask
  inpainting with fourier convolutions,'' \emph{arXiv preprint
  arXiv:2109.07161}, 2021.

\bibitem{canny1986computational}
J.~Canny, ``A computational approach to edge detection,'' \emph{IEEE
  Transactions on pattern analysis and machine intelligence}, no.~6, pp.
  679--698, 1986.

\bibitem{huan2021unmixing}
L.~Huan, N.~Xue, X.~Zheng, W.~He, J.~Gong, and G.-S. Xia, ``Unmixing
  convolutional features for crisp edge detection,'' \emph{IEEE Transactions on
  Pattern Analysis and Machine Intelligence}, 2021.

\bibitem{nazeri2019edgeconnect}
K.~Nazeri, E.~Ng, T.~Joseph, F.~Qureshi, and M.~Ebrahimi, ``Edgeconnect:
  Structure guided image inpainting using edge prediction,'' in
  \emph{Proceedings of the IEEE/CVF International Conference on Computer Vision
  Workshops}, 2019.

\bibitem{cao2021learning}
C.~Cao and Y.~Fu, ``Learning a sketch tensor space for image inpainting of
  man-made scenes,'' \emph{ICCV}, 2021.

\bibitem{guo2021image}
X.~Guo, H.~Yang, and D.~Huang, ``Image inpainting via conditional texture and
  structure dual generation,'' in \emph{Proceedings of the IEEE/CVF
  International Conference on Computer Vision}, 2021, pp. 14\,134--14\,143.

\bibitem{elharrouss2020image}
O.~Elharrouss, N.~Almaadeed, S.~Al-Maadeed, and Y.~Akbari, ``Image inpainting:
  A review,'' \emph{Neural Processing Letters}, vol.~51, no.~2, pp. 2007--2028,
  2020.

\bibitem{jo2019sc}
Y.~Jo and J.~Park, ``Sc-fegan: Face editing generative adversarial network with
  user's sketch and color,'' in \emph{Proceedings of the IEEE/CVF International
  Conference on Computer Vision}, 2019, pp. 1745--1753.

\bibitem{Bertalmo2000ImageI}
M.~Bertalm{\'i}o, G.~Sapiro, V.~Caselles, and C.~Ballester, ``Image
  inpainting,'' \emph{Proceedings of the 27th annual conference on Computer
  graphics and interactive techniques}, 2000.

\bibitem{Levin2003LearningHT}
A.~Levin, A.~Zomet, and Y.~Weiss, ``Learning how to inpaint from global image
  statistics,'' \emph{Proceedings Ninth IEEE International Conference on
  Computer Vision}, pp. 305--312 vol.1, 2003.

\bibitem{Roth2005FieldsOE}
S.~Roth and M.~J. Black, ``Fields of experts: a framework for learning image
  priors,'' \emph{2005 IEEE Computer Society Conference on Computer Vision and
  Pattern Recognition (CVPR'05)}, vol.~2, pp. 860--867 vol. 2, 2005.

\bibitem{Hays:2007}
J.~Hays and A.~A. Efros, ``Scene completion using millions of photographs,''
  \emph{ACM Transactions on Graphics (SIGGRAPH 2007)}, vol.~26, no.~3, 2007.

\bibitem{criminisi2003object}
A.~Criminisi, P.~Perez, and K.~Toyama, ``Object removal by exemplar-based
  inpainting,'' in \emph{2003 IEEE Computer Society Conference on Computer
  Vision and Pattern Recognition, 2003. Proceedings.}, vol.~2.\hskip 1em plus
  0.5em minus 0.4em\relax IEEE, 2003, pp. II--II.

\bibitem{krizhevsky2012imagenet}
A.~Krizhevsky, I.~Sutskever, and G.~E. Hinton, ``Imagenet classification with
  deep convolutional neural networks,'' \emph{Advances in neural information
  processing systems}, vol.~25, pp. 1097--1105, 2012.

\bibitem{goodfellow2014generative}
I.~Goodfellow, J.~Pouget-Abadie, M.~Mirza, B.~Xu, D.~Warde-Farley, S.~Ozair,
  A.~Courville, and Y.~Bengio, ``Generative adversarial nets,'' \emph{Advances
  in neural information processing systems}, vol.~27, 2014.

\bibitem{wan2021high}
Z.~Wan, J.~Zhang, D.~Chen, and J.~Liao, ``High-fidelity pluralistic image
  completion with transformers,'' in \emph{Proceedings of the IEEE/CVF
  International Conference on Computer Vision}, 2021, pp. 4692--4701.

\bibitem{yi2020contextual}
Z.~Yi, Q.~Tang, S.~Azizi, D.~Jang, and Z.~Xu, ``Contextual residual aggregation
  for ultra high-resolution image inpainting,'' in \emph{Proceedings of the
  IEEE/CVF Conference on Computer Vision and Pattern Recognition}, 2020, pp.
  7508--7517.

\bibitem{zhao2021large}
S.~Zhao, J.~Cui, Y.~Sheng, Y.~Dong, X.~Liang, E.~I. Chang, and Y.~Xu, ``Large
  scale image completion via co-modulated generative adversarial networks,''
  \emph{arXiv preprint arXiv:2103.10428}, 2021.

\bibitem{li2022mat}
W.~Li, Z.~Lin, K.~Zhou, L.~Qi, Y.~Wang, and J.~Jia, ``Mat: Mask-aware
  transformer for large hole image inpainting,'' in \emph{Proceedings of the
  IEEE/CVF Conference on Computer Vision and Pattern Recognition}, 2022, pp.
  10\,758--10\,768.

\bibitem{yu2018generative}
J.~Yu, Z.~Lin, J.~Yang, X.~Shen, X.~Lu, and T.~S. Huang, ``Generative image
  inpainting with contextual attention,'' in \emph{Proceedings of the IEEE
  conference on computer vision and pattern recognition}, 2018, pp. 5505--5514.

\bibitem{zeng2020high}
Y.~Zeng, Z.~Lin, J.~Yang, J.~Zhang, E.~Shechtman, and H.~Lu, ``High-resolution
  image inpainting with iterative confidence feedback and guided upsampling,''
  in \emph{European Conference on Computer Vision}.\hskip 1em plus 0.5em minus
  0.4em\relax Springer, 2020, pp. 1--17.

\bibitem{yu2021diverse}
Y.~Yu, F.~Zhan, R.~Wu, J.~Pan, K.~Cui, S.~Lu, F.~Ma, X.~Xie, and C.~Miao,
  ``Diverse image inpainting with bidirectional and autoregressive
  transformers,'' \emph{arXiv preprint arXiv:2104.12335}, 2021.

\bibitem{liao2020guidance}
L.~Liao, J.~Xiao, Z.~Wang, C.-W. Lin, and S.~Satoh, ``Guidance and evaluation:
  Semantic-aware image inpainting for mixed scenes,'' in \emph{Computer
  Vision--ECCV 2020: 16th European Conference, Glasgow, UK, August 23--28,
  2020, Proceedings, Part XXVII 16}.\hskip 1em plus 0.5em minus 0.4em\relax
  Springer, 2020, pp. 683--700.

\bibitem{song2018spg}
Y.~Song, C.~Yang, Y.~Shen, P.~Wang, Q.~Huang, and C.-C.~J. Kuo, ``Spg-net:
  Segmentation prediction and guidance network for image inpainting,''
  \emph{arXiv preprint arXiv:1805.03356}, 2018.

\bibitem{ren2019structureflow}
Y.~Ren, X.~Yu, R.~Zhang, T.~H. Li, S.~Liu, and G.~Li, ``Structureflow: Image
  inpainting via structure-aware appearance flow,'' in \emph{Proceedings of the
  IEEE/CVF International Conference on Computer Vision}, 2019, pp. 181--190.

\bibitem{liu2020rethinking}
H.~Liu, B.~Jiang, Y.~Song, W.~Huang, and C.~Yang, ``Rethinking image inpainting
  via a mutual encoder-decoder with feature equalizations,'' in \emph{European
  Conference on Computer Vision}.\hskip 1em plus 0.5em minus 0.4em\relax
  Springer, 2020, pp. 725--741.

\bibitem{xu2020e2i}
S.~Xu, D.~Liu, and Z.~Xiong, ``E2i: Generative inpainting from edge to image,''
  \emph{IEEE Transactions on Circuits and Systems for Video Technology},
  vol.~31, no.~4, pp. 1308--1322, 2020.

\bibitem{yang2020learning}
J.~Yang, Z.~Qi, and Y.~Shi, ``Learning to incorporate structure knowledge for
  image inpainting,'' in \emph{Proceedings of the AAAI Conference on Artificial
  Intelligence}, vol.~34, no.~07, 2020, pp. 12\,605--12\,612.

\bibitem{zhangijcai2021}
W.~Zhang, J.~Zhu, Y.~Tai, Y.~Wang, W.~Chu, B.~Ni, C.~Wang, and X.~Yang,
  ``Context-aware image inpainting with learned semantic priors,'' in
  \emph{Proceedings of the Thirtieth International Joint Conference on
  Artificial Intelligence, {IJCAI-21}}, Z.-H. Zhou, Ed.\hskip 1em plus 0.5em
  minus 0.4em\relax International Joint Conferences on Artificial Intelligence
  Organization, 8 2021, pp. 1323--1329, main Track.

\bibitem{huang2018learning}
K.~Huang, Y.~Wang, Z.~Zhou, T.~Ding, S.~Gao, and Y.~Ma, ``Learning to parse
  wireframes in images of man-made environments,'' in \emph{Proceedings of the
  IEEE Conference on Computer Vision and Pattern Recognition}, 2018, pp.
  626--635.

\bibitem{islam2020much}
M.~A. Islam, S.~Jia, and N.~D. Bruce, ``How much position information do
  convolutional neural networks encode?'' \emph{arXiv preprint
  arXiv:2001.08248}, 2020.

\bibitem{xu2020positional}
R.~Xu, X.~Wang, K.~Chen, B.~Zhou, and C.~C. Loy, ``Positional encoding as
  spatial inductive bias in gans,'' 2020.

\bibitem{lin2021infinitygan}
C.~H. Lin, H.-Y. Lee, Y.-C. Cheng, S.~Tulyakov, and M.-H. Yang, ``Infinitygan:
  Towards infinite-pixel image synthesis,'' 2021.

\bibitem{bachlechner2020rezero}
T.~Bachlechner, B.~P. Majumder, H.~Mao, G.~Cottrell, and J.~McAuley, ``Rezero
  is all you need: Fast convergence at large depth,'' in \emph{Uncertainty in
  Artificial Intelligence}.\hskip 1em plus 0.5em minus 0.4em\relax PMLR, 2021,
  pp. 1352--1361.

\bibitem{guo2022visual}
M.-H. Guo, C.-Z. Lu, Z.-N. Liu, M.-M. Cheng, and S.-M. Hu, ``Visual attention
  network,'' \emph{arXiv preprint arXiv:2202.09741}, 2022.

\bibitem{xie2015holistically}
S.~Xie and Z.~Tu, ``Holistically-nested edge detection,'' in \emph{Proceedings
  of the IEEE international conference on computer vision}, 2015, pp.
  1395--1403.

\bibitem{xu2012structure}
L.~Xu, Q.~Yan, Y.~Xia, and J.~Jia, ``Structure extraction from texture via
  relative total variation,'' \emph{ACM transactions on graphics (TOG)},
  vol.~31, no.~6, pp. 1--10, 2012.

\bibitem{dalal2005histograms}
N.~Dalal and B.~Triggs, ``Histograms of oriented gradients for human
  detection,'' in \emph{2005 IEEE computer society conference on computer
  vision and pattern recognition (CVPR'05)}, vol.~1.\hskip 1em plus 0.5em minus
  0.4em\relax Ieee, 2005, pp. 886--893.

\bibitem{zhou2017places}
B.~Zhou, A.~Lapedriza, A.~Khosla, A.~Oliva, and A.~Torralba, ``Places: A 10
  million image database for scene recognition,'' \emph{IEEE transactions on
  pattern analysis and machine intelligence}, vol.~40, no.~6, pp. 1452--1464,
  2017.

\bibitem{Silberman:ECCV12}
P.~K. Nathan~Silberman, Derek~Hoiem and R.~Fergus, ``Indoor segmentation and
  support inference from rgbd images,'' in \emph{ECCV}, 2012.

\bibitem{chang2017matterport3d}
A.~Chang, A.~Dai, T.~Funkhouser, M.~Halber, M.~Niessner, M.~Savva, S.~Song,
  A.~Zeng, and Y.~Zhang, ``Matterport3d: Learning from rgb-d data in indoor
  environments,'' \emph{arXiv preprint arXiv:1709.06158}, 2017.

\bibitem{karras2019style}
T.~Karras, S.~Laine, and T.~Aila, ``A style-based generator architecture for
  generative adversarial networks,'' in \emph{Proceedings of the IEEE/CVF
  conference on computer vision and pattern recognition}, 2019, pp. 4401--4410.

\bibitem{li2017localization}
H.~Li, W.~Luo, and J.~Huang, ``Localization of diffusion-based inpainting in
  digital images,'' \emph{IEEE transactions on information forensics and
  security}, vol.~12, no.~12, pp. 3050--3064, 2017.

\bibitem{ruzic2015context}
T.~Ruzic and A.~Pizurica, ``Context-aware patch-based image inpainting using
  markov random field modeling,'' \emph{IEEE transactions on image processing},
  vol.~24, no.~1, pp. 444--456, 2015.

\bibitem{pathak2016context}
D.~Pathak, P.~Krahenbuhl, J.~Donahue, T.~Darrell, and A.~A. Efros, ``Context
  encoders: Feature learning by inpainting,'' in \emph{Proceedings of the IEEE
  conference on computer vision and pattern recognition}, 2016, pp. 2536--2544.

\bibitem{liu2018image}
G.~Liu, F.~A. Reda, K.~J. Shih, T.-C. Wang, A.~Tao, and B.~Catanzaro, ``Image
  inpainting for irregular holes using partial convolutions,'' in
  \emph{Proceedings of the European Conference on Computer Vision (ECCV)},
  2018, pp. 85--100.

\bibitem{yu2019free}
J.~Yu, Z.~Lin, J.~Yang, X.~Shen, X.~Lu, and T.~S. Huang, ``Free-form image
  inpainting with gated convolution,'' in \emph{Proceedings of the IEEE/CVF
  International Conference on Computer Vision}, 2019, pp. 4471--4480.

\bibitem{zeng2022aggregated}
Y.~Zeng, J.~Fu, H.~Chao, and B.~Guo, ``Aggregated contextual transformations
  for high-resolution image inpainting,'' \emph{IEEE Transactions on
  Visualization and Computer Graphics}, 2022.

\bibitem{karras2020analyzing}
T.~Karras, S.~Laine, M.~Aittala, J.~Hellsten, J.~Lehtinen, and T.~Aila,
  ``Analyzing and improving the image quality of stylegan,'' in
  \emph{Proceedings of the IEEE/CVF Conference on Computer Vision and Pattern
  Recognition}, 2020, pp. 8110--8119.

\bibitem{dhariwal2021diffusion}
P.~Dhariwal and A.~Nichol, ``Diffusion models beat gans on image synthesis,''
  \emph{Advances in Neural Information Processing Systems}, vol.~34, pp.
  8780--8794, 2021.

\bibitem{rombach2022high}
R.~Rombach, A.~Blattmann, D.~Lorenz, P.~Esser, and B.~Ommer, ``High-resolution
  image synthesis with latent diffusion models,'' in \emph{Proceedings of the
  IEEE/CVF Conference on Computer Vision and Pattern Recognition}, 2022, pp.
  10\,684--10\,695.

\bibitem{saharia2022palette}
C.~Saharia, W.~Chan, H.~Chang, C.~Lee, J.~Ho, T.~Salimans, D.~Fleet, and
  M.~Norouzi, ``Palette: Image-to-image diffusion models,'' in \emph{ACM
  SIGGRAPH 2022 Conference Proceedings}, 2022, pp. 1--10.

\bibitem{nichol2021glide}
A.~Nichol, P.~Dhariwal, A.~Ramesh, P.~Shyam, P.~Mishkin, B.~McGrew,
  I.~Sutskever, and M.~Chen, ``Glide: Towards photorealistic image generation
  and editing with text-guided diffusion models,'' \emph{arXiv preprint
  arXiv:2112.10741}, 2021.

\bibitem{lugmayr2022repaint}
A.~Lugmayr, M.~Danelljan, A.~Romero, F.~Yu, R.~Timofte, and L.~Van~Gool,
  ``Repaint: Inpainting using denoising diffusion probabilistic models,'' in
  \emph{Proceedings of the IEEE/CVF Conference on Computer Vision and Pattern
  Recognition}, 2022, pp. 11\,461--11\,471.

\bibitem{chung2022come}
H.~Chung, B.~Sim, and J.~C. Ye, ``Come-closer-diffuse-faster: Accelerating
  conditional diffusion models for inverse problems through stochastic
  contraction,'' in \emph{Proceedings of the IEEE/CVF Conference on Computer
  Vision and Pattern Recognition}, 2022, pp. 12\,413--12\,422.

\bibitem{chung2022improving}
H.~Chung, B.~Sim, D.~Ryu, and J.~C. Ye, ``Improving diffusion models for
  inverse problems using manifold constraints,'' \emph{arXiv preprint
  arXiv:2206.00941}, 2022.

\bibitem{gao2020flow}
C.~Gao, A.~Saraf, J.-B. Huang, and J.~Kopf, ``Flow-edge guided video
  completion,'' in \emph{Computer Vision--ECCV 2020: 16th European Conference,
  Glasgow, UK, August 23--28, 2020, Proceedings, Part XII 16}.\hskip 1em plus
  0.5em minus 0.4em\relax Springer, 2020, pp. 713--729.

\bibitem{li2022towards}
Z.~Li, C.-Z. Lu, J.~Qin, C.-L. Guo, and M.-M. Cheng, ``Towards an end-to-end
  framework for flow-guided video inpainting,'' in \emph{Proceedings of the
  IEEE/CVF Conference on Computer Vision and Pattern Recognition}, 2022, pp.
  17\,562--17\,571.

\bibitem{zhang2022flow}
K.~Zhang, J.~Fu, and D.~Liu, ``Flow-guided transformer for video inpainting,''
  in \emph{Computer Vision--ECCV 2022: 17th European Conference, Tel Aviv,
  Israel, October 23--27, 2022, Proceedings, Part XVIII}.\hskip 1em plus 0.5em
  minus 0.4em\relax Springer, 2022, pp. 74--90.

\bibitem{chen2020generative}
M.~Chen, A.~Radford, R.~Child, J.~Wu, H.~Jun, D.~Luan, and I.~Sutskever,
  ``Generative pretraining from pixels,'' in \emph{International Conference on
  Machine Learning}.\hskip 1em plus 0.5em minus 0.4em\relax PMLR, 2020, pp.
  1691--1703.

\bibitem{vaswani2017attention}
A.~Vaswani, N.~Shazeer, N.~Parmar, J.~Uszkoreit, L.~Jones, A.~N. Gomez,
  {\L}.~Kaiser, and I.~Polosukhin, ``Attention is all you need,'' in
  \emph{Advances in neural information processing systems}, 2017, pp.
  5998--6008.

\bibitem{dosovitskiy2020image}
A.~Dosovitskiy, L.~Beyer, A.~Kolesnikov, D.~Weissenborn, X.~Zhai,
  T.~Unterthiner, M.~Dehghani, M.~Minderer, G.~Heigold, S.~Gelly \emph{et~al.},
  ``An image is worth 16x16 words: Transformers for image recognition at
  scale,'' \emph{arXiv preprint arXiv:2010.11929}, 2020.

\bibitem{esser2021taming}
P.~Esser, R.~Rombach, and B.~Ommer, ``Taming transformers for high-resolution
  image synthesis,'' in \emph{Proceedings of the IEEE/CVF Conference on
  Computer Vision and Pattern Recognition}, 2021, pp. 12\,873--12\,883.

\bibitem{ramesh2021zeroshot}
A.~Ramesh, M.~Pavlov, G.~Goh, S.~Gray, C.~Voss, A.~Radford, M.~Chen, and
  I.~Sutskever, ``Zero-shot text-to-image generation,'' 2021.

\bibitem{he2022masked}
K.~He, X.~Chen, S.~Xie, Y.~Li, P.~Doll{\'a}r, and R.~Girshick, ``Masked
  autoencoders are scalable vision learners,'' in \emph{Proceedings of the
  IEEE/CVF Conference on Computer Vision and Pattern Recognition}, 2022, pp.
  16\,000--16\,009.

\bibitem{chi2020fast}
L.~Chi, B.~Jiang, and Y.~Mu, ``Fast fourier convolution,'' \emph{Advances in
  Neural Information Processing Systems}, vol.~33, 2020.

\bibitem{ding2022scaling}
X.~Ding, X.~Zhang, J.~Han, and G.~Ding, ``Scaling up your kernels to 31x31:
  Revisiting large kernel design in cnns,'' in \emph{Proceedings of the
  IEEE/CVF Conference on Computer Vision and Pattern Recognition}, 2022, pp.
  11\,963--11\,975.

\bibitem{liu2022more}
S.~Liu, T.~Chen, X.~Chen, X.~Chen, Q.~Xiao, B.~Wu, M.~Pechenizkiy, D.~Mocanu,
  and Z.~Wang, ``More convnets in the 2020s: Scaling up kernels beyond 51x51
  using sparsity,'' \emph{arXiv preprint arXiv:2207.03620}, 2022.

\bibitem{xiao2021early}
T.~Xiao, P.~Dollar, M.~Singh, E.~Mintun, T.~Darrell, and R.~Girshick, ``Early
  convolutions help transformers see better,'' \emph{Advances in Neural
  Information Processing Systems}, vol.~34, 2021.

\bibitem{ho2019axial}
J.~Ho, N.~Kalchbrenner, D.~Weissenborn, and T.~Salimans, ``Axial attention in
  multidimensional transformers,'' \emph{arXiv preprint arXiv:1912.12180},
  2019.

\bibitem{huang2019ccnet}
Z.~Huang, X.~Wang, L.~Huang, C.~Huang, Y.~Wei, and W.~Liu, ``Ccnet: Criss-cross
  attention for semantic segmentation,'' in \emph{Proceedings of the IEEE/CVF
  international conference on computer vision}, 2019, pp. 603--612.

\bibitem{raffel2019exploring}
C.~Raffel, N.~Shazeer, A.~Roberts, K.~Lee, S.~Narang, M.~Matena, Y.~Zhou,
  W.~Li, and P.~J. Liu, ``Exploring the limits of transfer learning with a
  unified text-to-text transformer,'' \emph{arXiv preprint arXiv:1910.10683},
  2019.

\bibitem{xiong2020layer}
R.~Xiong, Y.~Yang, D.~He, K.~Zheng, S.~Zheng, C.~Xing, H.~Zhang, Y.~Lan,
  L.~Wang, and T.~Liu, ``On layer normalization in the transformer
  architecture,'' in \emph{International Conference on Machine Learning}.\hskip
  1em plus 0.5em minus 0.4em\relax PMLR, 2020, pp. 10\,524--10\,533.

\bibitem{xue2020holistically}
N.~Xue, T.~Wu, S.~Bai, F.~Wang, G.-S. Xia, L.~Zhang, and P.~H. Torr,
  ``Holistically-attracted wireframe parsing,'' in \emph{Proceedings of the
  IEEE/CVF Conference on Computer Vision and Pattern Recognition}, 2020, pp.
  2788--2797.

\bibitem{ZitnickECCV14edgeBoxes}
C.~L. Zitnick and P.~Doll\'ar, ``Edge boxes: Locating object proposals from
  edges,'' in \emph{ECCV}, 2014.

\bibitem{isola2017image}
P.~Isola, J.-Y. Zhu, T.~Zhou, and A.~A. Efros, ``Image-to-image translation
  with conditional adversarial networks,'' in \emph{Proceedings of the IEEE
  conference on computer vision and pattern recognition}, 2017, pp. 1125--1134.

\bibitem{ramachandran2017searching}
P.~Ramachandran, B.~Zoph, and Q.~V. Le, ``Searching for activation functions,''
  \emph{arXiv preprint arXiv:1710.05941}, 2017.

\bibitem{yu2015multi}
F.~Yu and V.~Koltun, ``Multi-scale context aggregation by dilated
  convolutions,'' \emph{arXiv preprint arXiv:1511.07122}, 2015.

\bibitem{gulrajani2017improved}
I.~Gulrajani, F.~Ahmed, M.~Arjovsky, V.~Dumoulin, and A.~Courville, ``Improved
  training of wasserstein gans,'' \emph{arXiv preprint arXiv:1704.00028}, 2017.

\bibitem{wang2018high}
T.-C. Wang, M.-Y. Liu, J.-Y. Zhu, A.~Tao, J.~Kautz, and B.~Catanzaro,
  ``High-resolution image synthesis and semantic manipulation with conditional
  gans,'' in \emph{Proceedings of the IEEE conference on computer vision and
  pattern recognition}, 2018, pp. 8798--8807.

\bibitem{ghazvininejad2019mask}
M.~Ghazvininejad, O.~Levy, Y.~Liu, and L.~Zettlemoyer, ``Mask-predict: Parallel
  decoding of conditional masked language models,'' \emph{arXiv preprint
  arXiv:1904.09324}, 2019.

\bibitem{zhou2019end}
Y.~Zhou, H.~Qi, and Y.~Ma, ``End-to-end wireframe parsing,'' in
  \emph{Proceedings of the IEEE/CVF International Conference on Computer
  Vision}, 2019, pp. 962--971.

\bibitem{poma2021dense}
X.~S. Poma, A.~Sappa, P.~Humanante, and A.~Arbarinia, ``Dense extreme inception
  network for edge detection,'' \emph{arXiv preprint arXiv:2112.02250}, 2021.

\bibitem{DBLP:journals/corr/ZhouZPFBT16}
B.~Zhou, H.~Zhao, X.~Puig, S.~Fidler, A.~Barriuso, and A.~Torralba, ``Semantic
  understanding of scenes through the {ADE20K} dataset,'' \emph{CoRR}, vol.
  abs/1608.05442, 2016.

\bibitem{zheng2021rethinking}
S.~Zheng, J.~Lu, H.~Zhao, X.~Zhu, Z.~Luo, Y.~Wang, Y.~Fu, J.~Feng, T.~Xiang,
  P.~H. Torr \emph{et~al.}, ``Rethinking semantic segmentation from a
  sequence-to-sequence perspective with transformers,'' in \emph{Proceedings of
  the IEEE/CVF conference on computer vision and pattern recognition}, 2021,
  pp. 6881--6890.

\bibitem{Lim_2017_CVPR_Workshops}
B.~Lim, S.~Son, H.~Kim, S.~Nah, and K.~M. Lee, ``Enhanced deep residual
  networks for single image super-resolution,'' in \emph{The IEEE Conference on
  Computer Vision and Pattern Recognition (CVPR) Workshops}, July 2017.

\bibitem{wang2004image}
Z.~Wang, A.~C. Bovik, H.~R. Sheikh, and E.~P. Simoncelli, ``Image quality
  assessment: from error visibility to structural similarity,'' \emph{IEEE
  transactions on image processing}, vol.~13, no.~4, pp. 600--612, 2004.

\bibitem{heusel2018gans}
M.~Heusel, H.~Ramsauer, T.~Unterthiner, B.~Nessler, and S.~Hochreiter, ``Gans
  trained by a two time-scale update rule converge to a local nash
  equilibrium,'' 2018.

\bibitem{zhang2018unreasonable}
R.~Zhang, P.~Isola, A.~A. Efros, E.~Shechtman, and O.~Wang, ``The unreasonable
  effectiveness of deep features as a perceptual metric,'' in \emph{Proceedings
  of the IEEE conference on computer vision and pattern recognition}, 2018, pp.
  586--595.

\bibitem{kulshreshtha2022feature}
P.~Kulshreshtha, B.~Pugh, and S.~Jiddi, ``Feature refinement to improve high
  resolution image inpainting,'' \emph{arXiv preprint arXiv:2206.13644}, 2022.

\bibitem{wei2021masked}
C.~Wei, H.~Fan, S.~Xie, C.-Y. Wu, A.~Yuille, and C.~Feichtenhofer, ``Masked
  feature prediction for self-supervised visual pre-training,'' \emph{arXiv
  preprint arXiv:2112.09133}, 2021.

\bibitem{marr2010vision}
D.~Marr, \emph{Vision: A computational investigation into the human
  representation and processing of visual information}.\hskip 1em plus 0.5em
  minus 0.4em\relax MIT press, 2010.

\bibitem{stevens2012vision}
K.~A. Stevens, ``The vision of david marr,'' \emph{Perception}, vol.~41, no.~9,
  pp. 1061--1072, 2012.

\bibitem{abid2019gradio}
A.~Abid, A.~Abdalla, A.~Abid, D.~Khan, A.~Alfozan, and J.~Zou, ``Gradio:
  Hassle-free sharing and testing of ml models in the wild,'' \emph{arXiv
  preprint arXiv:1906.02569}, 2019.

\bibitem{cimpoi2014describing}
M.~Cimpoi, S.~Maji, I.~Kokkinos, S.~Mohamed, and A.~Vedaldi, ``Describing
  textures in the wild,'' in \emph{Proceedings of the IEEE conference on
  computer vision and pattern recognition}, 2014, pp. 3606--3613.

\end{thebibliography}


\begin{IEEEbiography}[{\includegraphics[width=1in,height=1.25in,clip,keepaspectratio]{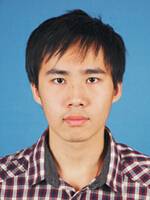}}]{Chenjie Cao} received the M.S. degree in Computer Science from East China University of Science and Technology, in 2019. He is currently pursuing a Ph.D. degree in Statistics from Fudan University. His research interests include machine learning, deep learning, image inpainting, image editing, 3D shape reconstruction, and multi-view stereo.
\end{IEEEbiography}

\begin{IEEEbiography}[{\includegraphics[width=1in,height=1.25in,clip,keepaspectratio]{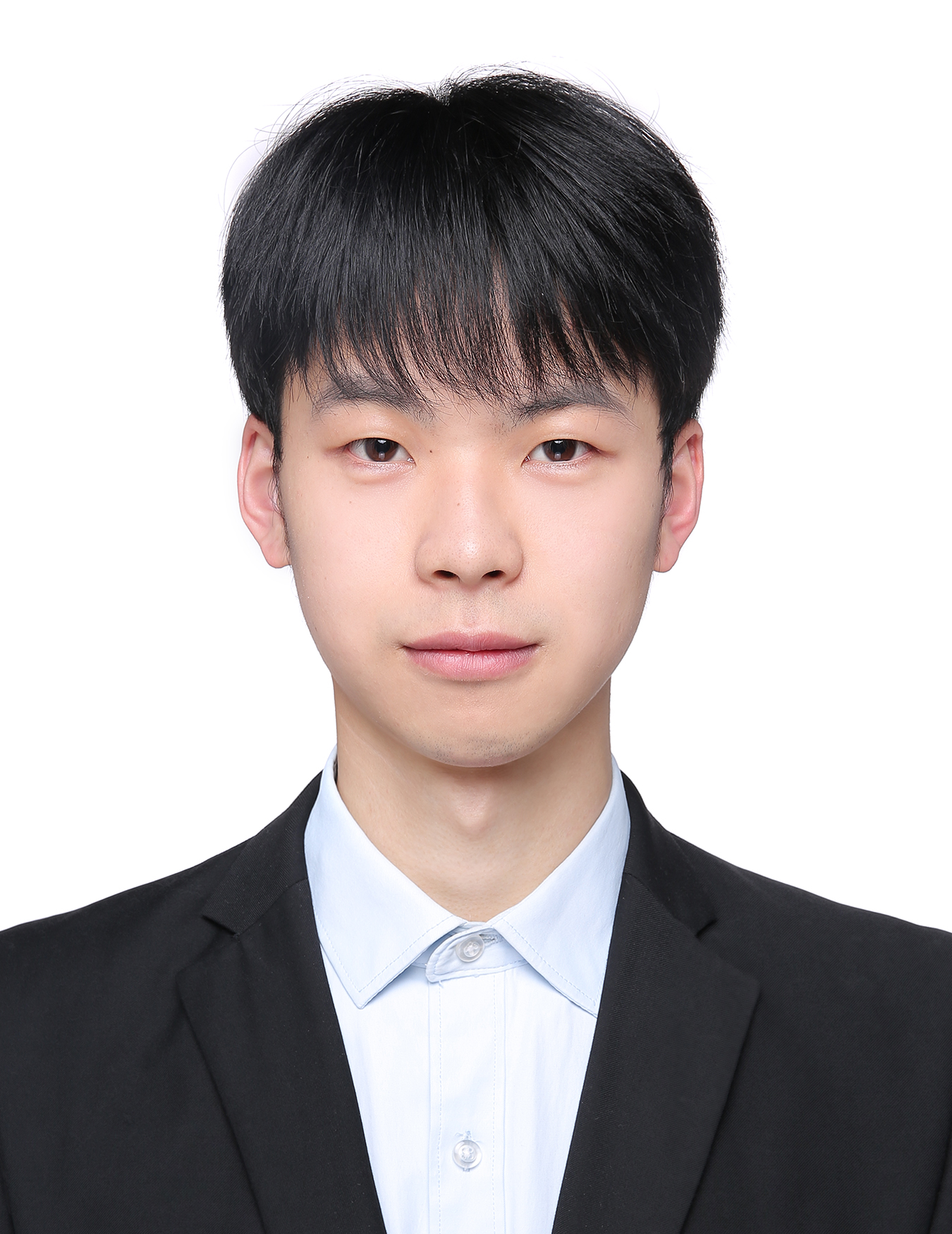}}]{Qiaole Dong} received the B.S. degree in data science from Fudan University, China, in 2022. He is currently pursuing a Ph.D. degree in Statistics from Fudan
University with supervisor Dr. Yanwei Fu. His research interests include machine learning, deep learning, image inpainting, image editing, and optical flow estimation.
\end{IEEEbiography}

\begin{IEEEbiography}[{\includegraphics[width=1in,height=1.25in,clip,keepaspectratio]{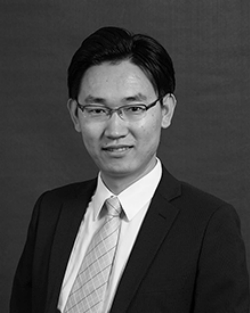}}]{Yanwei Fu} received the MEng degree from the Department of Computer Science and Technology, Nanjing University, China, in 2011, and the PhD degree from the Queen Mary University of London, in 2014. He held a post-doctoral position at Disney Research, Pittsburgh, PA, from 2015 to 2016. He is currently a tenure-track professor at Fudan University.  
He was appointed as the Professor of Special Appointment (Eastern Scholar) at Shanghai Institutions of Higher Learning.
His work has led to many awards, including the IEEE ICME 2019 best paper.
He published more than 100 journal/conference papers including IEEE TPAMI, TMM, ECCV, and CVPR. His research interests are one-shot learning, and learning-based 3D reconstruction.
\end{IEEEbiography}


%\vfill


\section*{Appendix}

\section{Broader Impacts}
All generated results of both the main paper and the supplementary are based on learned statistics of the training dataset. Therefore, the results only reflect biases in those data without our subjective opinion. This work is only researched for the algorithmic discussion, and related societal impacts should not be ignored by users.

\section{Introduction of Other Priors}

\noindent\textcolor{black}{\textbf{HOG}~\cite{dalal2005histograms} 
is a feature descriptor of gradient orientation distributions within local patches. 
% HOG is invariant to photometric and geometric changes with small translation and rotation, which helps to capture local shapes. 
MaskFeat~\cite{wei2021masked} finds that HOG-based self-supervised pre-training video model can achieve very good performance  in ImageNet.
Therefore, HOG has been studied as a prior to inpainting.
As to~\cite{wei2021masked}, we utilize a vanilla ViT~\cite{dosovitskiy2020image} to inpaint HOG with masked inputs in RGB channels. 
Each cell comprises $8\times8$ pixels with 9 orientations.
Thus a $256\times 256$ image results in $\mathbf{\hat{P}}_h\in\mathbb{R}^{32\times 32\times 27}$ HOG features. $L_2$ loss is used to minimize the distance between the predicted HOG features $\mathbf{\tilde{P}}_h$ and the original ones $\mathbf{\hat{P}}_h$.}
% Specifically, we extract HOG from each RGB channel for color information.
% Pixels per cell and cells per block are set to $(8,8)$ and $(1,1)$ respectively, the orientation number is 9. 

\noindent\textcolor{black}{\textbf{Low-Resolution RGB Pixels (LR-RGB).} 
RGB pixels are the most intuitive priors for inpainting. To address quadratically costly computation of transformer, ICT~\cite{wan2021high} leverages LR-RGB priors $\mathbf{\tilde{P}}_{rgb}$ ($32\times32$ in our implementation) to guide the inpainting. To further reduce the color space, all RGB combinations are clustered to $512$ discrete K-Means centers on the whole ImageNet as~\cite{chen2020generative}. Moreover, masked pixels are replaced with trainable tokens [\emph{MASK}]. Then generated $\mathbf{\tilde{P}}_{rgb}$ are bilinear upsampled to the stage-2 model. Thus stage-2 can be also seen as a super-resolution task, while stage-1 is working as LR inpainting. Since the color space is clustered to 512 discrete centers, the optimization can be seen as a classification. Therefore, the objective of $\mathbf{\tilde{P}}_{rgb}$ is to minimize the negative log-likelihood with cross-entropy (CE) for masked tokens.
}

\section{Deatils of E-NMS Fusion}

\begin{figure*}
\begin{centering}
\includegraphics[width=0.95\linewidth]{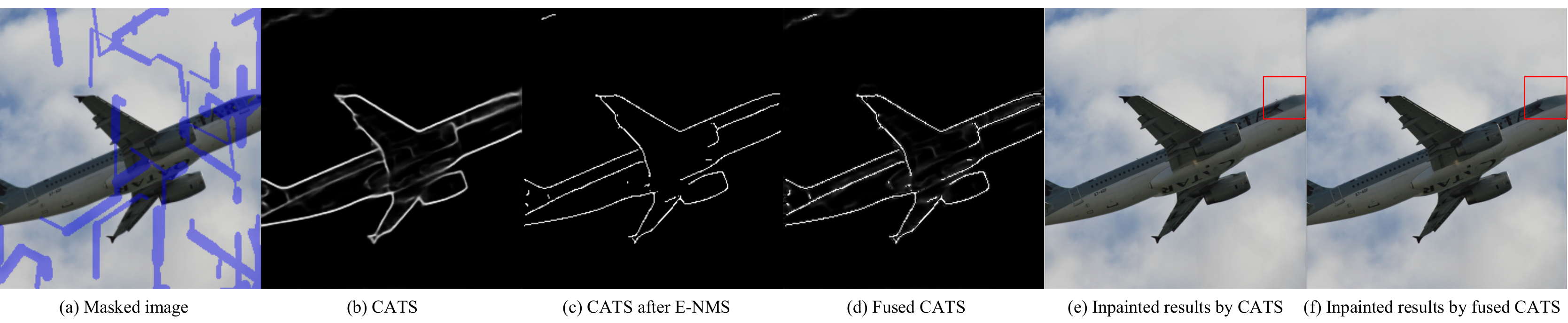}
\par\end{centering}
\caption{\textcolor{black}{The example of E-NMS fusion and corresponding inpainted results. Please zoom-in for details.}}
\label{fig:e-nms-display-images}
\end{figure*}

\textcolor{black}{Since the original CATS edge can be seen as a continuous probability image, which is blurred near the boundary; and the bilinear upsampling will further aggravate the blur, which eventually degrades the performance of image inpainting, as shown in Fig.~\ref{fig:e-nms-display-images}(b)(e). So we use E-NMS to work as a non-maximum suppression of the edge to preserve the sharp edge as in Fig.~\ref{fig:e-nms-display-images}(c). On the other hand, we find that E-NMS loses a lot of low-probability regions, which may contain informative clues for inpainting models. Therefore, we decide to make a trade-off between the continuous CATS and the E-NMS CATS. We keep areas as the original CATS with probability less than 0.25, while areas with probability larger than 0.25 are replaced with CATS after E-NMS. The fused edge map and the related inpainting result is shown in Fig.~\ref{fig:e-nms-display-images}(d) and Fig.~\ref{fig:e-nms-display-images}(f) respectively.}

\section{Detailed Network Settings}

We show some detailed network settings in Tab.~\ref{table:network_settings}. Besides, transformer block, Fast Fourier Convolution (FFC) block~\cite{suvorov2021resolution}, and the LKA inpainting block have been introduced in the main paper. The dilated resnet block is from the middle layer of~\cite{nazeri2019edgeconnect} with dilate=2. 

\begin{table*}
\caption{Model settings of Transformer Structure Restoration (TSR), Structure Feature Encoder (SFE), and Fourier CNN Texture Restoration (FTR). GC, BN mean Gated Convolution~\cite{yu2019free} and BatchNorm; TConv2d, TGC indicate Transposed Conv2d and GC.
\label{table:network_settings}}
\begin{centering}
\small
\begin{tabular}{c|c|c}
\hline 
Transformer Structure Restoration (TSR) & Structure Feature Encoder (SFE) & Fourier CNN Texture Restoration with LKA (FTR)\tabularnewline
\hline 
\hline 
Conv2d+ReLU($256\times256\times64$) & GC+BN+ReLU($256\times256\times64$) & Conv2d+LKA+FFN($256\times256\times64$)\tabularnewline
\hline 
Conv2d+ReLU($128\times128\times128$) & GC+BN+ReLU($128\times128\times128$) & Conv2d+LKA+FFN($128\times128\times128$)\tabularnewline
\hline 
Conv2d+ReLU($64\times64\times256$) & GC+BN+ReLU($64\times64\times256$) & Conv2d+LKA+FFN($64\times64\times256$)\tabularnewline
\hline 
Conv2d+ReLU($32\times32\times256$) & GC+BN+ReLU($32\times32\times512$) & Conv2d+LKA+FFN($32\times32\times512$)\tabularnewline
\hline 
TransformerBlock$\times8$ & DilatedResnetBlock$\times3$ & FFCBlock$\times9$\tabularnewline
\hline 
TConv2d+ReLU($64\times64\times256$) & TGC+BN+ReLU($64\times64\times256$) & TConv2d+LKA+FFN($64\times64\times256$)\tabularnewline
\hline 
TConv2d+ReLU($128\times128\times128$) & TGC+BN+ReLU($128\times128\times128$) & TConv2d+LKA+FFN($128\times128\times128$)\tabularnewline
\hline 
TConv2d+ReLU($256\times256\times64$) & TGC+BN+ReLU($256\times256\times64$) & TConv2d+LKA+FFN($256\times256\times64$)\tabularnewline
\hline 
Conv2d+Sigmoid($256\times256\times2$) & -- & Conv2d+Tanh($256\times256\times3$)\tabularnewline
\hline 
\end{tabular}
\par\end{centering}
\end{table*}

\section{Kanizsa Triangle Inpainting}

\begin{figure}[ht]
\begin{centering}
\includegraphics[width=0.95\linewidth]{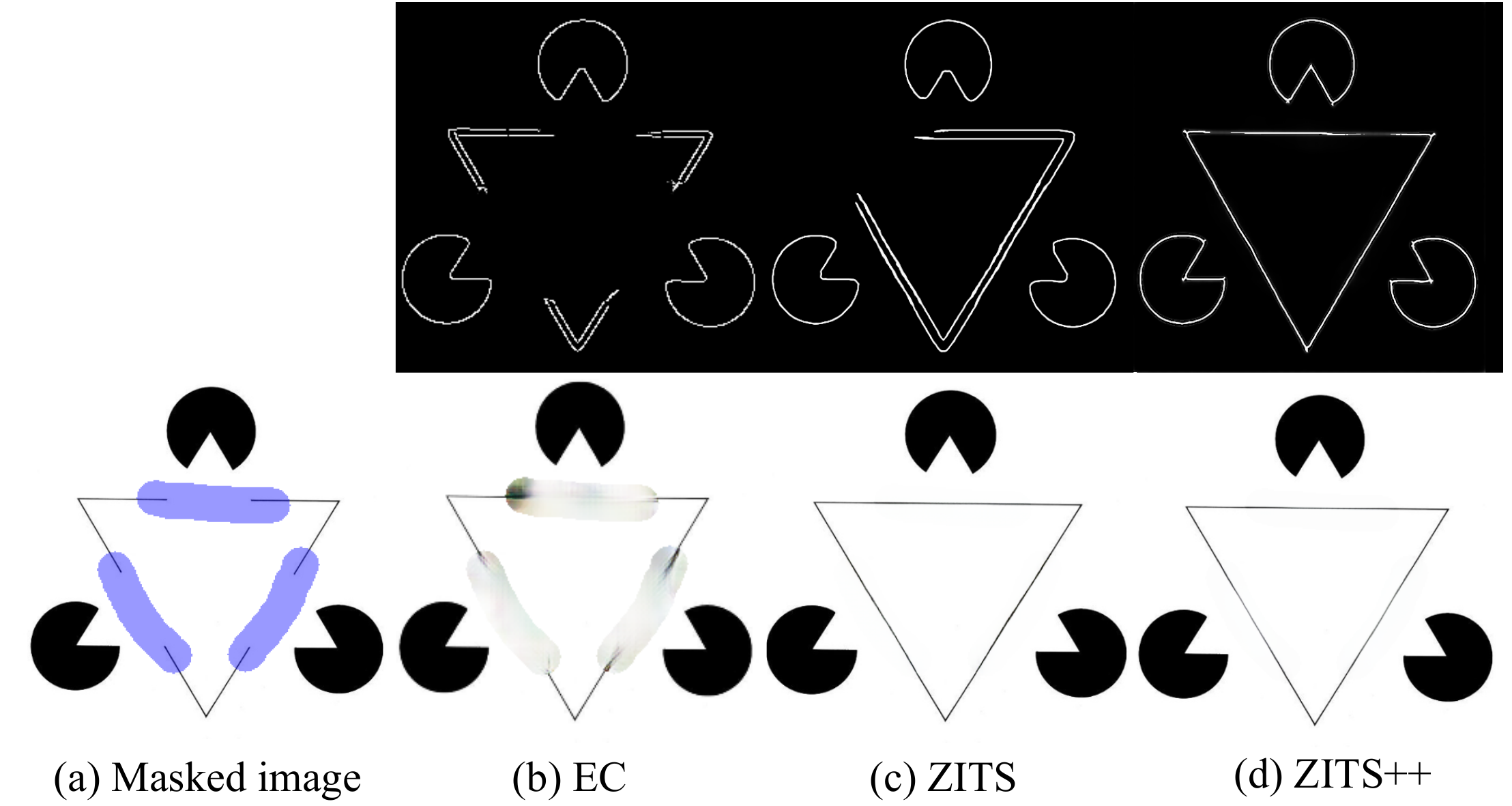}
\par\end{centering}
\vspace{-0.1in}
 \caption{Inpainted Kanizsa Triangle. First row is the inpainted canny edges (EC~\cite{nazeri2019edgeconnect} and ZITS~\cite{dong2022incremental}) and CATS (ZITS++).} \label{fig:triangle}
\vspace{-0.1in}
\end{figure}

We also conduct an interesting experiment on completing Kanizsa Triangle, which is utilized as the teaser figure of David Marr's book~\cite{marr2010vision}.
As contours in an outline drawing can convey meaning and play a crucial role in human vision, understanding and inpainting ideographic images are systematically tied to people's visual perception and cognitive processes. For instance, in Fig.~\ref{fig:triangle}(a), we can observe partially occluded disks and another triangle under a nonexistent white triangle in the Kanizsa triangle~\cite{stevens2012vision}. Benefited from transformer's better understand of global structure than CNN~\cite{wan2021high}, ZITS~\cite{dong2022incremental} and ZITS++ can inpaint more complete structure than CNN-based EC~\cite{nazeri2019edgeconnect} as in the first row of Fig.~\ref{fig:triangle}. The final inpainting results of our ZITS++ exchange the top/down positions of the black-sided triangle and the nonexistent white one in the Kanizsa triangle. It is worth noting that ZITS achieves a good inpainting result thanks to the completely inpainted lines rather than canny edges, while ZITS++ can address it with only L-Edges (CATS).

\section{More Training Details}

Training a model with dynamic resolutions of 256$\sim$512 reduces the training speed with frequent GPU memory swaps. Therefore, we train the model with regular resolutions, \emph{i.e.}, resizing images from 512 to 256 and then back to 512. For Indoor, there is one cycle for each epoch. For Places2, there are 64 cycles for each epoch. Such a local monotonic resizing makes the training smooth without missing diversity. And the dynamic resolution based training can effectively save the training cost compared with the training with a full 512 image size. Moreover, it benefits to learn relative position encoding for our proposed MPE as discussed in~\cite{xu2020positional}.

Our TSR can be trained in batch size 30 with 3 NVIDIA(R) Tesla(R) V100 16GB GPUs. 256$\times$256 based FTR and SFE can be trained in batch size 30 with 3 V100 GPUs. For the dynamic resolution based training, we use batch size 18 with 6 V100 GPUs. The ZeroRA based finetuning cost only about half a day and one day for 256$\times$256 and 256$\sim$512 resolutions respectively.

\section{Upsampling Iteratively with SSU}

\begin{figure*}
\begin{centering}
\includegraphics[width=0.99\linewidth]{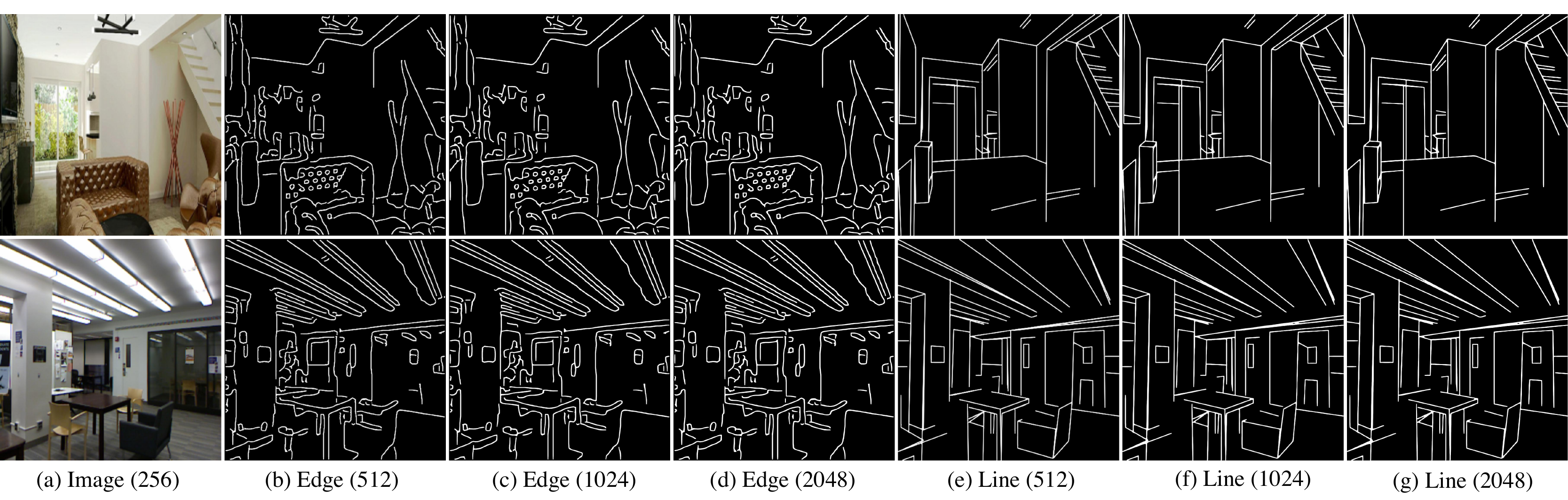}
\par\end{centering}
\caption{Iteratively outputs of SSU which have sizes from 512 to 2048. The results are consistent and robust.}
\label{fig:SSU_iterative}
\end{figure*}

Our Simple Structure Upsampler (SSU) introduced in Sec 3.2 can also work iteratively for larger image sizes. First, we should process the output edges $\mathbf{I}'_e$ and lines $\mathbf{I}'_l$ of SSU through shifted sigmoid as
\begin{equation}
\begin{split}
\mathbf{I}'_e&=\mathrm{sigmoid}[\gamma(\mathbf{I}'_e+\beta)],\\
\mathbf{I}'_l&=\mathrm{sigmoid}[\gamma(\mathbf{I}'_l+\beta)],
\end{split}
\label{eq:ssu_output}
\end{equation}
where $\gamma=2, \beta=2$ in our evaluation, and $\gamma,\beta$ are randomly selected from $[1.5, 3]$ for the finetuning. Since the output size of SSU is doubled, we can repeat the inputs $\mathbf{I}_e,\mathbf{I}_l\in\mathbb{R}^{h\times w}$ for $q$ times to achieve $\mathbf{I}'_e,\mathbf{I}'_l\in\mathbb{R}^{2^qh\times 2^qw}$. Then, the outputs can further be resized with the bilinear interpolation for the target size. In general, our SSU can get good and robust upsampled results for large sizes as shown in Fig.~\ref{fig:SSU_iterative}.

\section{Supplementary Experiments}
In this section, we provide some more qualitative and quantitative results to show the effects of our components. 
% Moreover, some details about the post-processing are also discussed.

\subsection{More Qualitative Results}

\begin{figure*}
\begin{centering}
\includegraphics[width=0.75\linewidth]{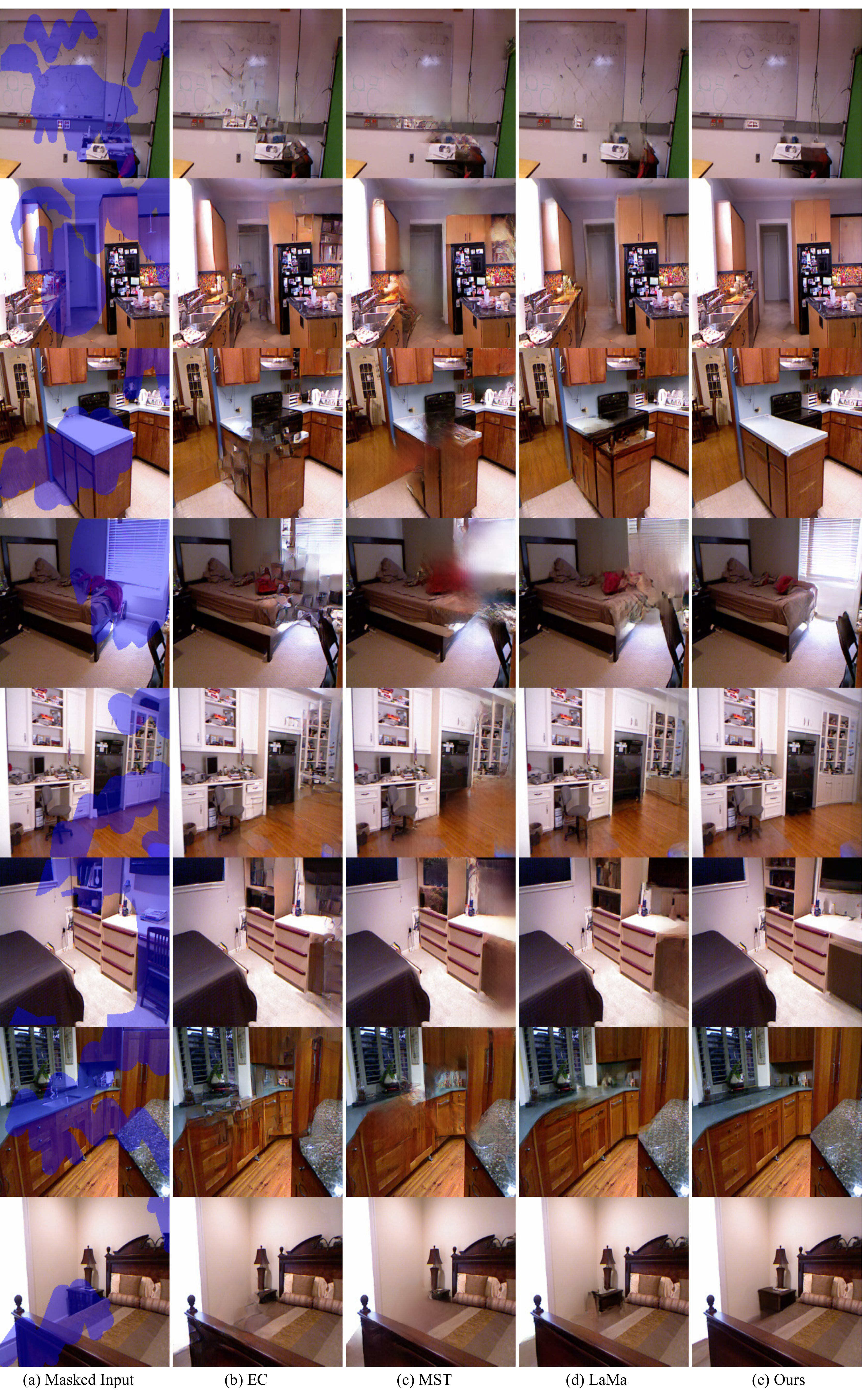}
\par\end{centering}
\caption{Qualitative 256$\times$256 results of Indoor dataset compared among EC~\cite{nazeri2019edgeconnect}, MST~\cite{cao2021learning}, LaMa~\cite{suvorov2021resolution}, and ours. Zoom-in for details.}
\label{fig:indoor_quali}
\end{figure*}

\begin{figure*}
\begin{centering}
\includegraphics[width=0.99\linewidth]{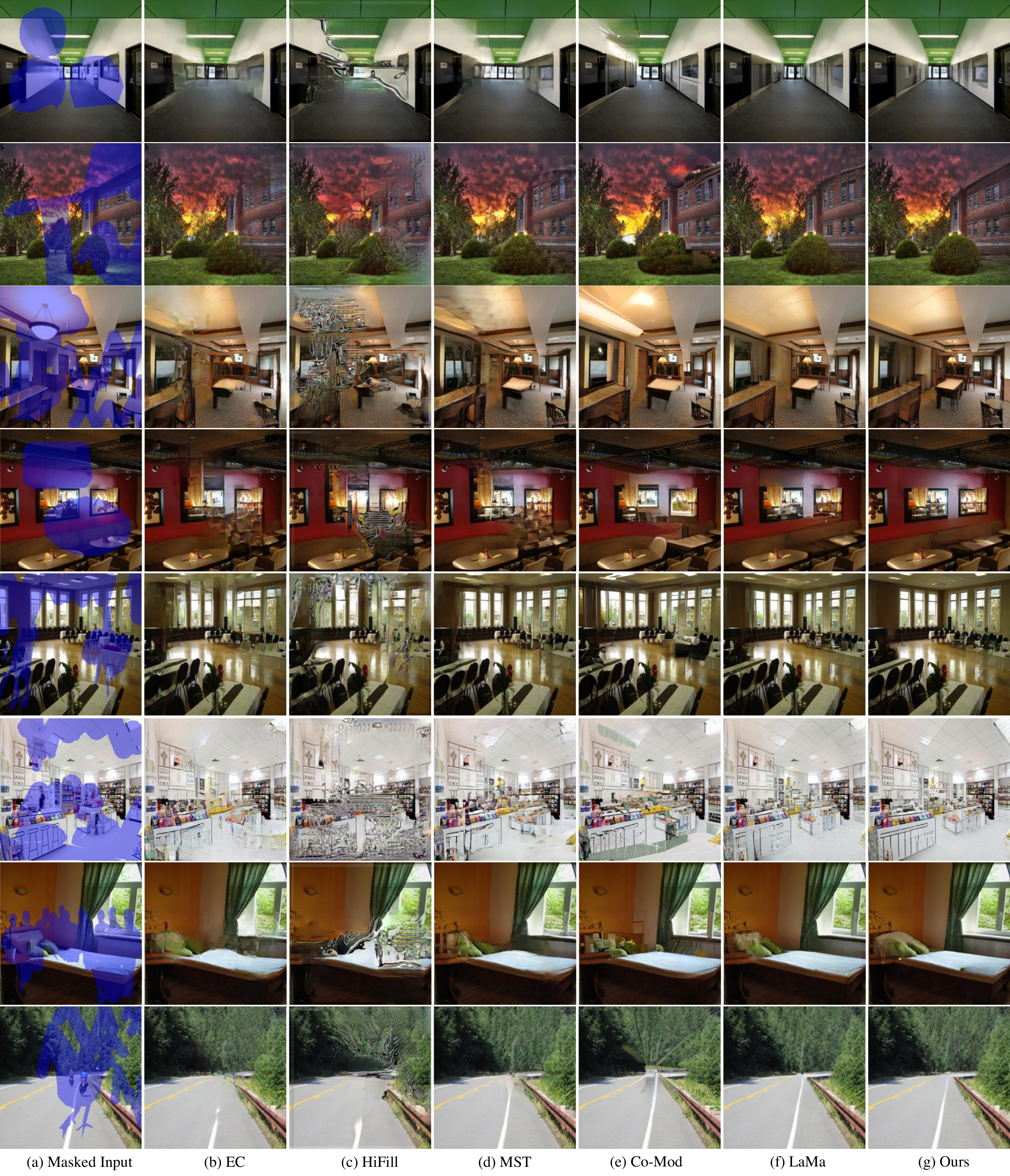}
\par\end{centering}
\caption{Qualitative 256$\times$256 results of Places2 dataset compared among EC~\cite{nazeri2019edgeconnect}, HiFill~\cite{yi2020contextual}, MST~\cite{cao2021learning}, Co-Mod~\cite{zhao2021large}, LaMa~\cite{suvorov2021resolution}, and ours. Zoom-in for details.}
\label{fig:places2_quali}
\end{figure*}

\begin{figure*}
\begin{centering}
\includegraphics[width=0.75\linewidth]{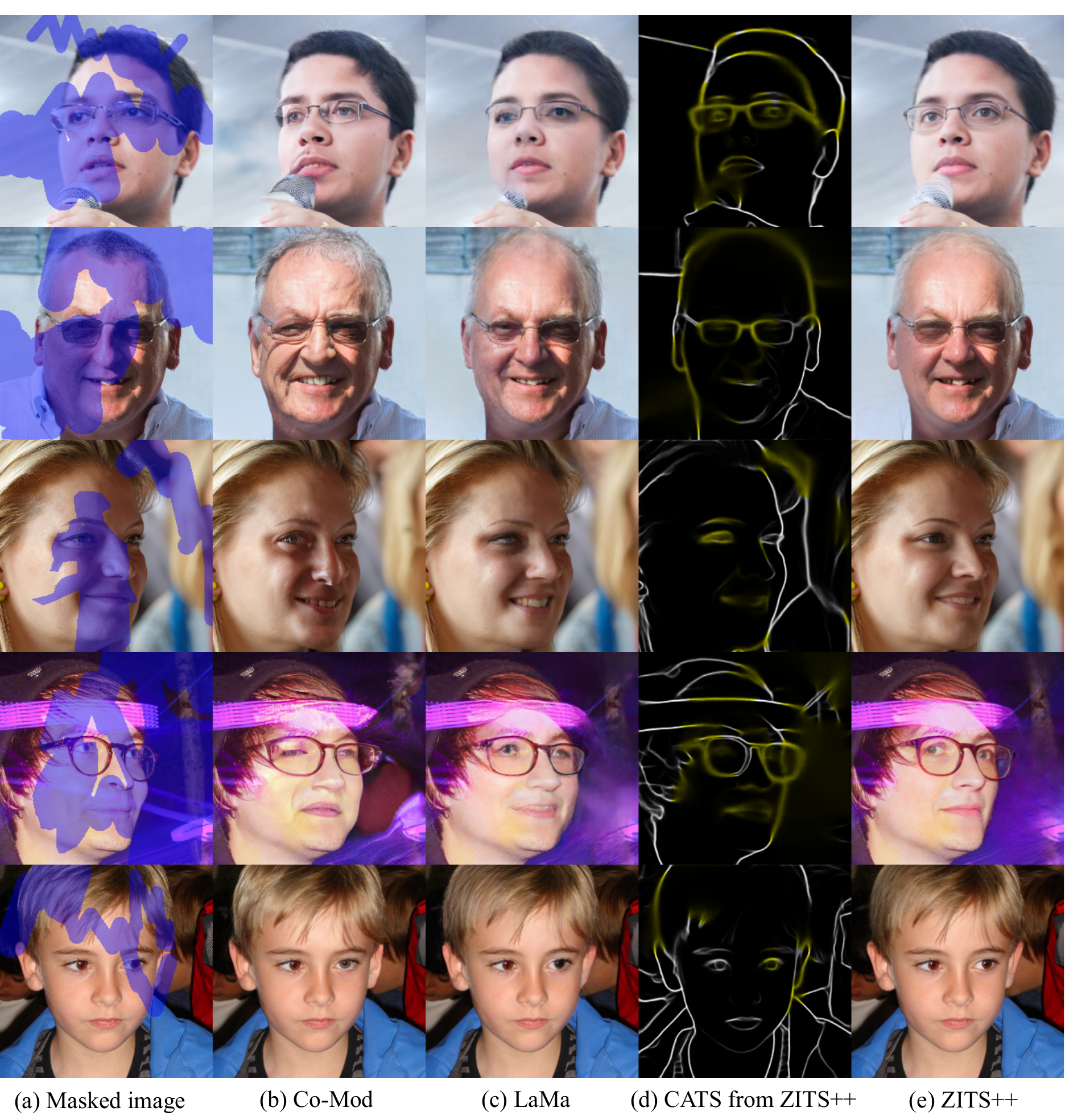}
\par\end{centering}
\caption{Qualitative 256$\times$256 results on FFHQ dataset compared among Co-Mod~\cite{zhao2021large}, LaMa~\cite{suvorov2021resolution}, and ours. Zoom-in for details.}
\label{fig:qualitative_ffhq_supp}
\end{figure*}

\begin{figure*}
\begin{centering}
\includegraphics[width=0.7\linewidth]{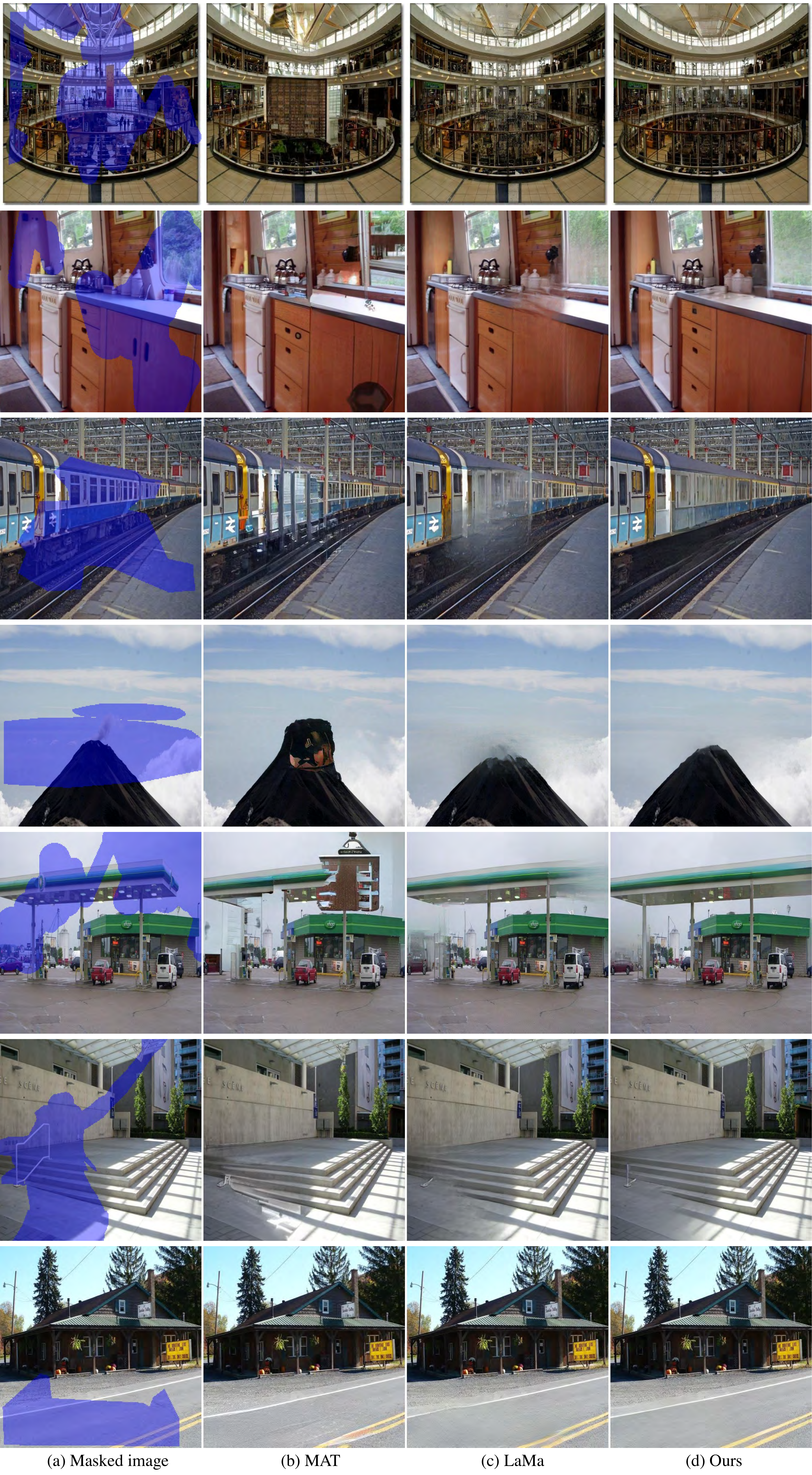}
\par\end{centering}
\caption{Qualitative 512$\times$512 results of Places2 dataset compared among MAT~\cite{li2022mat}, LaMa~\cite{suvorov2021resolution}, and ours. Zoom-in for details.}
\label{fig:places2_quali_512}
\end{figure*}

More qualitative 256$\times$256 results of Indoor, Places2, and FFHQ are shown in Fig.~\ref{fig:indoor_quali}, Fig.~\ref{fig:places2_quali}, and Fig.~\ref{fig:qualitative_ffhq_supp}. Qualitative 512$\times$512 Places2 results are shown in Fig.~\ref{fig:places2_quali_512}. Note that our method not only achieves better results in many man-made scenes, but also gets competitive results in natural scenes and face images benefited from MPE and edges.

\subsection{Quantitative Results with Different Masks}

\begin{table*}
\caption{Quantitative inpainting results on Indoor and Places2 with different mask ratios. \label{table:main_results}}
\vspace{-0.15in}
\begin{centering}
\small
\begin{tabular}{c|c|ccccc|ccccccc}
\hline
 &  & \multicolumn{5}{c|}{Indoor (256$\times$256)} & \multicolumn{7}{c}{Places2 (256$\times$256)}\tabularnewline
\hline
\hline
 & Mask & EC & MST & LaMa & ZITS & ZITS++ & EC & HiFill & Co-Mod & MST & LaMa & ZITS & ZITS++\tabularnewline
\hline
\multirow{5}{*}{PSNR$\uparrow$} & 10\textasciitilde 20\% & 28.18 & 28.72 & 29.05 & 29.87 & \textbf{30.39} & 26.60 & 24.04 & 26.40 & 28.13 & 28.23 & 28.31 & \textbf{29.15}\tabularnewline
 & 20\textasciitilde 30\% & 25.14 & 25.66 & 25.96 & 26.66 & \textbf{27.21} & 24.26 & 21.649 & 23.61 & 25.07 & 25.31 & 25.40 & \textbf{26.19}\tabularnewline
 & 30\textasciitilde 40\% & 23.02 & 23.53 & 23.87 & 24.64 & \textbf{25.16} & 22.59 & 19.96 & 21.73 & 23.07 & 23.43 & 23.51 & \textbf{24.23}\tabularnewline
 & 40\textasciitilde 50\% & 21.55 & 22.02 & 22.39 & 23.13 & \textbf{23.59} & 21.27 & 18.63 & 20.28 & 21.53 & 22.03 & 22.11 & \textbf{22.76}\tabularnewline
 & Mixed & 24.073 & 24.52 & 25.20 & 25.57 & \textbf{26.15} & 23.31 & 20.76 & 22.57 & 24.02 & 24.37 & 24.42 & \textbf{25.13}\tabularnewline
\hline
\multirow{5}{*}{SSIM$\text{\ensuremath{\uparrow}}$} & 10\textasciitilde 20\% & 0.951 & 0.954 & 0.956 & 0.961 & \textbf{0.965} & 0.913 & 0.883 & 0.926 & 0.941 & 0.942 & 0.942 & \textbf{0.950}\tabularnewline
 & 20\textasciitilde 30\% & 0.916 & 0.922 & 0.925 & 0.933 & \textbf{0.940} & 0.872 & 0.818 & 0.880 & 0.898 & 0.901 & 0.902 & \textbf{0.915}\tabularnewline
 & 30\textasciitilde 40\% & 0.876 & 0.886 & 0.890 & 0.901 & \textbf{0.912} & 0.828 & 0.751 & 0.831 & 0.852 & 0.859 & 0.860 & \textbf{0.877}\tabularnewline
 & 40\textasciitilde 50\% & 0.835 & 0.848 & 0.855 & 0.870 & \textbf{0.882} & 0.783 & 0.682 & 0.781 & 0.803 & 0.814 & 0.817 & \textbf{0.837}\tabularnewline
 & Mixed & 0.884 & 0.894 & 0.902 & 0.907 & \textbf{0.917} & 0.839 & 0.770 & 0.843 & 0.862 & 0.869 & 0.870 & \textbf{0.884}\tabularnewline
\hline
\multirow{5}{*}{FID$\downarrow$} & 10\textasciitilde 20\% & 9.56 & 8.56 & 8.01 & 7.18 & \textbf{6.62} & 1.95 & 4.71 & 0.52 & 0.76 & 0.45 & 0.43 & \textbf{0.36}\tabularnewline
 & 20\textasciitilde 30\% & 16.223 & 15.88 & 13.23 & 12.13 & \textbf{11.10} & 3.791 & 11.93 & 1.00 & 1.86 & 0.95 & 0.88 & \textbf{0.72}\tabularnewline
 & 30\textasciitilde 40\% & 23.48 & 22.69 & 18.77 & 16.51 & \textbf{15.61} & 6.98 & 25.16 & 1.64 & 3.83 & 1.72 & 1.55 & \textbf{1.24}\tabularnewline
 & 40\textasciitilde 50\% & 31.16 & 31.06 & 23.47 & 20.87 & \textbf{19.60} & 11.49 & 44.68 & 2.38 & 6.80 & 2.81 & 2.51 & \textbf{2.01}\tabularnewline
 & Mixed & 22.02 & 21.65 & 16.97 & 15.93 & \textbf{14.61} & 6.21 & 21.33 & 1.49 & 3.53 & 1.63 & 1.47 & \textbf{1.19}\tabularnewline
\hline
\multirow{5}{*}{LPIPS$\downarrow$} & 10\textasciitilde 20\% & 0.054 & 0.050 & 0.044 & 0.038 & \textbf{0.036} & 0.073 & 0.119 & 0.053 & 0.047 & 0.047 & 0.042 & \textbf{0.038}\tabularnewline
 & 20\textasciitilde 30\% & 0.094 & 0.087 & 0.078 & 0.068 & \textbf{0.062} & 0.111 & 0.189 & 0.098 & 0.082 & 0.083 & 0.073 & \textbf{0.066}\tabularnewline
 & 30\textasciitilde 40\% & 0.140 & 0.129 & 0.117 & 0.101 & \textbf{0.093} & 0.152 & 0.265 & 0.140 & 0.120 & 0.121 & 0.107 & \textbf{0.097}\tabularnewline
 & 40\textasciitilde 50\% & 0.189 & 0.172 & 0.156 & 0.136 & \textbf{0.126} & 0.194 & 0.343 & 0.184 & 0.160 & 0.161 & 0.143 & \textbf{0.131}\tabularnewline
 & Mixed & 0.135 & 0.122 & 0.112 & 0.098 & \textbf{0.090} & 0.149 & 0.137 & 0.246 & 0.122 & 0.155 & 0.108 & \textbf{0.094}\tabularnewline
\hline
\end{tabular}
\par\end{centering}
\vspace{-0.1in}
\end{table*}

We show more quantitative results with different masking rates from 10\% to 50\% and mixture of segmentation and irregular masks in Tab.~\ref{table:main_results}.

\subsection{More Structural Experiments}

\noindent\textbf{TSR Ablations.} 
For the Indoor dataset, we conducted several ablation experiments on our Transformer Structure Restoration (TSR), and the results are displayed in Tab.~\ref{table:axial_memory_speed} and  Tab.~\ref{table:abla_TSR}. As illustrated in Tab.~\ref{table:axial_memory_speed} and the first two rows of Tab.~\ref{table:abla_TSR}, replacing one standard self-attention module~\cite{vaswani2017attention} with an axial attention module~\cite{ho2019axial} in our Transformer Block can greatly reduce the GPU memory usage and speed up the model inference while keeping all metrics basically unchanged. Furthermore, we add the relative position encoding (RPE)~\cite{raffel2019exploring} into our axial attention module, which can boost our results. Note that the RPE must be incorporated with the axial attention module in row-wise and column-wise, while standard attention based RPE costs much more GPU memory due to the long sequence. On the other hand, as we think that a higher recall will benefit the later image inpainting, we further multiply the line logits by 4 before feeding it through the sigmoid activation function in all the experiments. This strategy enhances recall while only compromising a little precision.

\begin{table}
\small
\caption{Efficient ablations of axial attention module. FPS is the Frames Per Second during the inference. The GPU memory is test on single Tesla V100 with batch size 8.\label{table:axial_memory_speed}}
\begin{centering}
\begin{tabular}{ccc}
\hline
 & FPS & GPU Memory (MB)\tabularnewline
\hline
\hline
w./o. Axial & 6.41 & 14845\tabularnewline
with Axial & \textbf{7.89} & \textbf{10547}\tabularnewline
\hline
\end{tabular}
\par\end{centering}
\end{table}

\begin{table*}
\small
\caption{Ablation studies of TSR on the Indoor dataset, where P., R., F1 mean Precision, Recall, and F1-score.\label{table:abla_TSR}}
\begin{centering}
\begin{tabular}{cc|cccccc|c}
\hline
 &  & \multicolumn{3}{c|}{Edge} & \multicolumn{3}{c|}{Line} & Avg\tabularnewline
Axial & RPE & P. & R. & \multicolumn{1}{c|}{F1} & P. & R. & F1 & F1\tabularnewline
\hline
\hline
 &  & 38.27 & 33.12 & 34.78 & 52.93 & 65.79 & 57.73 & 46.26\tabularnewline
\CheckmarkBold{} &  & \textbf{38.30} & 32.90 & 34.64 & 52.74 & \textbf{66.48} & 57.87 & 46.26\tabularnewline
\CheckmarkBold{} & \CheckmarkBold{} & 37.34 & \textbf{34.25} & \textbf{35.10} & \textbf{53.60} & 66.23 & \textbf{58.35} & \textbf{46.72}\tabularnewline
\hline
\end{tabular}
\par\end{centering}
\end{table*}

\subsection{Effects of ZeroRA}
\begin{figure}
\begin{centering}
\includegraphics[width=1.0\linewidth]{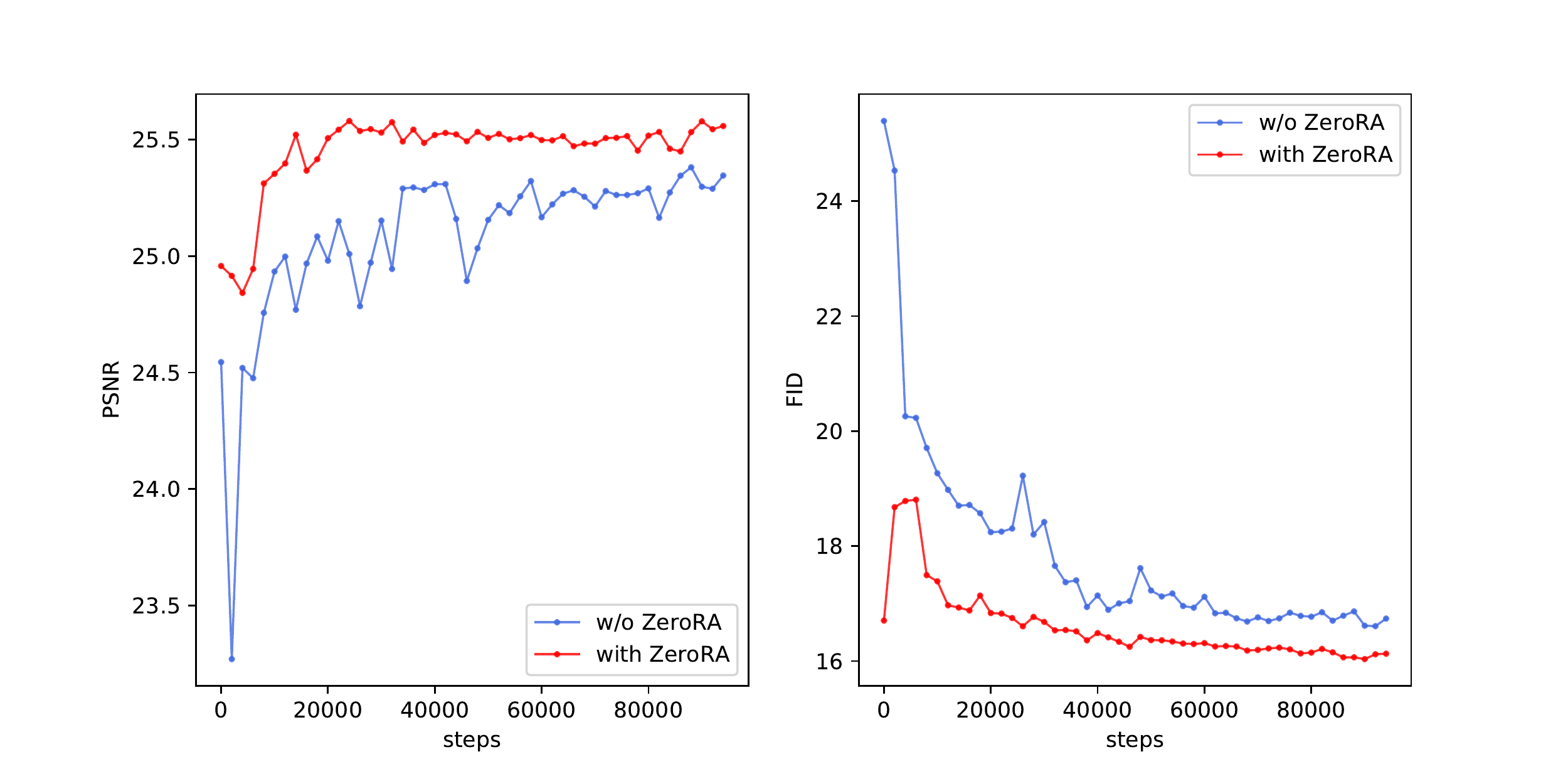}
\par\end{centering}
 \caption{Structure enhanced finetuning with and without ZeroRA.
 \label{fig:zerora_abla}}
\end{figure}

We also show line charts of PSNR and FID during the finetuning in Fig.~\ref{fig:zerora_abla} with and without ZeroRA. The blue curve without ZeroRA is unstable at the beginning of the finetuning, while the red one with ZeroRA enjoys better convergence and stability. Because adding extra structural features without ZeroRA leads to dramatic output changing, which harms the vulnerable GAN training.

\subsection{User Study}

\begin{figure}
\centering
\includegraphics[width=0.95\linewidth]{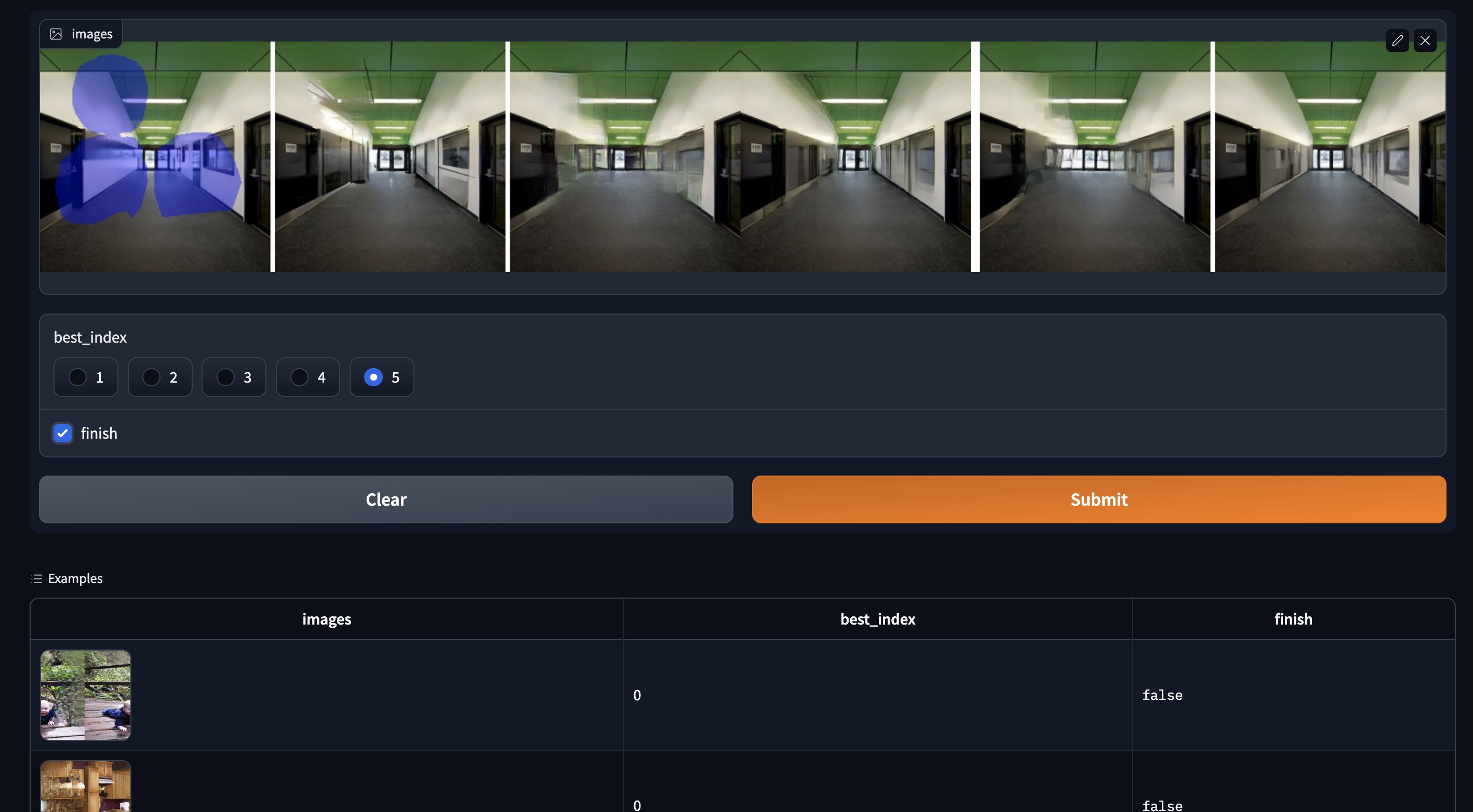}
\vspace{-0.1in}
\caption{Our concise user study interface built by Gradio~\cite{abid2019gradio}.\label{fig:user_study_IDE}}
\vspace{-0.1in}
\end{figure}

\begin{figure}
\begin{centering}
\includegraphics[width=0.99\linewidth]{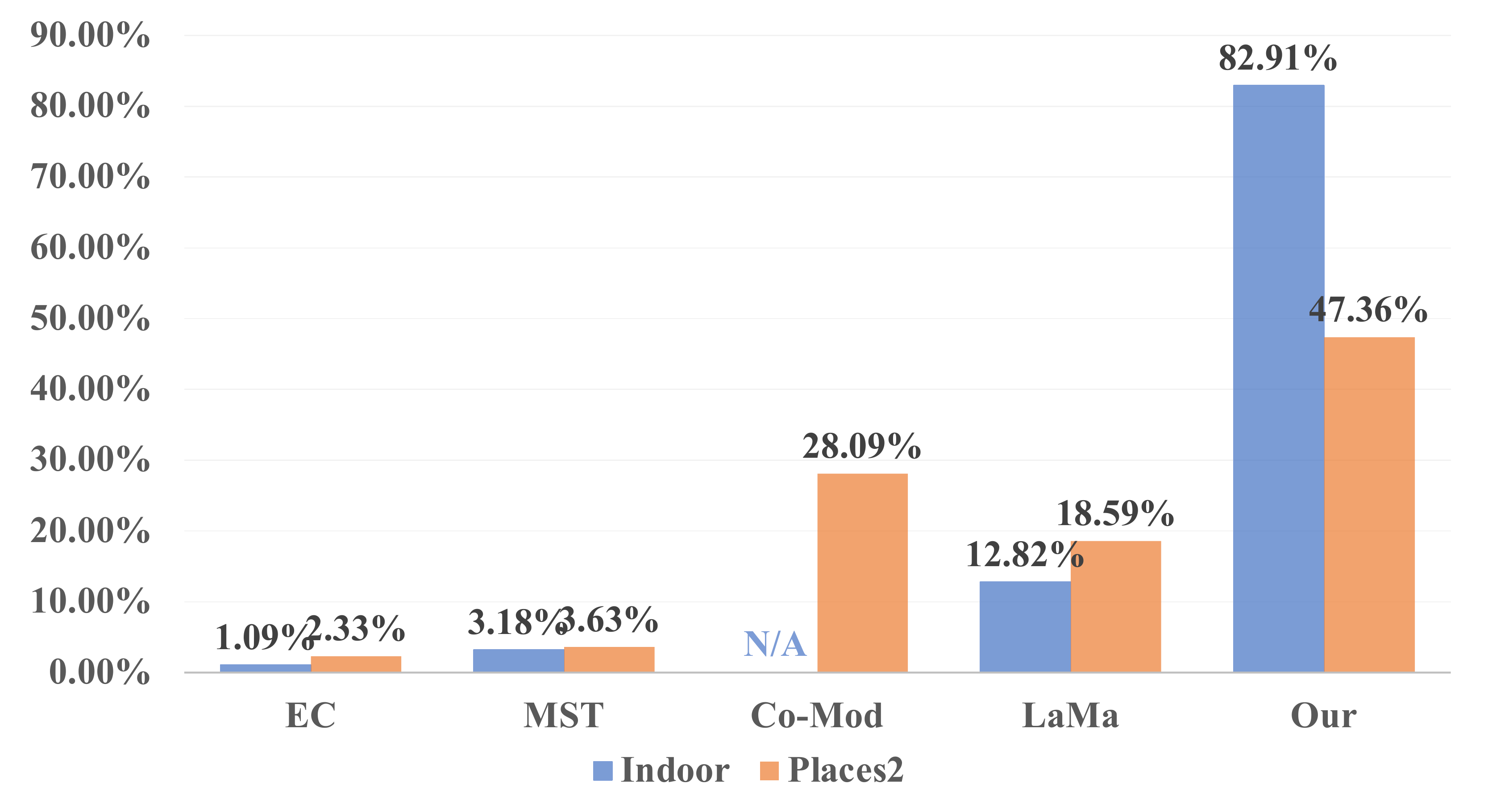}
\par\end{centering}
\caption{Average scores of Indoor and Places2 for user studies, which are
collected from volunteers who select the best one from shuffled inpainted images.}
\label{fig:user_study}
\end{figure}

We conduct user studies on several models to validate the effectiveness of our model from the perspective of humans. Specifically, we invite 10 volunteers who are not familiar with image inpainting to judge the quality of inpainted images.  On Indoor and Places2, five methods are compared, which including EC~\cite{nazeri2019edgeconnect}, MST~\cite{cao2021learning} LaMa~\cite{suvorov2021resolution}, Co-Mod~\cite{zhao2021large} and ours. Given the masked inputs, we randomly shuffle and combine the results of five methods together. Then, volunteers are required to choose the best one from each group. 
To improve the experience and quality of the user study, a Gradio-based~\cite{abid2019gradio} interface shown in Fig.~\ref{fig:user_study_IDE} is built.
As shown in Fig.~\ref{fig:user_study}, our method outperforms the other three competitors on both two datasets. Especially, our method can achieve a great advantage compared with the baseline method \emph{i.e.}, LaMa.

\subsection{Results of Rectangular Masks}

\begin{figure}
\begin{centering}
\includegraphics[width=0.99\linewidth]{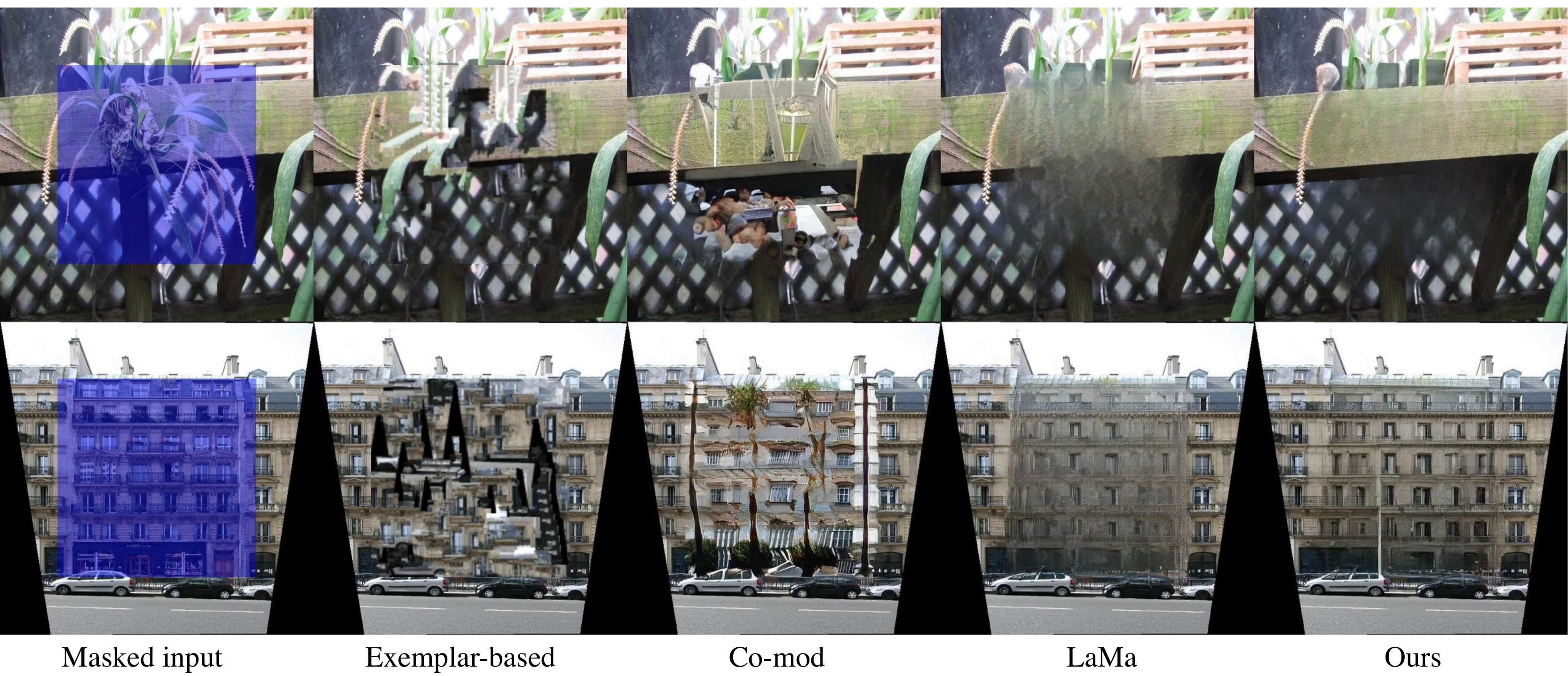}
\par\end{centering}
\caption{Inpainting results of 512$\times$512 images compared with Exemplar-based inpainting~\cite{criminisi2003object}, Co-Mod, LaMa, and ours.}
\label{fig:rec}
\end{figure}

Here we provide some results of 40\% center rectangular masks of 1k Places(512) images without any retraining in Tab.~\ref{table:rectangular_mask}. Note that Co-Mod~\cite{zhao2021large} is the only one trained with some rectangular masks while other methods have not been trained with similar masks. Moreover, we compare related qualitative results in Fig.~\ref{fig:rec}. And the classical exemplar-based inpainting~\cite{criminisi2003object} is also included. Traditional exemplar-based method fails to work properly and is time-consuming. Co-Mod has hallucinated artifacts instead of generating plausible results. And LaMa results are blur with still high PSNR.

\begin{table}
\small
\caption{Quantitative results on 1k Places 512 images with 40\% center rectangular masks.\label{table:rectangular_mask}}
\begin{centering}
\begin{tabular}{cccc}
\hline 
 & {PSNR} & {FID} & {LPIPS}\tabularnewline
\hline 
{Co-Mod} & {17.59} & \textbf{52.38} & {0.262}\tabularnewline
{LaMa} & \textbf{19.69} & {61.67} & {0.268}\tabularnewline
{Ours} & {19.65} & {55.85} & \textbf{0.239}\tabularnewline
\hline 
\end{tabular}
\par\end{centering}
\end{table}

\subsection{Comparisons of Texture Images}

\begin{table}
\small
\caption{Quantitative results on 512 texture images from~\cite{cimpoi2014describing}.\label{table:texture}}
\begin{centering}
\begin{tabular}{ccc}
\hline 
 & {LaMa} & {Ours}\tabularnewline
\hline 
{PSNR} & \textbf{25.82} & {25.67}\tabularnewline
{SSIM} & \textbf{0.875} & {0.869}\tabularnewline
{FID} & {12.86} & \textbf{11.67}\tabularnewline
{LPIPS} & {0.138} & \textbf{0.134}\tabularnewline
\hline 
\end{tabular}
\par\end{centering}
\end{table}

\begin{figure}
\begin{centering}
\includegraphics[width=0.99\linewidth]{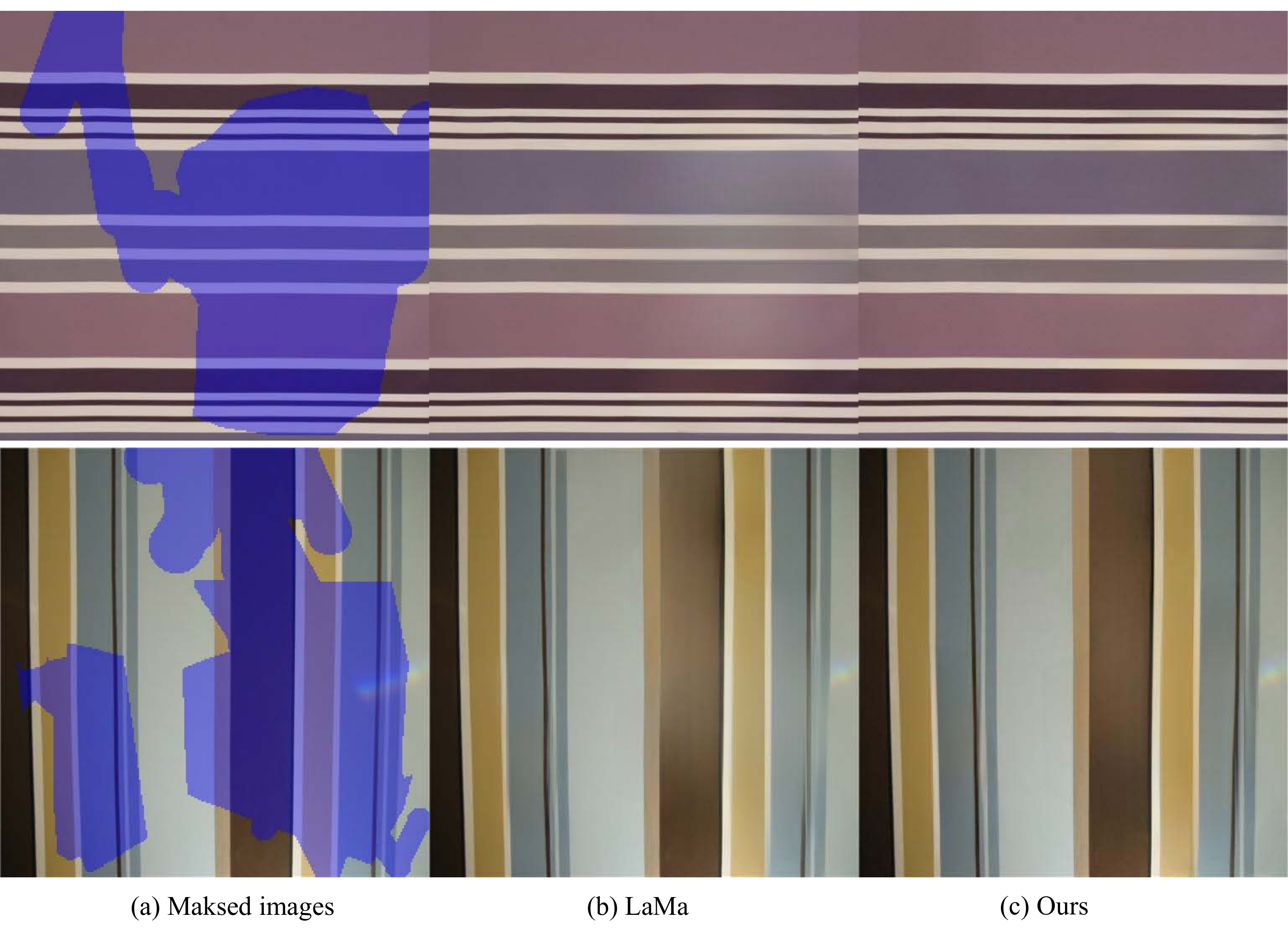}
\par\end{centering}
\caption{Inpainting results of 512$\times$512 texture images~\cite{cimpoi2014describing} compared with LaMa and ours.}
\label{fig:texture}
\end{figure}

We further compare our method with LaMa on 1,880 texture images~\cite{cimpoi2014describing} in Tab.~\ref{table:texture} and Fig.~\ref{fig:texture}, which contain strong periodic textures. Since this dataset is very suitable to LaMa~\cite{suvorov2021resolution}, our method still has competitive performance.

\begin{figure}
\begin{centering}
\includegraphics[width=0.95\linewidth]{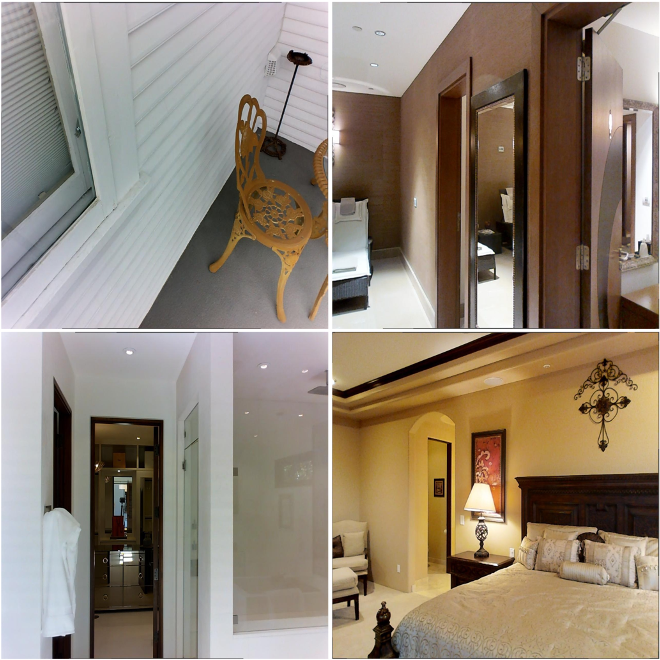}
\par\end{centering}
\caption{Examples of resized 1024$\times$1024 MatterPort3D images.}
\label{fig:matterport3d}
\end{figure}

\begin{figure}
\begin{centering}
\includegraphics[width=0.95\linewidth]{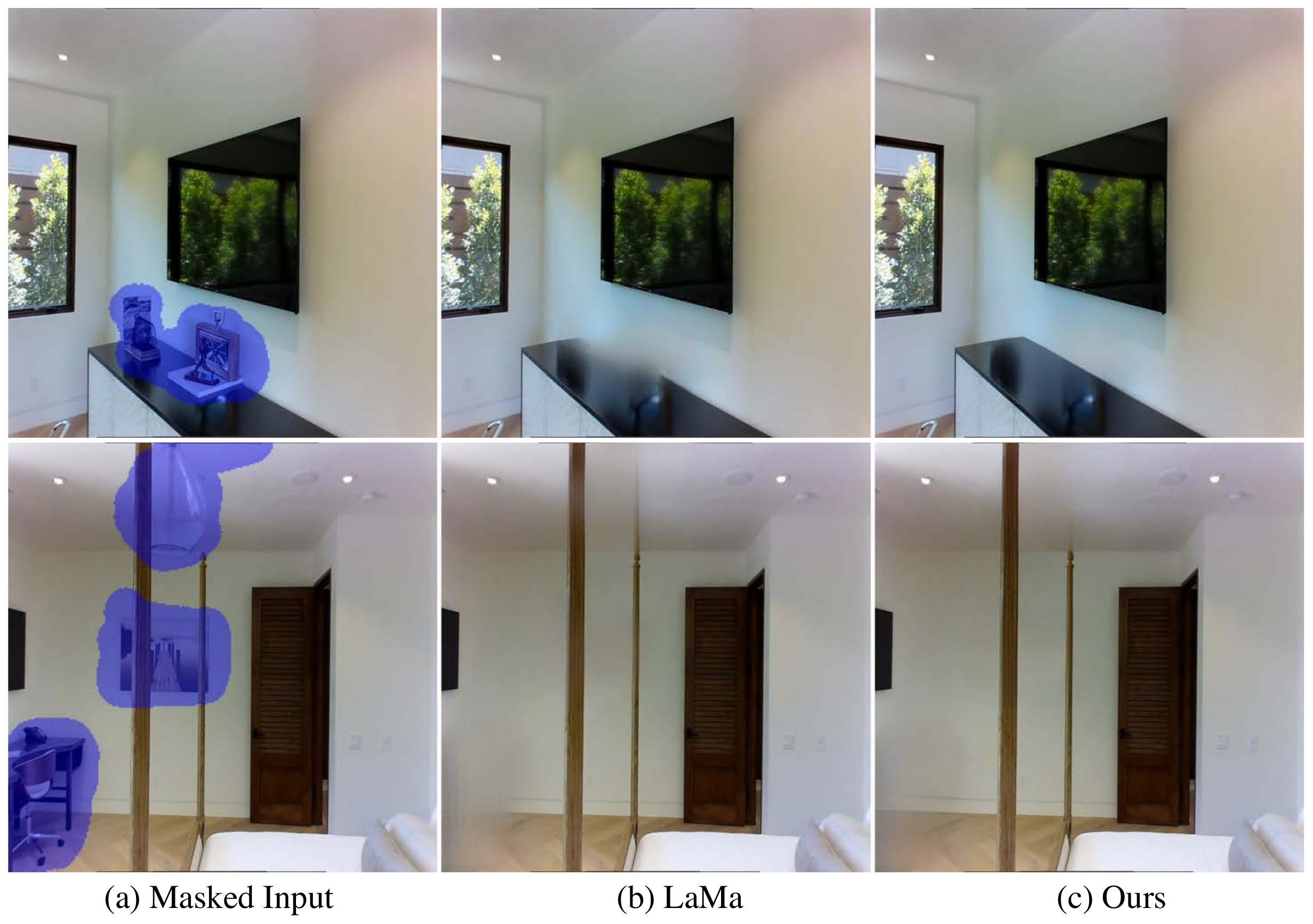}
\par\end{centering}
\caption{Inpainting results of LaMa~\cite{suvorov2021resolution} and ours in 1024$\times$1024 MatterPort3D images.}
\label{fig:matterport3d_res}
\end{figure}

\subsection{Results of MatterPort3D}
We use the test set of MatterPort3D~\cite{chang2017matterport3d} to evaluate the effectiveness of our method in the high-resolution structure recovery. MatterPort3D images tested in this paper consist of 1,965 indoor images in 1280$\times$1024. We resized them into 1024$\times$1024 as shown in Fig.~\ref{fig:matterport3d}. We provide some qualitative results of our method and LaMa compared on MatterPort3D in Fig.~\ref{fig:matterport3d_res}. For these structural images, our results enjoy better structures.

\subsection{Compare with the inpainting of PhotoShop}

\begin{figure}
\begin{centering}
\includegraphics[width=0.95\linewidth]{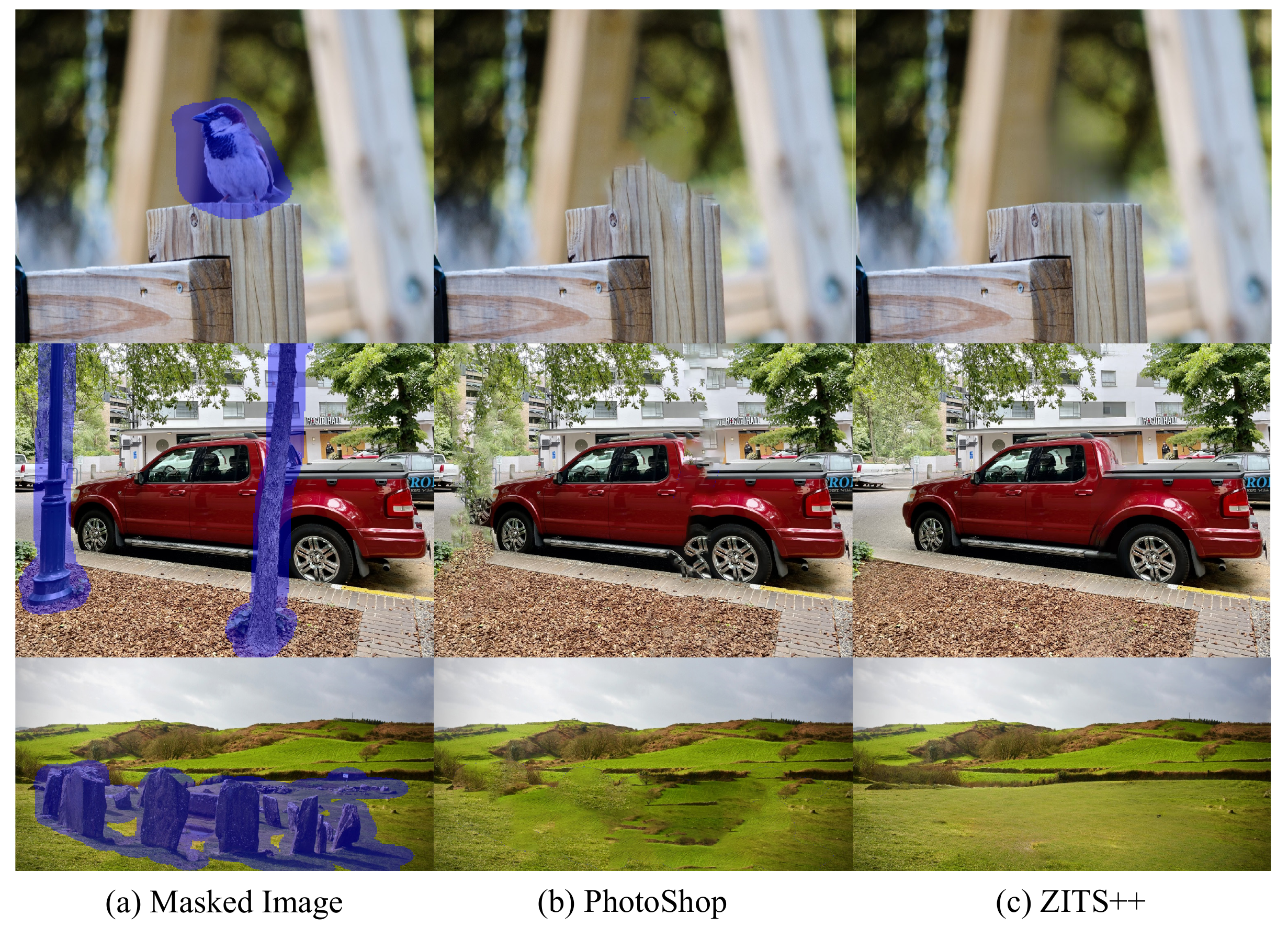}
\par\end{centering}
% \vspace{-0.1in}
\caption{Inpainting results from PhotoShop2023 and ZITS++ of high-resolution images.}
% \vspace{-0.1in}
\label{fig:compare_ps}
\end{figure}

To further explore the effectiveness of our proposed method. We compare our ZITS++ with the content-aware fill tool of a famous commercial software--PhotoShop2023 in Fig.~\ref{fig:compare_ps}. The reference-guided regions of the PhotoShop's filling tool are automatically decided. 
Note that PhotoShop fails to preserve structures (row 1); it also suffers from complex environments (row 2). Although both PhotoShop and ZITS++ could work well in nature scenes (row 3), ZITS++'s result is more harmonious.

\section{More High-Resolution Results}

\begin{figure*}
\begin{centering}
\includegraphics[width=0.95\linewidth]{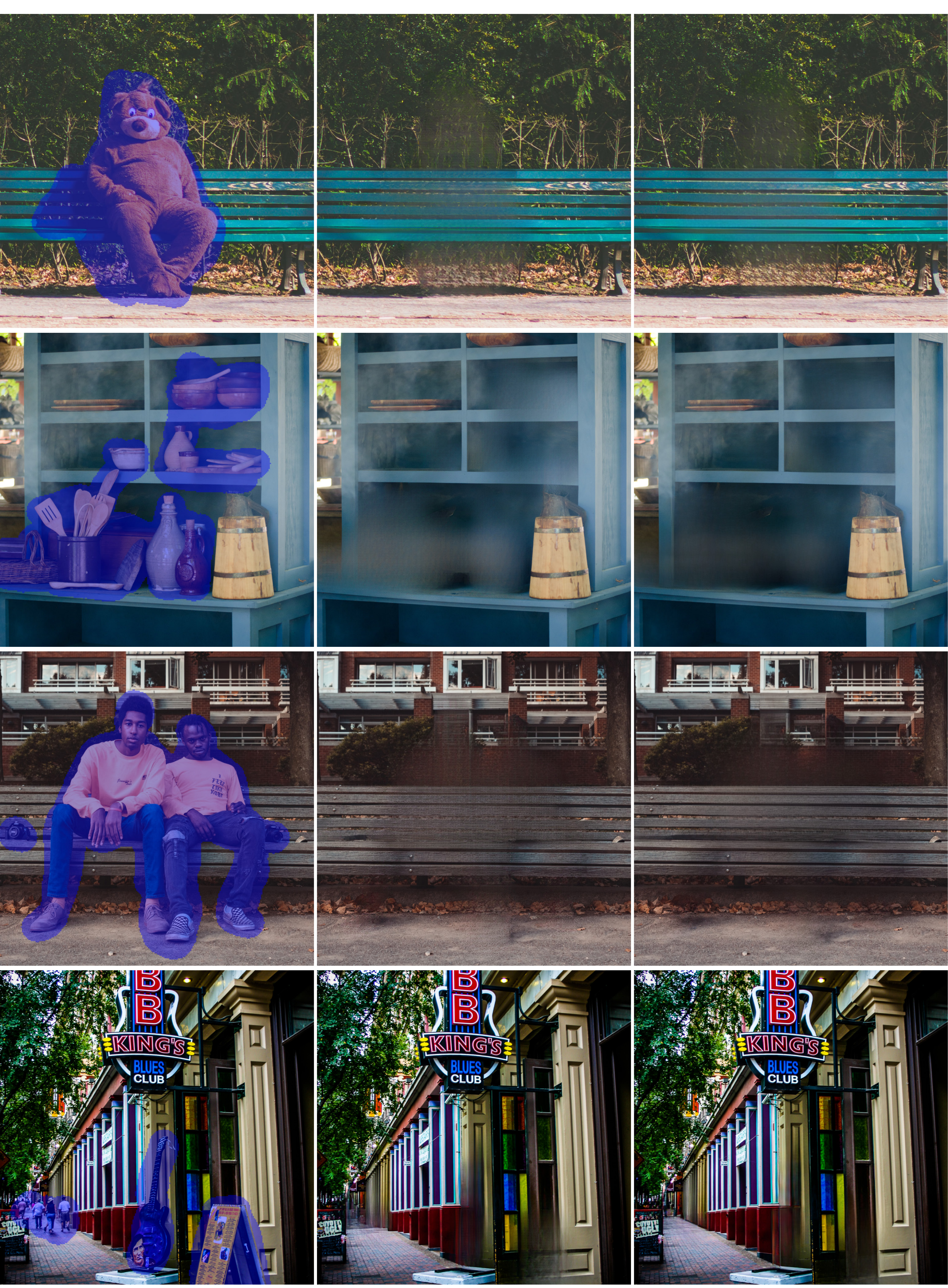}
\par\end{centering}
\caption{High-resolution(1K) object removal results. From left to right are masked inputs, results from LaMa~\cite{suvorov2021resolution}, and results from our method. Please zoom-in for more details.}
\label{fig:high_res1}
\end{figure*}

\begin{figure*}
\begin{centering}
\includegraphics[width=0.95\linewidth]{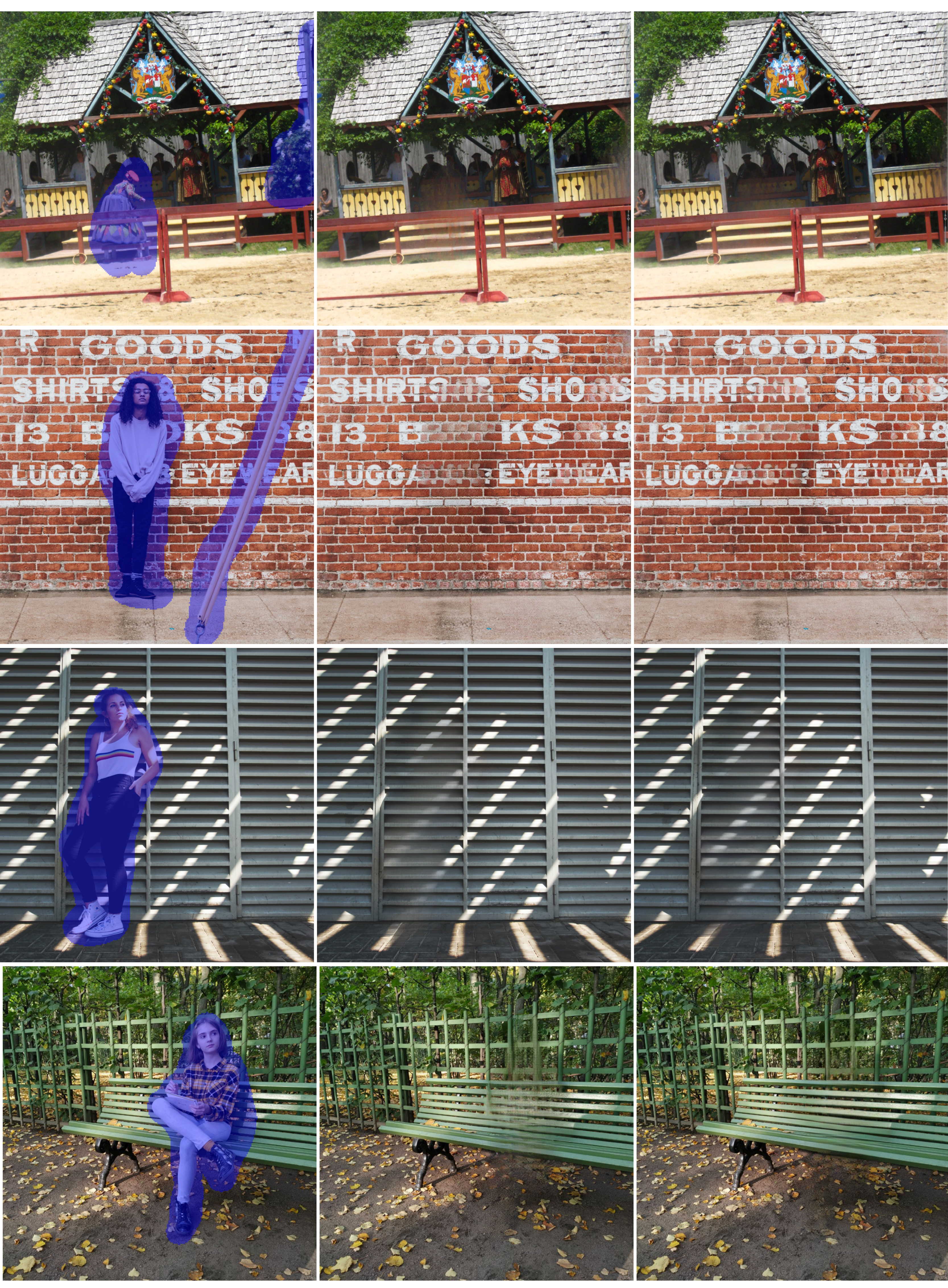}
\par\end{centering}
\caption{High-resolution(1K) object removal results. From left to right are masked inputs, results from LaMa~\cite{suvorov2021resolution}, and results from our method. Please zoom-in for more details.}
\label{fig:high_res2}
\end{figure*}

\begin{figure*}
\begin{centering}
\includegraphics[width=0.95\linewidth]{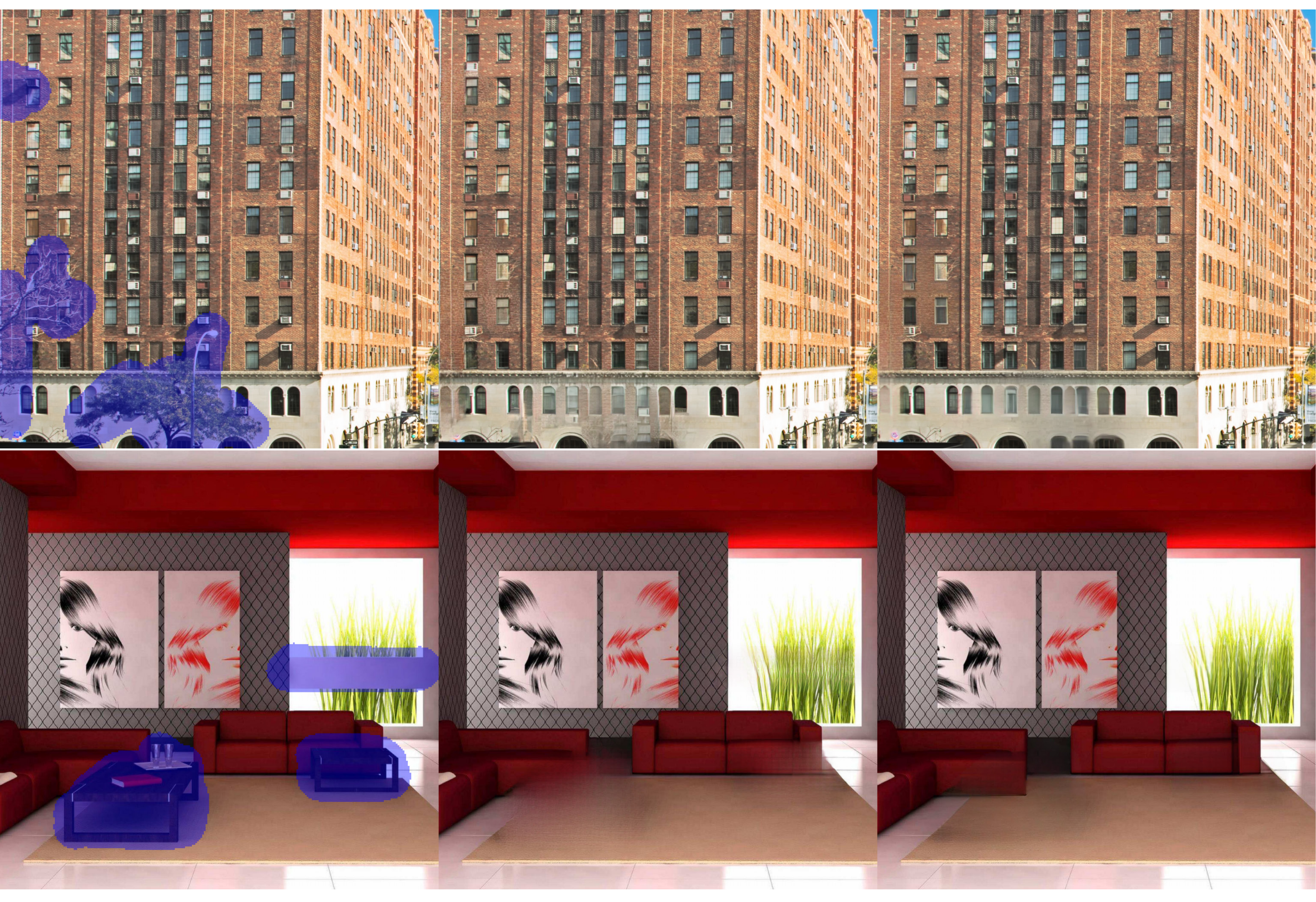}
\par\end{centering}
\caption{\textcolor{black}{More high-resolution(1K) object removal results from LaMa~\cite{suvorov2021resolution} examples. From left to right are masked inputs, results from LaMa, and results from our method. Please zoom-in for more details.}}
\label{fig:high_res4}
\end{figure*}

\begin{figure*}
\begin{centering}
\includegraphics[width=0.95\linewidth]{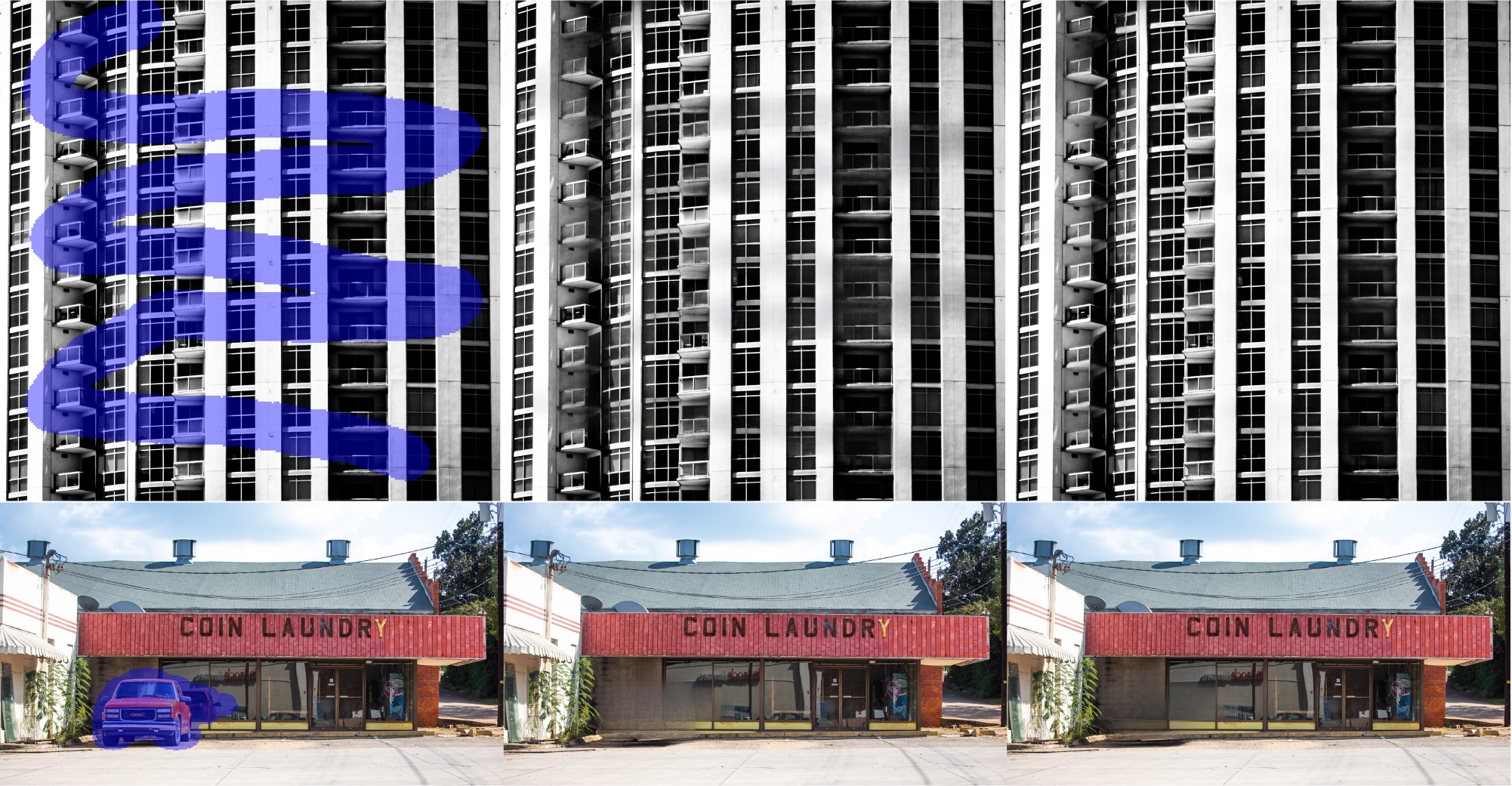}
\par\end{centering}
\caption{The high-resolution inpainting comparison of \textcolor{black}{2K examples from our HR-Flickr}. From left to right are the masked input, the result from LaMa~\cite{suvorov2021resolution}, and the result from our method. Please zoom-in for more details.}
\label{fig:high_res3}
\end{figure*}

\begin{figure*}
\begin{centering}
\includegraphics[width=0.95\linewidth]{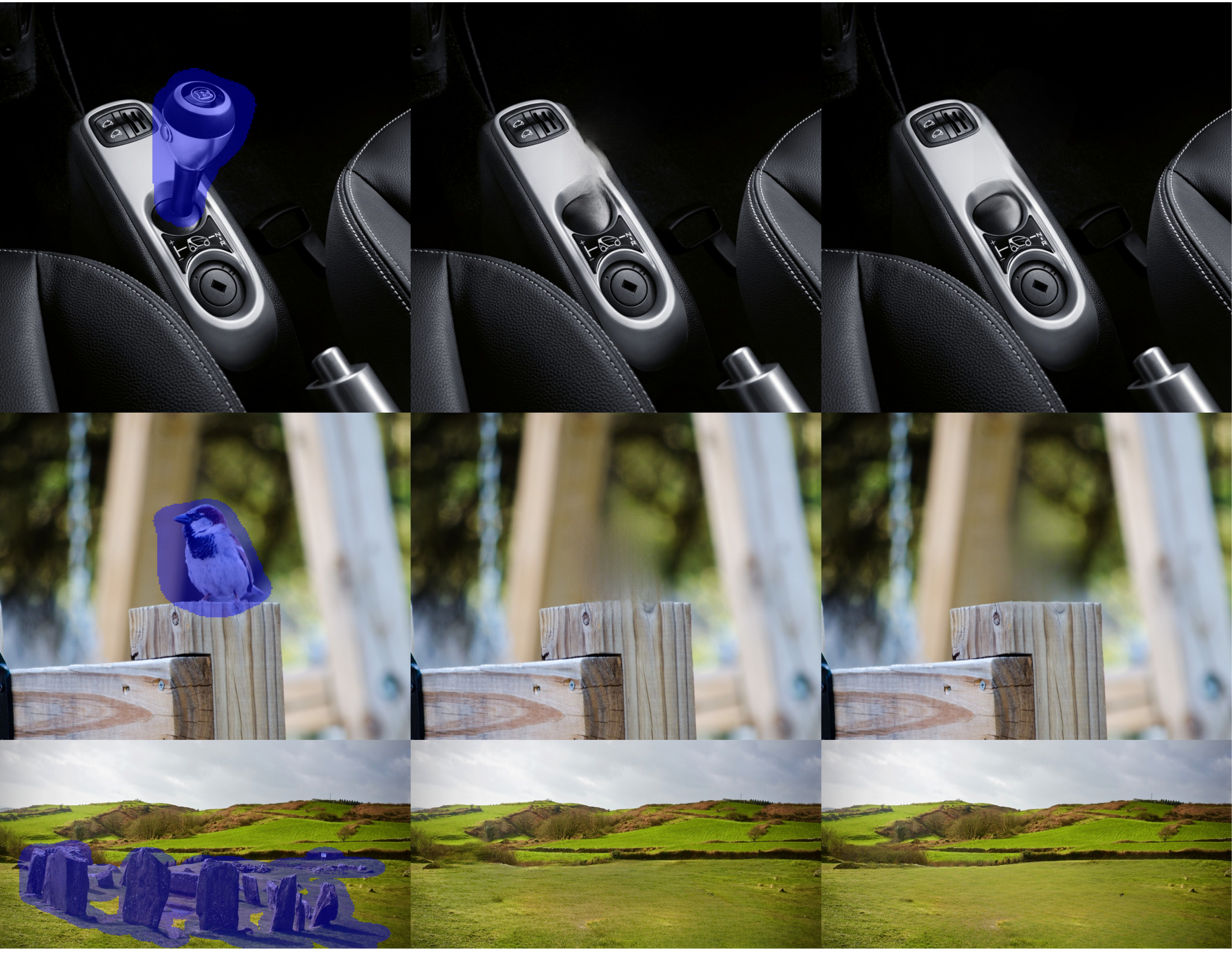}
\par\end{centering}
\caption{\textcolor{black}{More high-resolution 1K object removal results from HR-Flickr. From left to right are masked inputs, results from LaMa~\cite{suvorov2021resolution}, and results from our method. Please zoom-in for more details.}}
\label{fig:high_res5}
\end{figure*}

In Fig.~\ref{fig:high_res1}, Fig.~\ref{fig:high_res2}, Fig.~\ref{fig:high_res3}, Fig.~\ref{fig:high_res4}, and Fig.~\ref{fig:high_res5} we provide some object removal instances in large images from 1k to 2k resolutions compared with LaMa~\cite{suvorov2021resolution}. Some cases are selected from the open-source testset of LaMa \textcolor{black}{and others are selected from our HR-Flickr}. Note that our method outperforms LaMa in scenes with weak textures such as row 2 in Fig.~\ref{fig:high_res1} and row 1 in Fig.~\ref{fig:high_res2}. For the cases with sparse regular textures and lines (rows 1,3 of Fig.~\ref{fig:high_res1}), our method can still achieve more clear borderlines. For the cases with dense regular textures (rows 2,3 of Fig.~\ref{fig:high_res2}), LaMa gets competitive results, which shows that FFC in frequency fields has solved these problems properly. However, our method can also achieve results with less blur that benefited from precise structural constraints. For the larger case with 2048 image size in Fig.~\ref{fig:high_res3}, our method can still get more consistent results compared with LaMa.

\end{document}